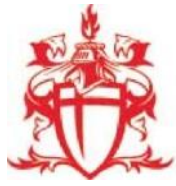

CITY UNIVERSITY
LONDON

**Segmentation of Soft atherosclerotic plaques using active**

**contour models**

TRANSFER REPORT

# MPHIL/PhD IN INFORMATION ENGINEERING

**Supervisor: Prof. PANOS LIATSIS**

**Prepared by: M Moazzam Jawaid**

**Department of Electrical and Electronic Engineering**

**City University London**

**November 2015**

# Table of Contents

## Chapter 5 Proposed Framework

## List of Tables

# Abstract

Detection of non-calcified plaques in the coronary tree is a challenging problem due to the nature of comprising substances. Hard plaques are easily discernible in CTA data cloud due to apparent bright behaviour, therefore many approaches have been proposed for automatic segmentation of calcified plaques. In contrast soft plaques show very small difference in intensity with respect to surrounding heart tissues & blood voxels. This similarity in intensity makes the isolation and detection of soft plaques very difficult. This work aims to develop framework for segmentation of vulnerable plaques with minimal user dependency. In first step automatic seed point has been established based on the fact that coronary artery behaves as tubular structure through axial slices. In the following step the behaviour of contrast agent has been modelled mathematically to reflect the dye diffusion in respective CTA volume. Consequently based on detected seed point & intensity behaviour, localized active contour segmentation has been applied to extract complete coronary tree. Bidirectional segmentation has been applied to avoid loss of coronary information due to the seed point location whereas auto adjustment feature of contour grabs new emerging branches. Medial axis for extracted coronary tree is generated using fast marching method for obtaining curve planar reformation for validation of contrast agent behaviour. Obtained coronary tree is to be evaluated for soft plaques in second phase of this research.

# Chapter 1    Cardio-Vascular Diseases & CTA Imaging

Recently non invasive imaging techniques have emerged as powerful tool for diagnosis of cardiovascular diseases. Use of cardiac CTA is prominent example which offers high spatial & temporal resolution. This accuracy makes CTA a substitute for complex catheterization process for detecting coronary abnormalities; however interpreting bulk amount of data is cumbersome task & depends on clinician's previous knowledge & expertise. Object segmentation aims to facilitate radiologist for quick diagnosis of abnormal & suspected regions. This chapter starts with basics of human anatomy (It is important because prior anatomical knowledge is often incorporated in segmentation process to speed up the computation). Followed with the importance of research question, a very brief review of segmentation process is presented. At the end of this chapter, aims and objectives of this work are presented.

## 1.1    Heart Anatomy & Coronary Artery Basics

Heart is located between two lungs in the centre of the human chest. Anatomically it is divided into 4 chambers each performing a specific task to uphold the blood circulation in the body. Strongest chamber of the heart is Left ventricle that is responsible for impelling blood through aortic valve to different organs of the body. In cardiovascular system heart works as a pump that circulates purified & contaminated blood simultaneously inside human body. A complex vessel network is used for blood transportation. According to purpose they serve, vasculatures are named as arteries (moving purified blood away from heart) or veins (responsible for dispatching polluted blood of organs to heart for purification). Heart itself is a muscular organ comprising of cells called Cardiomyocytes. For continuous contraction operation a rich supply of oxygen & nutrients is demanded by Cardiomyocytes. This is accomplished through coronary circulation i.e. provision of the oxygenated blood to the heart muscles. The constant motion of the heart is accommodated in terms of 'vasoconstriction', a process of cyclic peaks & troughs of coronary circulation that allows coronary arteries to adjust blood flow according to the requirement of the tissue and muscles. Failure to meet increased oxygen demand causes ischemia (condition of the oxygen deficiency) leading to the angina attack. Angina is the result of reduced blood flow to heart muscles and requires immediate restoration of blood flow otherwise the section of the heart begins to die as shown in Figure1.1(a).

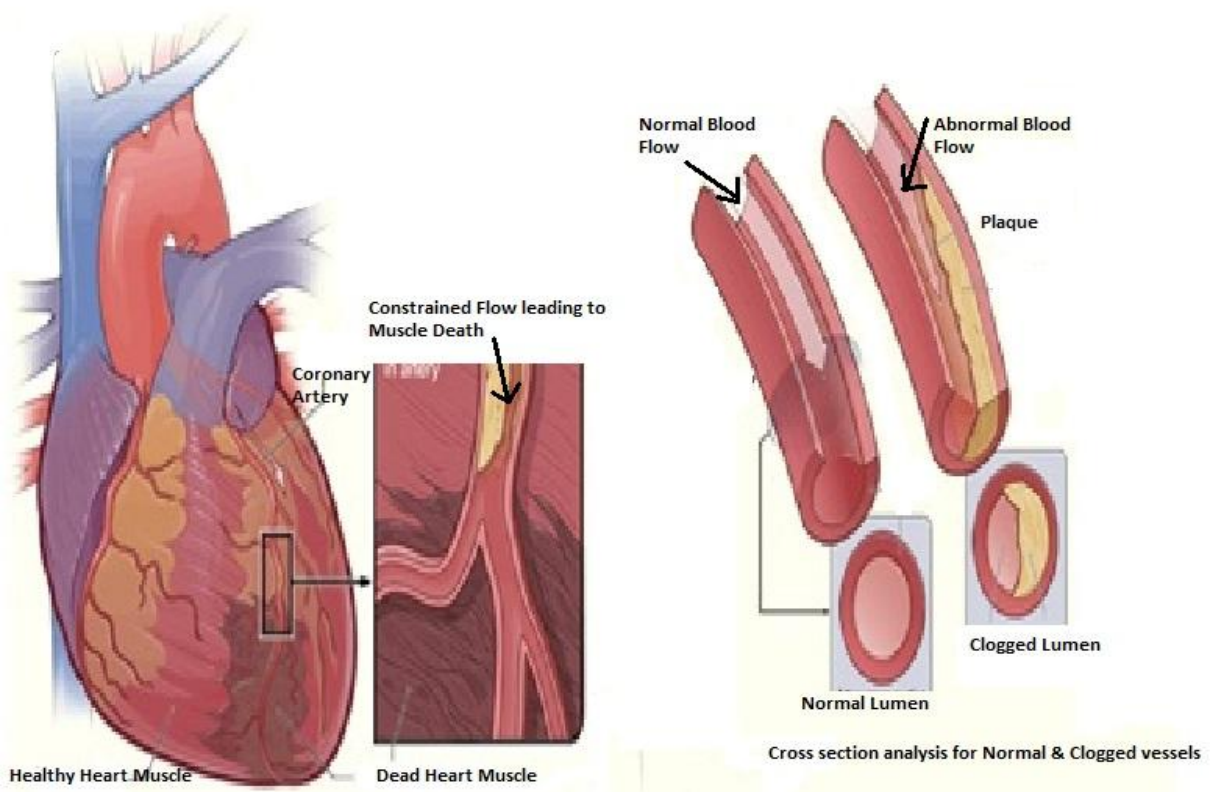

Figure 1.1 Cardiovascular diseases (a)   Death of heart muscle leading to Angina (b) Normal versus abnormal lumen

Coronaries run on the external surface of the heart and termed as "end circulation" as these are the only supply to the myocardium. In contrast to other organs, smaller arterial branches are very refined & do not offer interconnections for blood flow diversions. This makes the coronary blockage a severe threat leading to the myocardial infarction & fatal casualties.

 Coronary tree in human heart comprises of two (left & right) arteries that originates from descending aorta (main vessel coming out of left ventricle).  Left coronary artery (LCA) serves left chambers of the heart whereas right and posterior muscles are nourished by right coronary artery (RCA) with the help of posterior descending artery. For LCA, segment running from aorta to first bifurcation point is termed as left main (LM) artery. Two bifurcated branches are named as left circumflex artery (LCX) and left anterior descending (LAD) artery. Usually right artery branches into few marginal arteries (OM1, OM2) and posterior descending arteries (PDA). Sinuatrial nodal artery arises from RCA in 55% hearts whereas for remaining 45% it comes out of LCA [3]. Figure 1.2 (a) shows the left & right arteries & role of arterial tree in heart nourishment is depicted in (b).

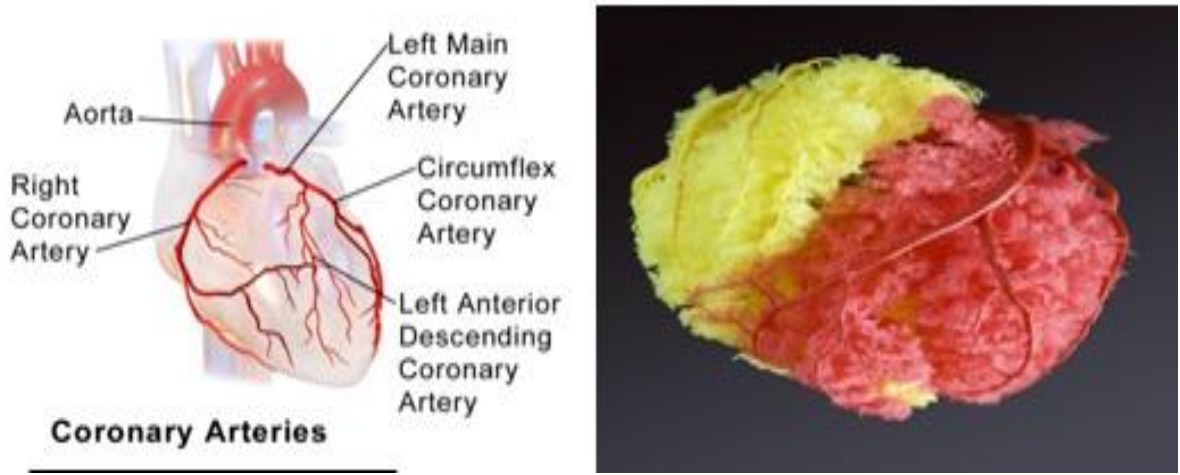

Figure 2.2 Coronary artery tree on heart surface (a) Coronary branches (b) contribution of coronary arteries in heart muscle nourishment

Instead of a following a standard architecture, wide inter patient variability has been observed that makes coronary tracking a challenging task. Some typical variations of coronary arteries & their anatomy can be found in [4]. Side dominance is used to classify subjects according to coronary behaviour. Dominance is determined by identifying the nutritional source to the posterior muscle of heart. Almost 70% of the population is right dominant where posterior tissues are nourished by right coronary artery. For 10 % cases, left coronary artery is dominant where as remaining 20% are identified as co-dominant where both left & right arteries are feeding PDA.

## 1.2    Introduction to Coronary Heart Disease

Cardiovascular Diseases (CVD) refers to all abnormalities of heart & related blood vasculatures in human body. CVD leads to different disorders including high blood pressure, stroke, and congenital cardiovascular effects. Recently CVDs have become a major cause of discomfort & sickness in developed countries. Approximately 40% of total deaths per year in Unites States are related with cardiovascular diseases [1]. According to NHS statistics, over 1.6 Million men & around 1M females are suffering from chronic heart disease (CHD) in United Kingdom. Each year CVD's claim about 88,000 deaths (an average of one death every six minutes). Moreover, around 124,000 heart attacks & 152,000 stroke attacks occur every year due to the cardiovascular abnormalities resulting in thousands of mortalities. Table 1 shows the percentage breakdown of deaths related to all CVDs. It can be observed from table.1 that coronary heart disease has become the worst enemy of human race contributing to almost 50% of total CVD related deaths.

Table 1.1 Percentage ratio of cardio vascular diseases death toll [2]

| Type of CVD Deaths in (%) | |
|---|---|
| Coronary Heart Disease | 49.9 |
| Stroke | 16.5 |
| High Blood Pressure | 7.5 |
| Congestive Heart Failure | 7.0 |
| Diseases of Arteries | 3.4 |
| Other | 15 |

Coronary heart disease (CHD) is a state in which fatty material builds up inside the coronary arteries as shown in Fig.1.1 (b). This fatty material (plaque) resides either inside lumen or within the wall & causes obstruction in the flow of oxygenated blood to the heart muscles. Development of arterial plaques is termed as "atherosclerosis" & it takes many years before it becomes intimidating. In case of calcified depositions, blood supply to heart tissues is significantly restricted leading to the myocardial infarction & Angina attacks. Especially left main coronary artery is very sensitive to occlusions due to the wall thickness of left ventricles. In contrast, development of non calcified arterial plaque makes vessels prone to sudden rupture and blood-clot deposition which consequences in sudden unexpected casualties. Coronary heart disease also weakens heart muscles leading to "arrhythmias" or heart failure i.e. heart fails to push blood to different body organs. Electrical activity or rhythm of the heart beat is also affected by coronary heart disease. Threat imposed by CHD makes it essential to diagnose coronary artery atherosclerosis at early stages of development. Accordingly, medications & surgical procedures can avoid or at least delay worst cardiac events in future. CHD identification requires sufficient clinical expertise & previous knowledge to interpret risk assessment tests. Specialized medical procedures have been designed for detecting coronary heart disease; however no single investigation can draw final conclusion & multiple investigations are advised for confirmation of CHD. For example, Electrocardiogram that is used for initial diagnosis of CHD monitors heart electrical activity whereas Stress test records blood flow, heart & breathe rate in excited state to identify abnormalities. Echocardiography test uses sound waves for imaging dynamic heart. These images are used for identification of any injury caused by poor blood flow. Based on the initial results, clinicians advise for coronary angiography that is an advanced and more reliable test for diagnosis. This invasive test makes use of contrast medium injection in arteries followed with special x-ray imaging. This process "cardiac catheterization" ensures the release of dye in coronaries that makes them prominent in the recorded images. Increased visibility in the x-ray images facilitates clinicians to trace blood filled coronaries & mark any restriction in blood flow.

## 1.3    Medical Imaging (Invasive versus Non-Invasive)

Medical imaging refers to phenomena of visual interpretation of internal body organs for clinical analysis & intervention. By combining radiology with imaging technologies, internal body structures are revealed for detailed examination. In recent years state of the art imaging techniques have been developed for imaging dynamic structures like heart. In context of clinical diagnosis, imaging methods are divided into invasive & non invasive categories depending upon the procedural requirements.

### 1.3.1   Invasive Imaging:

Invasive imaging refers to scheme involving induction of apparatus into body cavities. Some examples include X-ray angiography, optical coherent tomography (OCT) & intravascular ultrasound (IVUS). These methods are catheter guided techniques as it carries contrast medium to the desired locations inside vessels. An example 2D angiogram obtained from X-ray angiography is shown in Figure.1.3 (a-b).

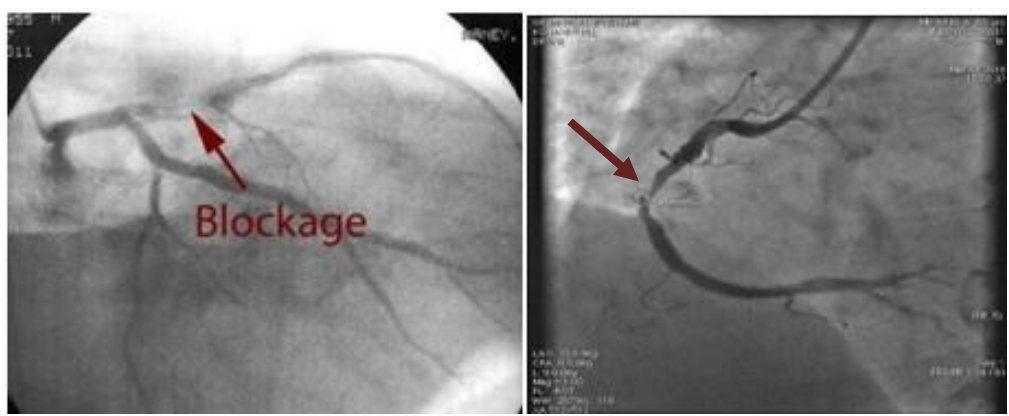

Figure 3.3 Vessel visualization for clinical diagnosis (a, b)   2D Angiogram showing vessel clogged regions

The ability to reflect luminal diameter and occlusion points are the strengths of this method & it remained as gold standard for CHD diagnosis for a long time. However 2D portrayal is the inherent limitation of this method which can underestimate the total plaque burden as well as vessel remodelling is not precisely reflected. The possible superposition of venous structures over the arterial tree can mislead clinicians during diagnosis. For effective 3D reconstruction of vessels, different variations of X-ray angiography have been proposed including rotational system for acquiring 2D angiograms at different angles. Obtained image sequence is used to create a 3D volume representing vascular structures. However, rotational mechanism involved in this technique brought trouble during reconstruction & image quality becomes inferior to a real CT scan. Catheter guider OCT and IVUS are also used for imaging vessel structures. These techniques allow direct and real time imaging of vessel occlusions by providing cross sectional views. Image acquisition is performed by recording signals from reflected laser beams (OCT) or ultrasound beams (IVUS). Generally OCT is capable of providing more accurate images because of higher spatial resolution (around $10\,\mu$m). Comparatively, OCT suffers from low penetration problem, as laser beams are attenuated quickly in arterial walls. A comprehensive review of IVUS, OCT is presented in [5]. OCT and IVUS are preferably used for vessel wall imaging where different layers of arterial wall & atherosclerotic plaque components can be visualized. Limitations associated with these methods are difficulty in reconstruction of 3D structures with the help of 2D cross sections. Besides, these methods are applied to small segments of vessels to reduce the complications of catheterization procedure, so it is not possible to trace all the important branches of the arterial tree.

### 1.3.2 Non Invasive Imaging:

Despite the fact that invasive imaging (x-ray angiography) acquires valuable coronary information, the shortcoming is involvement of complicated clinical procedures and associated risk to patient. Operational requirements make angiography a time consuming procedure & it demands high expertise of clinician. Specially, this method becomes impractical when motive is to track the progress of disease at regular intervals in routine examinations. Advancements in the computing technology have made 3D non invasive imaging of human body quite simple. Two state of the art imaging modalities being used for acquiring internal body details are magnetic resonance imaging(MRI) and computed tomography angiography (CTA). Besides being non-invasive, these imaging methods provide complete information about 3D structure of internal organs rather than 2D projection in conventional angiograms.

CTA is being used to capture 3D shape/behaviour information of different body organs including head, neck, abdomen and heart with precise details (sub millimetre resolution in all three dimensions). Especially for cardiac imaging, CTA appears striking technique as blood voxels & calcified components (plaque components) exhibit high density values. Along with blood filled vessels, Calcium deposits also become visible in CTA images which are not discernible in traditional angiograms. Due to 3D data acquisition, CTA can also be used for generating cross sectional images for lumen analysis as produced in OCT & IVUS. On the basis of medial axis, directional vector calculation helps to extract 2D orthogonal planes that facilitate lumen diameter & vessel wall estimation. 2D oblique plane is often constructed by re-sampling intensities in 3D space, so quality is inferior to OCT but still good enough for evaluating luminal changes. A state of the art development is multi phase CTA that captures image as distinct time-points over the cardiac cycle. It allows clinician to analyze individual coronary branch at best phase of the cardiac cycle. A substitute for non-invasive imaging of vasculature is magnetic resonance angiography (MRA) but due to less spatial resolution (high slice thickness) this modality is for recording larger vessels only. MRA based coronary imaging has not been reported yet.

Due to the amount of data generated, it becomes inconvenient for radiologists to analyze axial images one by one. Moreover, relating information from consecutive axial slices to establish substantial structures is also cumbersome task. A techniques for purposeful interpretation of CTA data is termed as digital subtraction angiography. Basically DSA is a visualization procedure used to apprehend blood filled vessels by subtracting normal CTA image from a contrast enhanced version. In practice this procedure becomes difficult because of requirement of two CTA data sets that leads to additional exposure to radiation. Another issue related is the registration of two data sets for precise subtraction. It is very difficult to obtain identical images of coronary arteries due to the constant motion of heart. Therefore despite of the advantages DSA offers, it is not widely used for cardiac imaging & visualizations. Here arises the need of robust segmentation & visualization algorithms, i.e. a precise combination of segmentation & visualization techniques can help clinician in accurate & fast diagnosis of coronary related abnormalities.

## 1.4 Cardiac Computed Tomography Angiography (CTA)

CT imaging technique associate special X-ray equipment with sophisticated digital geometry processing for generating 3D image of inside of an object. 3D construction is done from a sequence of 2D mages acquired around a single axis of rotation, i.e. a number of images of same area are recorded from different angles and placed together to produce a 3D image. Ability to provide fast & precise internal details has made CT exam ultimate choice for clinical diagnosis of body organs like head, neck, abdomen & cardiac chambers. Despite of impressive results, CTA scan has not been used widely for imaging cardiac system in last decade because of the "motion" of heart. Coronary arteries being very refine structures with diameter in sub-millimetre range demands high spatial resolution acquisition systems. Moreover high temporal resolution is required that can emulates dynamic heart as "static organ". Recently, state of the art advancements in medical imaging has resolved this problem by introducing sub-second rotation combined with multi slice CT that guarantees high speed & high resolution at same time. Dual source technique has reduced the acquisition time by imaging data in half rotation whereas multi-detector approach helped to increase spatial resolution. This makes CTA a clinical reality for assessment of cardiac vascular system (i.e. imaging the dynamic heart for coronary analysis). ECG gating is used during CT scan process to achieve synchronization between heart motion & image acquisition / reconstruction. Every potion of the heart is imaged multiple times along with ECG traces & corresponding phases of cardiac contraction are correlated by using ECG data. After correlation, systole related data is discarded & images are constructed using static phase (diastole) data. Figure 1.4 illustrates the acquisition process of CT data showing axial slice dependence on ECG values. For prominent visualization of blood filled coronary arteries iodine based contrast medium is injected prior to scan (sometimes can be dangerous for kidney patients).

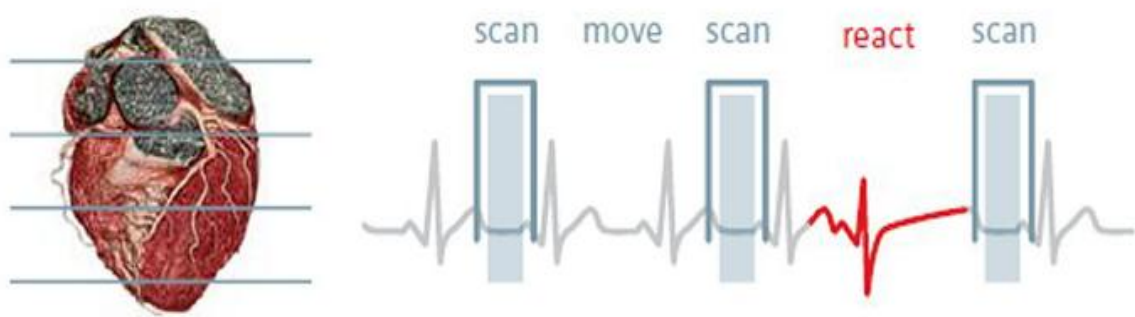

Axial Slices

Figure 4.4 ECG triggered CT scan for compensation of heart motion, (Each slice is scanned during same ECG phase).

Images for different body organs in CTA are recorded on the basis of "ability to block x-rays", called as radio-density. CT scanner records the attenuation through a plane of finite thickness for whole cross section. Every component of this cross section image (pixel) represents the mean attenuation value of the segment. Mathematically attenuation is represented as

$$I_t = I_0 \, e^{-\mu.\Delta x} \qquad (1.1)$$

Where $I_t$ represents attenuation value for object under examination, $I_o$ represents intensity measured in the beam path without any obstruction & $\mu$ is the linear transformation coefficient for specific material. Third dimension information for volumetric objects can be incorporated by adding all attenuation values along the beam path according to equation1.2.

$$I_t = I_o e^{-\sum_{i=1}^{k} \mu_i \Delta x} \qquad (1.2)$$

In clinical CTA dataset attenuation value is represented in terms of HU scale [-1024 to +3071] that transforms intensity value into a Standard temperature pressure(STP) metric where distilled water is assigned 0 HU & air has -1000 HU. Mathematical model for HU transformation is given in Equation (1.3).

$$HU = 1000 * \frac{\mu - \mu water}{\mu - \mu air} \qquad (1.3)$$

For medical diagnosis & investigation, HU values are reserved according to the behaviour/structure of the organs. Lungs=-500, Fat=-100 to -50, Blood=30 to 45, soft tissues=100 to 300. These reserved values are useful during investigation of abnormalities in different organs.

## 1.5    Aim & Objectives of this work

In a broader view this work aims to develop an automated framework for diagnosis of atherosclerosis in coronary arteries. Application will perform 3D reconstruction of the arterial surface(s) from contrast enhanced clinical CTA data set. Suspected clogged locations in major branches of the arterial tree will be tracked with a focus on soft plaques. In first phase of this work, fully automatic seed detection mechanism will be developed to make this application robust & independent of clinician prior knowledge. Moreover, segmentation process will be guided by improved vessel tracking filter that will incorporate the local behaviour of contrast medium in respective CTA volume. In the second phase of this work, soft plaque detection & quantification will be performed. Usually, soft plaques resides inside vessel walls, so focus of this stage will be geometric modelling of vessel walls to record & detect abnormal behaviour. Geometric shape analysis combined with intensity information will allow delineation of non-calcified plaques in CTA volumes which are not identifiable easily due to close similarity with contrast enhanced blood. This work will effectively help clinicians to suggest preventive measures well in time to avoid worst cardiac events in future. A brief summary of research objectives is given below.

### 1.5.1 Automatic Detection of Seed points

High spatial & temporal resolution of CT scanners produces a bulk amount of data which is very useful for revealing internal information; however it become tedious for clinicians to explore 2D cross sectional axial slices for detecting abnormalities. Interestingly coronaries represent only 2% of data cloud that contains a number of irrelevant internal organs having same visual appearance. It becomes difficult for clinicians to quickly isolate the coronary voxels (segments) in a 2D slice. Even manual selection of coronary points from a 2D axial slice remains suspected because it involves previous knowledge of radiologist whereas huge inter-patient variability in coronary architecture has been observed. This work aims to detect the coronary points automatically without interaction from radiologist. Geometrical shape analysis in combination with available anatomical knowledge will be used in this work to identify coronary pixels from 2D axial slice of contrast enhanced CTA. Identified coronary pixel will be used in subsequent stage as seed points for 3D surface construction of complete arterial tree.

### 1.5.2 Vessel Filter For tracking coronary

Non vessel structures found in CTA volume including aorta & blood filled heart chambers should be suppressed for effective delineation of coronary arteries. Geometric & shape characteristics are examined usually to extract the structure of interest from data cloud. Hessian matrix based Eigen value analysis will be used for detection of vascular /tubular objects in an image. An improved Eigen-vector driven method will be used for tracking coronary arteries in CTA which follows a cylindrical model in axial view. The proposed vessel enhancement measure (incorporating intensity information) will be used in to ensure that only coronary based voxels are selected during curve evolution process.

### 1.5.3   3D Arterial Surface Construction

Based upon the automatic seed detected in step 1.5.1, region based active-contour model will be enforced for precise segmentation & 3D reconstruction of arterial tree from CTA data. Inherent problem of intensity in-homogeneity in medical images will be addressed by incorporating localized regional intensity information in curve evolution. Proposed two-way segmentation (forward & backward in axial direction) will be controlled by described vessel tracking filter for effective recording of coronary progression to proximal & distal endpoints respectively.

### 1.5.4   Centreline Extraction & MPR Analysis

After precise 3D reconstruction of coronary arteries, clogged locations will be identified in the subsequent stage. Centreline of the respective arterial tree will be generated by applying fast marching implementation of sub voxel skeletonization technique. Based on skeleton centreline, multi planar reformation (MPR) will be used for vessel branch representation in real time. For investigating localized vessel behaviour in terms of lumen & vessel-wall geometric information 2D orthogonal cross sections will be extracted (guided by direction vector of the vessel). A combination of intensity & geometric reasoning will be applied for detection of abnormal arterial regions (containing plaque or remodelling) in phase II.

### 1.5.5   Soft Plaque segmentation & Quantification

In final stage of this work, potential clogged locations will be categorized into hard & soft plaques. Often, hard plaques are clearly identifiable due to associated high intensity that makes them visible in CTA. Non-calcified plaques normally lie inside vessel walls & cannot be identified easily. Critical geometric analysis of arterial wall will be performed for segmentation & quantification of non-calcified plaque & results are to be validated by cardiac imaging expert.

## 1.6    Organization of the report

Chapter 1 introduces the basic theme of this work. Importance of the research problem is highlighted with the help of statistical data & visual evidences. Starting with human cardiac anatomy, an overview of medical imaging techniques is presented. It is followed with a focused analysis of Coronary CTA mechanism as imminent research is based upon CTA data. Chapter 2 addresses the basic problem of image segmentation. Existing algorithms starting from simple threshold based classification to state of the art geometric active contour models are briefly reviewed. Explicit curve representation in terms of level set formulation is explored in succeeding section. Chapter 3 highlights intensity in-homogeneity problem that is specifically associated with medical images. Different proposed solutions and their drawbacks are presented from literature in this chapter. Atherosclerosis detection and quantification problem is discussed in chapter 4. Developments for segmentation of the calcified and non calcified plaques are reviewed in this section with a special focus on "soft" plaque delineation. Chapter 5 represents the proposed framework & initial results obtained in phase-I of this work. Automatic seed detection mechanism is illustrated followed with segmented coronary tree for 12 CTA volumes. At the end of chapter sub-voxel based skeleton & re-sampled cross sectional planes are presented (to be used in next phase of this work for vessel wall analysis).

# Chapter2        Image Segmentation Methods

Segmentation refers to mechanism of dividing a digital image into various pieces (sub images) i.e. set of pixels. Ultimate goal of segmentation process is to present an image in a meaningful way that can be analyzed easily. In short, segmentation process assigns a label to individual pixels of image such that pixels having same label belong to one object (share certain features). Result of segmentation is bunch of segments that cover the entire image whereas adjacent regions differ significantly with respect to certain features. For 3D data cloud, isolated contours can be used to construct solid surface(s) using interpolation algorithms for individual objects. Image segmentation has been applied in several fields such as conception and visualization, satellite imaging, intelligent transportation systems and biomedical imaging. Among them most important application is medical imaging where it can be used as an effective tool to isolate vascular pathologies in imaged data like detection of tumour & vessel remodelling index.

Generally gray level images contain enormous amount of data, much of which is irrelevant e.g. the background of the scene is always unwanted. For effective analysis of local features of particular object, segmentation is necessary i.e. to extricate ROI from unwanted background in the image. For instance, cardiac CTA data to be used for diagnosis of coronary heart disease (CHD) represents a number of organs in data cloud. They exhibit similar visual behaviour because of same intensity characteristics. It makes difficult to trace small individual organs through a sequence of consecutive axial slices. The object of interest for clinician in cardiac CTA is coronary artery which constitutes only about 2-2.5 % of whole data cloud. Segmentation process can effectively differentiate coronaries from other anatomical organs on the basis of certain metrics. For Example all unwanted components of the image can be suppressed on the basis of visual appearance (intensity), connectivity or texture information. Depending upon the complexity of data set, a combination of features (intensity & geometric) can be used for effective delineation of object boundaries. This chapter starts with the classification of image segmentation methods where a brief review of different methods is presented in section 2.1. A comprehensive revision of active contour model, their geometric deformable counterpart & level set formulation is presented in section 2.3-2.5 as the subsequent research is based upon these ideas.

## 2.1 Classification of Segmentation Algorithms

Different segmentation techniques rely on certain image features for associating pixel similarity; however it is notable that no single method perfectly works for all images. For successful extraction of objects from image (volume), domain specific knowledge is usually associated to make segmentation process robust & realistic. Categorization of general segmentation algorithms is presented here, whereas detailed review of specific techniques related to this work will be presented in the subsequent section.

### 2.1.1   Threshold Based Segmentation

Straightforward method for image segmentation is termed as "Thresholding". This method is based on a cut-off value to transform a complex gray-scale image into binary one. The key of this method is selection of threshold value (value(s) in case of multiple-levels to allow intermediate gray shades between white and black colors). Popular threshold based segmentation methods include Otsu's method (maximum variance) and maximum entropy method.

### 2.1.2   Clustering Based Segmentation

Cluster analysis is task of grouping a data set such that objects in the same group (cluster) show more similarity (in some sense or another) to each other, than to those in other groups (clusters). Mainly this is exploratory data mining task used for statistical analysis in machine learning, information retrieval & pattern recognition. One famous clustering example is "K-means" cluster that assigns different pixels to obtain clusters on the basis of "difference" from cluster centre (difference based on texture, intensity, distance etc).

### 2.1.3   Histogram-based methods

In comparison to conventional approaches, histogram based method are more efficient. They typically require only one pass through the pixels & global behaviour of the image is used in classification. Peaks & valleys of histogram are employed to establish cluster centres. In the following stage, object segmentation is achieved by cluster analysis that minimizes the difference with respect to cluster centre. Recursive application ensures the optimal segmentation by obtaining meaningful cluster in every iteration.

### 2.1.4    Edge based methods

Edge refers to image point where a significant intensity discontinuity is present. Object boundaries are usually associated with edges (gradient shift) since the intensity profile shows an abrupt change in intensity at regional boundaries. Edge detection can be used as effective technique for object segmentation but it is prone to image noise & weak intensity variations. Indeed edge detection itself has become a developed field in digital image processing. It can be formulated as a binary-classification task at pixel level with a motive of identifying individual pixels as edge or non-edge entity. Successful edge detection leads to quality segmentation whereas weak & degraded edges results in leakage and over segmentation. Different methods for edge based segmentation utilize edge detection including Sobel, Canny, Prewitt or Roberts detectors.

### 2.1.5    Region-growing methods

Region-growing is based on the fact that pixels inside one region share similar intensity behaviour. Approach is to equate pixel intensity with its neighbours for satisfaction of some similarity criteria. On successful validation pixel is assigned similar label & vice versa. The selection of the similarity criterion is very important for successful extraction of required object and presence of noise degrades the quality of segmentation. Statistical region merging (SRM) method merges a pixel with its 4 neighbours subject to criterion fulfilment. Another famous approach used for segmentation is **"seeded region growing"** based on seed points for every object in image. Region starts growing by integrating pixels satisfying the difference criteria i.e. (Pixel intensity – Region Mean Intensity). Efficiency of this method is based upon the proper selection of seeds. Another variation "**Split-and-merge**" technique is based on a quad-tree partition that starts from the root (whole image). On non-homogeneity, image is split into four/eight squares whereas homogenous squares are combined to form one component.

### 2.1.5    Partial Differential Equation Based Methods (Deformable Contours)

Several algorithms have been reported for image segmentation that relies on partial differential equation (PDE) solution. The fundamental idea is to evolve a curve iteratively approaching lowest potential of a cost function whereas cost function is defined to reflect the intended task, for instance minimization at region boundaries. The smoothing constraints of PDE solution are imposed in terms of geometrical constraints on curve deformation. Two different approaches including implicit & explicit representation have been investigated for evolving curve. Parametric representation termed as **"Active Contour"** model performs fast & efficient segmentation but it is unable to handle topological changes. In contrast, **"Level set"** method handles the segmentation problem more accurately because of implicit representation of evolving curve but demands more computational resources. Majority of the recent image segmentation approaches are based on level set formulation due to easy implementation in discrete systems.

### 2.1.6  Trainable segmentation

Mostly segmentation methods rely on the colour (intensity) information for object labelling however in real life humans use much more knowledge in differentiating objects. Incorporating this knowledge in computer based segmentation is not feasible as it demands huge domain-related knowledge base & state of the art computational resources. Trainable segmentation methods suggest the integration of useful knowledge in object delineation. NN segmentation processes image in terms of small areas using neural network. In the following stage decision-making mechanism labels the image pixels accordingly. Kohonen Map is one major example of NN based segmentation system whereas PCNN (pulse coupled neural networks) also performs biomimetic image-processing.

### 2.2 Threshold based Segmentation

Thresholding is the simplest image segmentation method. It can be defined as a mapping operation for the existing (gray level) value of pixels, given by g(v) such that

$$g(v) \ = 0 \ if \ v < t \hspace{3cm} (2.1)$$
$$= 1 \ \text{if v} \geq \text{t}$$

Where **(v)** denotes the intensity value of a pixel, and **(t)** represents the pre-determined threshold value. Although there are several automated methods available for threshold selection but the most effective is to set threshold value interactively, by using the hit & trial technique until a required segmentation has been achieved. Histogram analysis is often used for selection of the appropriate threshold value as it represents the overall behavior of the image. The output image is a binary version having two distinct segments identified by the labels 0 (background) and 1 (object) respectively. Multiple objects can be segmented by extending equation (2.1) for multiple thresholds. The resultant object detection is shown in figure (2.1) where multiple objects are segmented in (d). In case of ambiguous images where foreground cannot be distinguished easily from background, threshold based segmentation fails & results in loss of data.

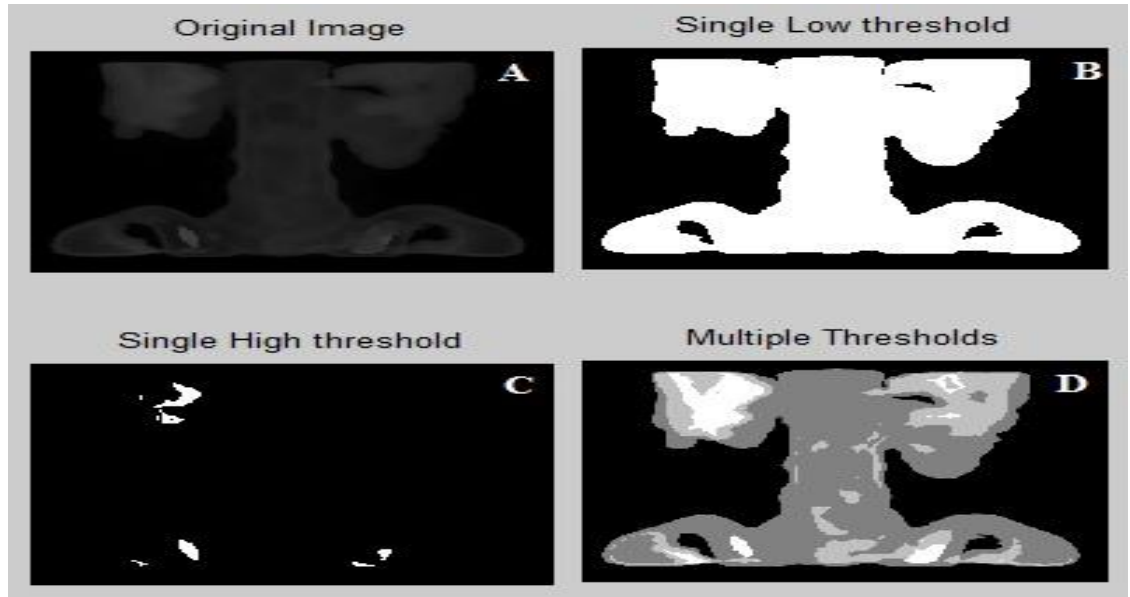

Figure 2.1 Threshold based segmentation (a) Original Image (b) Single LOW-threshold based segmentation (c) Single HIGH threshold based segmentation (d) Multiple threshold based segmentation

Selection of the threshold value is very crucial for final segmentation. For instance, in case of notable difference between object & image background, **"Mid"** of two peak values is often selected from bimodal histogram. More efficient threshold can be established by using Equation 2.2 that selects **"Minimum"** value between two peaks of histogram as shown in figure 2.2(a). This threshold selection strategy ensures the significant differentiation of pixels in background & foreground.

$$t = arg_{v \epsilon \{p1, p2\}} \min H(v) \tag{2.2}$$

Where H($v$) represents histogram value at intensity ($v$) & it is assumed that a that **p1** is smaller than **p2**. An improved threshold value can be established by using optimal threshold method as illustrated in Figure 2.2(b). This method suggests the Gaussian fitting of image histogram to obtain individual distribution for object & background. Cut-off value is chosen as intersection point of Gaussian curves, that ensures minimization of segmentation error i.e. the number of pixels to be mis-segmented.

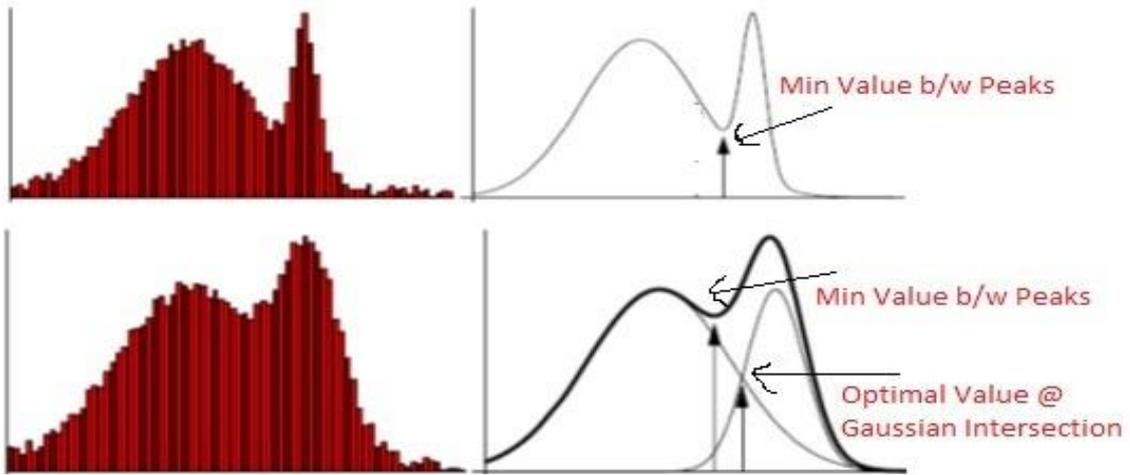

Figure 2.2 Histogram based threshold value (a)Minimum between two peak values (b) Optimal threshold selection by Gaussian curve fitting

Another efficient technique called K-means clustering establishes threshold value dynamically by minimizing the variance between object pixels. It is an iterative process than converges to the optimal segmentation by assigning pixels to segment such that internal segment variance decreases sequentially as shown in figure 2.3(a-h). Initially image is divided into K segments using (K − 1) static threshold values, followed with step wise subsequent optimization. The within-segment variance $\sigma^2_w$ is defined by

$$\sum_{i=0}^{K-1} hi.\sigma_{yi^2}\ \sigma_{w^2} = ho\sigma_{0^2} + h1\sigma_{1^2} + h2\sigma_{2^2} + h3\sigma_{3^2} \tag{2.3}$$

Where h represents the normalized histogram of the image, & variance is obtained using mean gray value of image.

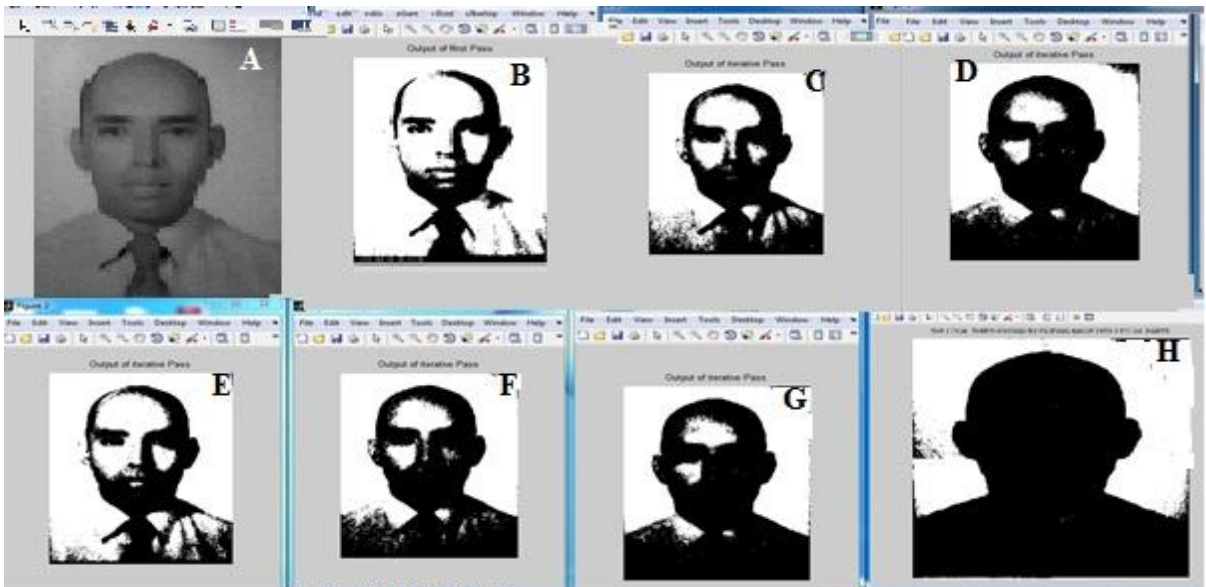

Figure 2.3 Illustration of K-means algorithm for image segmentation. (A) Original image for segmentation (B) output after 1st iteration. (C-H) improved segmentation after subsequent iterations

Conventional threshold segmentation methods fail to handle intensity artifacts in image. For instance unexpected shift in illumination across image region considerably degrades segmentation quality. Due to the intensity variation global histogram becomes diffused & consequently no single threshold can generate good segmentation for particular image. Threshold selection process can be further improved by incorporating spatial information of image i.e. bad lighting conditions & impact of noise can be repaid by localized thresholding. "Adaptive", i.e. threshold value is selected based on localized regional behavior of image to handle unexpected intensity gradients. Instead of using a global threshold for whole image, image is divided into 4 or 8 sub-images for addressing intensity variations. Consequently grey values in each tile remains relatively constant as gradient shift becomes relatively small in each sub image. Now each sub-image is treated individually by finding a local threshold based on localized gray level values in a restricted neighborhood. At the end the results are merged into a single output image by using Boolean operations. Figure 2.4 shows the advantage of using local/adaptive threshold over global to recover missing information.

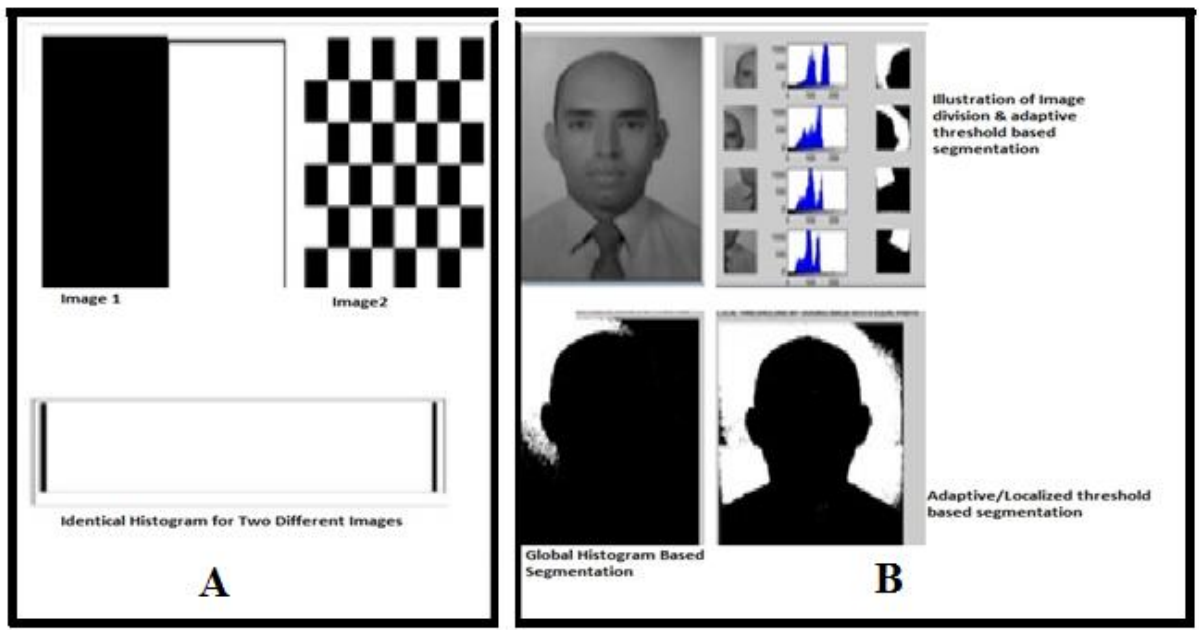

Figure2.4  Failure of Global histogram approximation in threshold value selection. (A)Two distinct images & their identical histogram  (B) Illustration of adaptive threshold based segmentation

## 2.3 Edge based Segmentation

Edge based segmentation is accomplished by employing simple idea that objects in binary image can be fully represented by its regional boundaries. Edge detection process can help in substantial classification of image contents by isolating individual objects. This process is very important because success of algorithms for higher level processing heavily relies on good edges. Usually edge detectors are designed to respond image discontinuity in terms of intensity gradient or texture which often occurs at regional boundaries. Detected edges are inspected for closed contours to identify solid shapes. Intelligence based operations are generally incorporated at this stage for bridging small gaps & removal of false edges. In the final step, all edges are filled for appropriate shape representation. Algorithm for edge based segmentation mechanism is presented below.

Start with given 2D image f(x, y).

1. Obtain 2-D gradient image ∇f(x, y) by using suitable gradient operator.

2. Threshold gradient image to obtain binary edge map **(∇f(x, y))t**

3. Compute Laplacian of image **Δf(x, y)** to sharpen the original image, using Laplacian of Gaussian.

4. Computer output image g(x, y) = (∇f(x, y))t * sign(Δf).

5. Fill established boundaries to identify shape of objects.

G(x, y) returned after application of sign operator, will contain +/-1 (start / end) for edge pixels whereas all non edge pixels will be suppressed by assigning value 0.

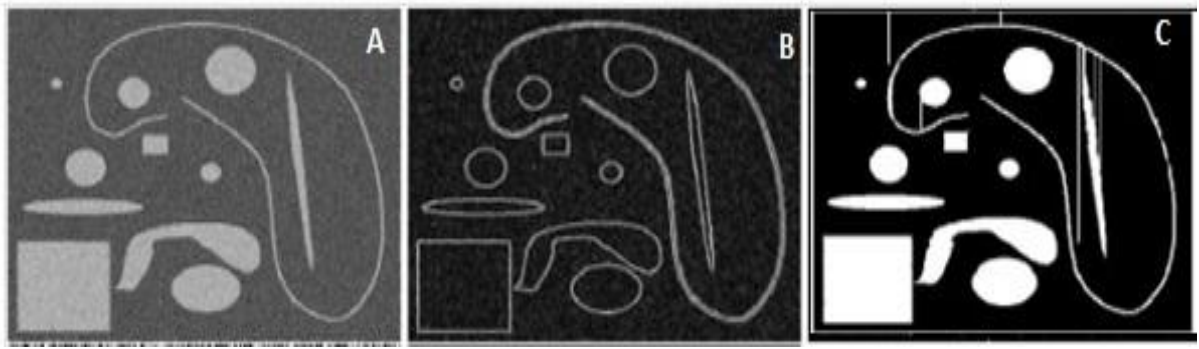

Figure2.5. Step wise example of edge-based segmentation.(A) Original Image (B) Edge detection (C) Filled segmented objects having closed boundaries

Mathematically, gradient calculation is the most effective way of locating discontinuity in image intensity values. Fortunately, it provides both magnitude and directional information of regarding edges. Gradient value itself represents the magnitude strength whereas the direction can be determined by rotating gradient direction by 90 degrees. Noise sometimes behaves as sudden discontinuity in images & lead to false edges, so smoothing filter is usually applied in preprocessing stage of edge detection. In the smoothed image, differentiation is applied using finite differences to record intensity variations in spatial directions respectively, as defined in equation 2.3 - 2.4.

$$\nabla f_x = \frac{f\left(x + \Delta x, y\right) - f\left(x, y\right)}{\Delta x} \tag{2.4}$$

$$\nabla f_y = \frac{f\left(x, y + \Delta y\right) - f\left(x, y\right)}{\Delta y} \tag{2.5}$$

Above equations represents derivative for continuous function. In case of discrete images, minimum step distance in respective direction is set equal to 1 i.e. $\Delta x = \Delta x = 1$.

$$\nabla f_x = \frac{f\left(x + 1, y\right) - f\left(x, y\right)}{1} \tag{2.6}$$

$$\nabla f_y = \frac{f\left(x, y + 1\right) - f\left(x, y\right)}{1} \tag{2.7}$$

Images are two dimensional therefore intensity variations in both directions are calculated to reflect the net gradient/shift in pixel values. The derivative of two dimensional image f(x, y) is term as Gradient vector G(fx, fy) where fx & fy represent the partial derivatives in the respective directions. Magnitude and direction for gradient can be obtained as follows.

$$\left|\nabla f\left(x, y\right)\right| \equiv \sqrt{\left(\nabla f_x^{\,2} + \nabla f_y^{\,2}\right)} \tag{2.8}$$

$$\theta(x, y) \equiv \arctan\left(\frac{\nabla f_x}{\nabla f_y}\right)$$

Edges are established by comparing normalized gradient magnitude with a suitable threshold value as defined in equation 2.9.

$$E(x, y) = \begin{cases} = & 1 & If\ N(x,y) > T \\ = & 0 & Elsewhere \end{cases} \tag{2.9}$$

Where

$$N(x, y) = \frac{M(x, y)}{Max\ M(i, j)\ i, j = 1 : n}\ x\ 100$$

Different operators exist for gradient calculation. Simple mechanism proposed by Sobel, Prewitt computes gradient at a pixel location along the horizontal and vertical principal directions. For optimal results some operators also accounts change in diagonal & anti-diagonal directions.

The gradient based edge detection usually results in frequent gaps between edge pixels whereas segmentation aims to represent distinct objects separated by closed boundary contours. Moreover false edges appear in edge map representing the influence of image noise and quantization error. In final stage of segmentation process these gaps are filled & false edges are optimally removed for precise segmentation. Hough transformation helps to identify & track objects in edge map if prior shape information is available, however it is rigid with particular defined features & does not show flexibility to incorporate slightly different shapes. Neighborhood search is another method for linking nearby pixels to establish edges. However the efficiency of this method depends upon the linking criteria. It can be simple scalar value or a compound metric integrating intensity, gradient & directional information. Constraints imposed on linking criteria ensure the linkage on right path at the cost of computational complexity.

## 2.4     Region based segmentation

Edge based segmentation detects individual objects by delineating outer boundaries whereas region based segmentation interprets individual objects as consolidated regions. Starting from a single identified point, the object shape is adapted gradually by including all the pixels that satisfy pre-defined similarity criteria (texture, intensity, shape). Although the efficiency depends upon the nature of image, but in general region based segmentation has shown more potential for accurate segmentation for different image classes. Especially noise induced images are handled more efficiently by region based algorithms & leakage is avoided that generally occurs in aforementioned method. Two basic operations at the core of region based mechanism are 'split' & 'merge' of image regions. Majority of the algorithms utilizes one of these operations; however combination of these two operations can produce more accurate segmentation at the cost of computational resources. Merging operation begins with the assumption of over segmentation in image i.e. every element is assumed a different object. It then starts fusing adjacent pixels having similarity until no more merging is possible. In contrast, splitting is based on the assumption of under segmentation. It starts splitting image regions having dis-similarity into distinct regions until no more splitting is possible. Similarity criteria used is important & generally rely on intensity value conformity. Edge strength factor can be incorporated in similarity test for optimized segmentation results (for instance, a weak edge implies similarity between two segments and a strong edge represents separate identity of two adjacent segments).

Region growing algorithm is practical implementation of merging segmentation. Starting with a set of pixels (seed values), neighborhood is scanned for pixels that fulfill the similarity criteria & added consequently to segment. This neighborhood search is recursively performed for recently added pixels till no pixels can be added to segment. Fig. 2.6 shows the region growing implementation for single seed based on intensity value.

<div align="center">

**Region Growing Algorithm**

</div>

1. Initially break down input image into several small segments such that H (Ri) = TRUE, for i =1, 2, 3...S
2. Detect seed point for initialization of segmentation (x, y)
3. Search neighbourhood around seed point (s) & apply similarity criteria test
4. Add successful pixels to seed point list & merge pixels in segment
5. Repeat step 3-4 until seed point list become empty
6. Segmented image output is H (Ri) = TRUE, i =1, 2, 3...S

It is extremely valuable to use region intensity heuristics in combination with edge based rules to ensure the best possible segmentation. An image may contain two structures having similar intensity values but separated with strong edge. In this case two individual objects should be delineated instead of merging them into one.

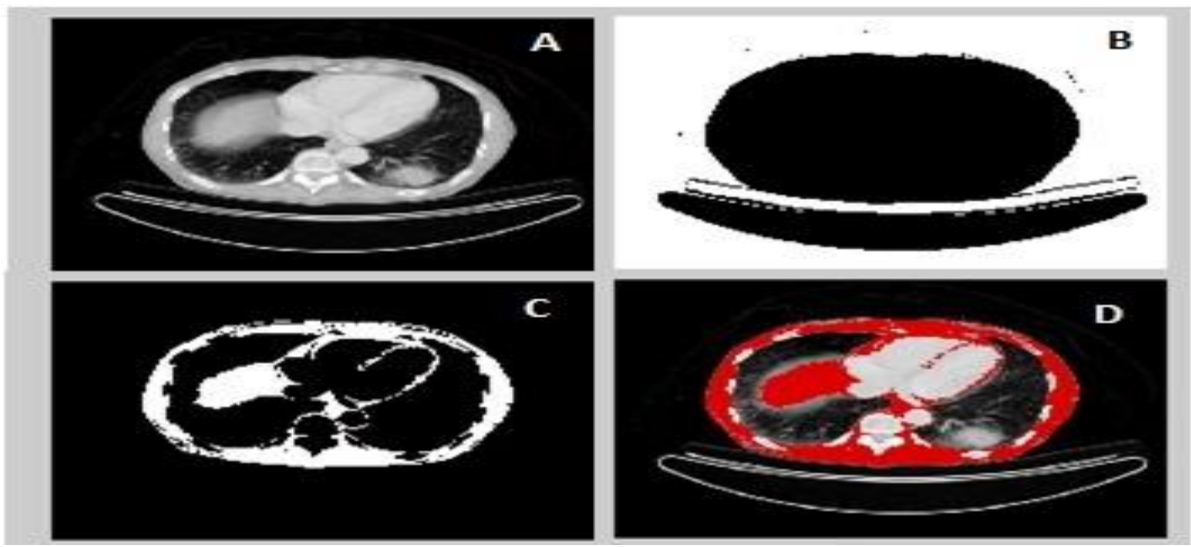

Figure 2.6. Implementation of Region Growing with single seed. (A) Original Image (B) segmentation with random seed point.(c) segmentation with object based seed point (D) segmented structure highlighted

Apparently "split" and "merge" operations seems quite similar, but a fundamental difference makes splitting operation technically complex. Merging is standard combining procedure, which adapts the shape of the objects present in image as segment grows. In contrast, standard splitting does not follow the shape of objects but dividing component is replicated. Figure 2.7 (a-c) shows the splitting mechanism where the object boundaries are violated leading to poor segmentation due to square based splitting. Consequently, "splitting" operation is often used in combination with "merging" in a sequential manner to minimize the shape deficiencies. However small 'blocky' impact can be observed in the final segmentation figure 2.7 (e). This technique fails is different similarity criteria is adopted for two operations.

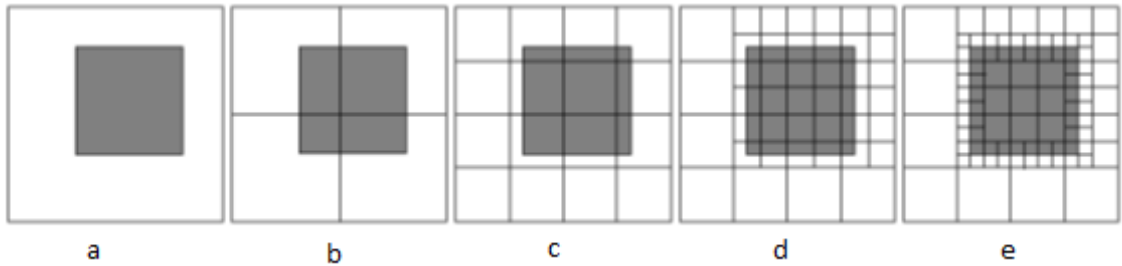

Figure 2.7. Implementation of split & merge segmentation

**Algorithm 3.2: Split and merge**

1. Start with initial segmentation ( various regions), establish a homogeneity criteria & initialize a pyramid tree structure.

2. In case the initial region is not homogeneous i.e. ($H(R) = FLASE$) then split into 4 children squares & vice versa. Once splitting & merging is completed goto next step.

3. Merge all adjacent regions satisfying the homogeneity criteria (no matter if they belong to different pyramid levels).

## 2.5    PDE based Deformable Models

### 2.5.1   Classical snake Model

An effective delineation technique based on edge information is named as "Active Contour Model". Originally proposed by Kass *et al* [6], segmentation process is defined as evolution of a dynamic spline to catch useful features inside an image. Propagation speed & direction of the curve growth is influenced by image characteristics in terms of internal & external energy. Internal energy is derived from curve itself (smoothness & elasticity) whereas external energy is realized from image. Individual objects can be segmented by active contour snake effectively as image based external energy pushes curve towards object boundaries. An appropriate combination of internal & external energy leads to a smooth segmented object. Mathematically, active contour can be described as parametric curve defined as

$$C(s) = \big(x(s), y(s)\big) \qquad\qquad 0 < s < 1 \qquad\qquad (2.10)$$

Intermediate values of "s" define the curve control points representing deformable snake. Evolution of the snake towards object borders can be interpreted as energy minimization problem of matching deformable model to object boundaries in an image. Designed cost function produces optimal values at high gradient (associated with object boundaries). Total energy of the parametric snake can be defined as in Equation 2.11

$$E = \int [E_{int}(c(s)) + E_{ext}(c(s)) + E_{con}(c(s))] ds \qquad (2.11)$$

$E_{int}$ represent the internal force of the curve, $E_{ext}$ refers to image behavior in terms of intensity information and $E_{con}$ represents external constraints imposed by user by employing prior knowledge to speed up the computation. The Internal energy reflecting the snake behavior can be modeled according to equation 2.12. Sharp twists & lengthy spans increases this term, so minimization ensures a smooth concise curve.

$$E_{int} = \alpha(s) \left| \frac{dc(s)}{ds} \right|^2 + \beta(s) \left| \frac{d^2 c(s)}{ds^2} \right|^2 \qquad (2.12)$$

The norm of the first derivative $\frac{dC(s)}{ds}$ represents elasticity & second derivative $\frac{d2.C(s)}{ds2}$ represents curvature measure. Two constants ($\alpha$ & $\beta$) controls the importance of individual factors (elasticity & stiffness) in overall cost calculation. Setting $\alpha = 0$ may lead to infinitely long snake whereas low values of $\beta$ will allows sharp twists in curve. An appropriate combination of two values ($\alpha$ & $\beta$) is application sensitive & obtained after experimentation. $E_{ext}$ represents the external energy (derived from image) that forces the snake towards specific features of image including line, corners & terminations. Energy value is modeled mathematically such that it decreases as snake comes closer to a particular feature of interest in image. Required features can be assigned high weights (significance importance) in the cost function defined by Equation 2.13.

$$E_{ext} = w_{line} E_{line} + w_{edge} E_{edge} + w_{term} E_{term} \qquad (2.13)$$

$$E_{ext} = -W. ||\nabla f||.^2 \qquad (2.14)$$

For instance, edges are most important features in images. External energy for establishing edges can be represented by Equation 2.14 where $\nabla f$ represents the gradient of image at a given point. Accordingly, highest gradient locations (edges) will attain minimum energy & force moving snake to stick with the edges. Other terms (line & corner) can be incorporated with suitable weights for a reasonable cost function that incline snake towards lines & corners. $E_{cons}$ represents external constraints by user for explicit control of snake movement.

This can be used to penalize the snake if it moves too away from the initial position, or into some undesired region. For many applications constraints are not imposed, means simply this is set equal to Zero & snake moves under influence of internal and external energy only. The optimization is achieved not by direct minimization of snake energy functional of Equation 2.11, but solving a numerical model based on Euler-Lagrange equations as reported in [6]. Equation 2.15 represents snake model used to find boundaries of an object inside image.

$$- \frac{d}{ds}\left(\alpha.\,||\,\frac{d.C(s)}{ds}\,||.^2\,\right) + \frac{d^2}{ds^2}\left(\beta.\,\left|\left|\frac{d2.C(s)}{ds2}\right|\right|.^2\,\right) + \nabla(Eext[C(s,t)] + Econ[C(s,t)]) = 0 \qquad (2.15)$$

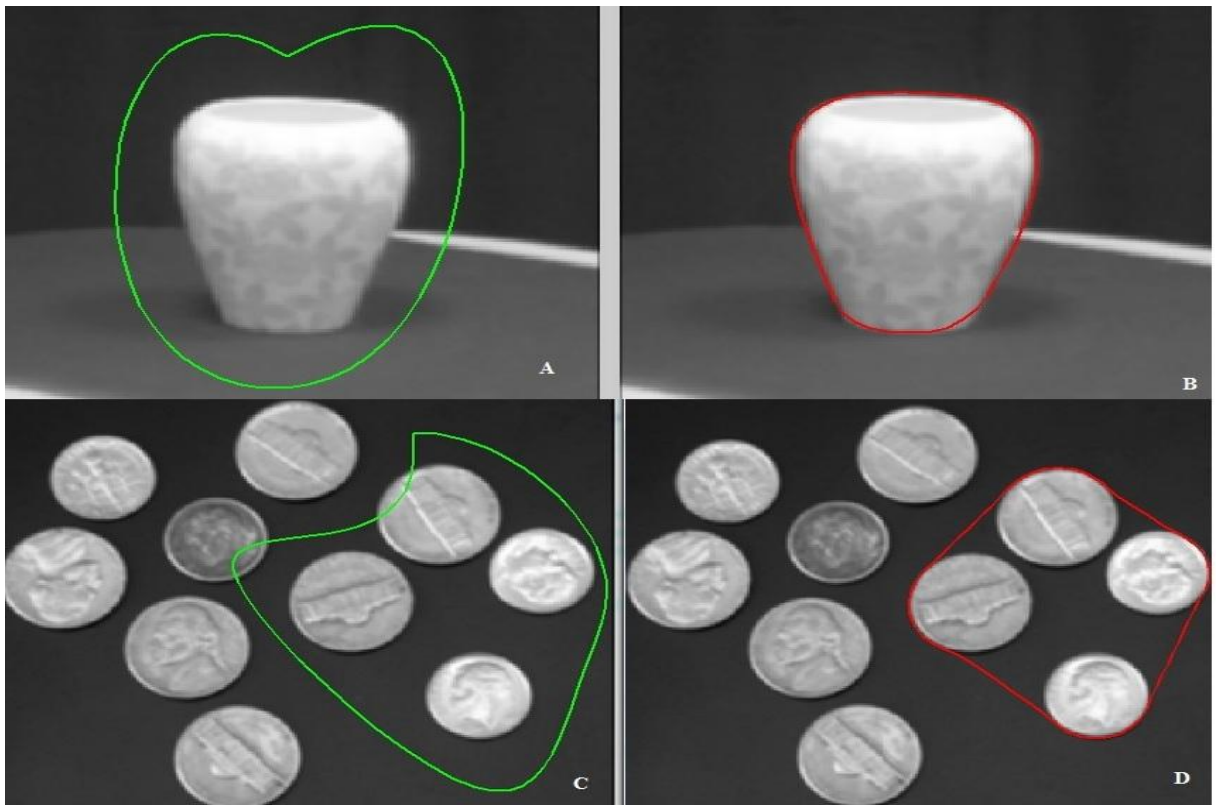

Figure 2.8.Segmentation based on active contours. (Green) Initial contours (Red) Final Contours. (B) Successful segmentation of object. (D) Snake unable to handle topological changes.

"Active Contour Snake" (parametric representation) of the curve deformation has been used for many years in image segmentation as it dircetly handles the control points of the curve resulting in fast segmentation. Practically active contour model is implemented in terms of "series of splines" where spatial & temporal derivatives are approximated by finite differences. Figure 2.8 represents the implementation and associated shortcomings of active contour model. Inspite of being intuitive method, certain limitation are also associated with snake model. Firstly, the convergence of snake is dependent upon initial estimation of curve. Snake will never be able to capture object accurately if initialization is done away from the object. Growing capability of the snakes is another addressable issue. If a snake stretches beyond a specific limit then extra control points are required for making precise estimation else boundaries may not be captured. Poor response to topological changes is another major drawback associated with active contour model.

## 2.5.2 Geometric Deformable Models

In the classical snake model, deformation is obtained via minimization of cost function that attains the optimal value at object boundaries. The idea behind this snake expansion is technique of deforming curves to reflect image features by energy minimization [7]. Unfortunately, the energy model used is not capable of handling the topological changes that becomes problematic when multiples objects are to be detected simultaneously. Another drawback is the non intrinsic behavior as energy depends on curve parameterization instead of relating object geometry & shape. In geometric models proposed by Casselles [8], Malladi et al [9] curve propagation is controlled by velocity containing two terms (first controlling the regularity & second responsible for expansion or shrinkage towards boundary). Casselles [10] reported geodesic active contour model that relates boundary detection problem with energy minimization. Derived curve evolution model represents geometric flow (PDE based on mean curvature motion) that is more close to curve evolution theory instead on energy minimization problem. Level set formulation for this curve shortening flow allows automatic detection of topological changes. A brief derivation of mathematical model for geodesic curve evolution is presented in this section, for detailed derivation readers are referred to [10].

According to definition, active contour snake moves under influence of two forces. First is driving force conceived from image that pushes snake towards edges & other salient regions in image. Second is the physical property of contour itself including elasticity & stiffness that ensures regularity of the curve. Starting with the closed curve represented by C(s), length functional can be defined by Equation 2.16.

$$L(t) = \int_0^1 \left| \frac{\partial c}{\partial s} \right| ds \qquad (2.16)$$

Length functional is differentiated to obtain curve shortening flow, as Euclidean curve shrinks as quickly as possible during evolution. Equation 2.17 represents curve flow where "N" represents unit inward Normal & "k" represents local curvature. For a detailed derivation of curve shortening flow, we refer readers to [89].

$$\frac{\partial C}{\partial t} = k.N \tag{2.17}$$

Because of the nature of contour (closed curve), continue evolution in unit normal direction will shrink to a point & eventually curve will vanish. To avoid elimination of the curve, an external inflation term is added in the curve evolution/shortening equation [2.17]. The inflation term will be used for growing curve & negative value assigned to inflation can wipeout the impact of curvature, as represented in Equation 2.18. (If inflation $v$ is set to negative, i.e. $v = -v$, then impact of curvature will be decreased by value of inflation term as $(k - v)$).

$$\frac{\partial C}{\partial t} = (k + v).N \tag{2.18}$$

Equation 2.18 shows that curve is moving based on snake properties explicitly. Images based information can be introduced by multiplication of an image based conformal factor g(x, y). Equation 2.16 is updated to represent both curve length & image information as given in Equation 2.19.

$$L(t) = \int_0^1 g(x,y)\left|\frac{\partial C}{\partial s}\right| ds \tag{2.19}$$

g(x, y) is termed as "conformal factor & it is derived from image I(x, y) representing useful information. Generally g(x, y) represents image behavior in terms of intensity information, expressed as gradient. Gradient is helpful to mark object boundaries where evolution should be stopped in segmentation. As shown in Equation 2.20, denominator term of g(x, y) contains additional value "1" to avoid numerical error possible because of divide by zero. Gradient is calculated by applying a Gaussian smoothing filter to image for noise suppression, as gradient based contour evolution is very sensitive to noise.

$$g(x,y) = \frac{1}{1+|\nabla G\sigma * I(x,y)|^2} \tag{2.20}$$

Accordingly curve shortening /evolution equation given by 2.17 is updated by introducing the image based information g(x, y) as shown in Equation 2.21

$$\frac{\partial C}{\partial t} = g.kN - \nabla g \tag{2.21}$$

By adding the inflation term to avoid curve shrinkage to because of normal curve shortening flow, curvature $k$ is replaced by $(k + v)$ as given in Equation 2.22. A complete derivation of the geodesic curve evolution can be found in work of Casselles *et al* [10].

$$\frac{\partial c}{\partial t} = g.(k + v)N - \nabla g \qquad (2.22)$$

Level set formulation [13, 14] defining curve flow for discrete implementation of geodesic active contour is defined by equation 2.23.

$$\frac{\partial \emptyset}{\partial t} = g.\left(div(\frac{\nabla \emptyset}{|\nabla \emptyset|}) + v\right)|\nabla \emptyset| + \nabla g.\nabla \emptyset \qquad (2.23)$$

Where $\emptyset$ represents the level set function (curve embedded into higher dimensional space), & $div(\frac{\nabla \emptyset}{|\nabla \emptyset|})$ denotes the curvature of the level set. For 2 Dimensional level set representation, curvature (kappa) is calculated using finite difference according to Equation 2.24

$$\kappa = \frac{\phi_{yy}\phi_x^2 + \varphi_{xx}\phi_y^2 - 2\phi_x\phi_y\phi_{xy}}{(\varphi_x^2 + \varphi_y^2)^{3/2}} \qquad (2.24)$$

Yang [11] reported that geodesic active contour model works efficiently for images where the region of interest is almost "rounded" & separated via strong gradients. For complex images that contain weak edges, efficiency of the geodesic flow decreases due to the nature of "conformal" factor. It does not forcefully stop curve at boundaries (conformal factor never reaches to zero because of additional 1 present in denominator). Another important problem highlighted in their work was the inflation parameter behavior in terms of uni-directional flow. This uni-directional flow demands that initialization must be completely inside or outside the object for a substantial segmentation as curve cannot move simultaneously in both directions. An adaptive inflation parameter was proposed by Yang that enables curve evolution in terms of shrinkage & expansion. This adaptive inflation in geodesic model emulated the behavior of traditional region based segmentation.

### 2.5.3 Level set formulation

Level set formulation proposed for front propagation by Osher & Sethian [13] has become a standard implementation framework for image segmentation. Limitations of traditional snake model are well addressed in level set method because of implicit representation of the curve. 2D curve to be evolved is represented as an iso-contour embedded into a higher dimensional space. Generally level set function is chosen as a signed-distance-function such that curve to be evolved is defined as zero level set. Numerical representation of the level set function is given by Equation 2.25.

$$\Gamma(x,t) = \{x | \phi(x,t) = 0\} \qquad (2.25)$$

Phi (∅) denotes signed distance function; x is spatial parameter describing the contour such that contour comprises of all pixels of signed distance function $\phi(x,t)$ equal to 0. $\phi(x,t) > 0$ represents the external region of curve whereas $< \phi(x,t)\, 0$ defines the interior of curve i.e. the object of interest.

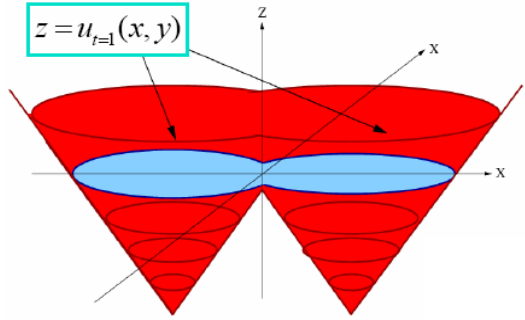

Figure 2.9: Illustration of the implicit contour representation.

For mathematical derivation of level set implementation, let us represent the velocity of a curve point by V(x) & entire surface (x) is to be moved with same velocity. This can be accomplished by solving the ordinary differential equation (2.18) for every point of the interface curve i.e. $\{x | \varphi(x,t) = 0\}$.

$$\frac{dx}{dt} = V(x) \qquad (2.26)$$

Practically it is not easy to solve this ordinary differential equation, since velocity field V(x) can lead to distortion of boundary elements of the surface. Besides, special procedures are required for maintaining smoothness & regularity of the curve to avoid elimination & intersections. Level set formulation can be used to overcome these problems. Evolution of the implicit function can be defined by Equation 2.27.

$$\phi_t + V \cdot \nabla\phi = 0 \qquad (2.27)$$

Where subscript "t" denotes temporal partial derivative with respect to time i.e.

$$\phi_t = \frac{\phi^t(x) - \phi^{t-1}(x)}{\Delta t} \qquad (2.28)$$

According to derivation of Casselles[10], the level set formulation for curve evolution in image can be specified by equation 2.29.

$$\phi_t = g(div(\frac{\nabla \phi}{\|\nabla \phi\|}) + \nu)\|\nabla \phi\| + \nabla g \cdot \nabla \phi \qquad (2.29) \approx (2.23)$$

In order to guarantee the numerical stability during surface propagation, the derivatives to the surface should be approximated by upwind schemes. Therefore Equation 2.29 can be presented as:

$$\begin{aligned}
\phi_{ijk}^{t+1}(x) = \phi_{ijk}^{t+1} + \Delta t[ & g_{ijk}k_{ijk}^{t}((D_{ijk}^{0x})^2 + (D_{ijk}^{0y})^2 + (D_{ijk}^{0z})^2)^{0.5} \\
& + (\max(v_{ijk}\phi_{ijk},0)\nabla^{+} + \min(v_{ijk}\phi_{ijk},0)\nabla^{-} \\
& + \max(v_{ijk}\phi_{ijk},0)D_{ijk}^{-x} + \min(v_{ijk}\phi_{ijk},0)D_{ijk}^{+x} \\
& + \max(v_{ijk}\phi_{ijk},0)D_{ijk}^{-y} + \min(v_{ijk}\phi_{ijk},0)D_{ijk}^{+y} \\
& + \max(v_{ijk}\phi_{ijk},0)D_{ijk}^{-z} + \min(v_{ijk}\phi_{ijk},0)D_{ijk}^{+z}]
\end{aligned} \qquad (2.30)$$

Where $D_{ijk}^{0x}$, $D_{ijk}^{0y}$ and $D_{ijk}^{0z}$ are the central difference that for the first order derivatives, and

$$\begin{aligned}
\nabla^{+} = (&\max(D_{ijk}^{-x},0)^2 + \min(D_{ijk}^{+x},0)^2 + \max(D_{ijk}^{-y},0)^2 + \min(D_{ijk}^{+y},0)^2 \\
& + \max(D_{ijk}^{-z},0)^2 + \min(D_{ijk}^{+z},0)^2)^{1/2},
\end{aligned} \qquad (2.31)$$

$$\begin{aligned}
\nabla^{-} = (&\max(D_{ijk}^{+x},0)^2 + \min(D_{ijk}^{-x},0)^2 + \max(D_{ijk}^{+y},0)^2 + \min(D_{ijk}^{-y},0)^2 \\
& + \max(D_{ijk}^{+z},0)^2 + \min(D_{ijk}^{-z},0)^2)^{1/2},
\end{aligned} \qquad (2.32)$$

Active contour deformation model has been used effectively for medical image segmentation as reported by several authors. Cerebral cortex segmentation in MR images is reported by Zeng *et al*[15]. Cardiac MR Segmentation is reported by Yezzi *et al* in [16] and successful bone delineation is reported by Leventon *et al* [17]. Main advantage of active contour based deformation is ability to ensure curve smoothness while localizing image features. A comprehensive review of deformable model utilization for image segmentation can be found in [18].

Two different representations of deformable models including parametric (classical snakes) & implicit (level set formulation of geodesic snakes) have their own advantages & disadvantages. For complex images, especially for medical data classical snake model fails to segment efficiently. The ability of level set formulation to address topological changes by accommodating split & merge operations make them an ideal choice for medical segmentation. The main shortcoming of level set formulation is inherent problem of computational cost, because of the fact that whole interface is evolved in every iteration. Computational burden is compensated by employing narrow band method to evaluate only restricted region around zero level set for new position of the curve. Narrow band approach exploits the fact that object edges do not show abrupt changes & iteration wise a certain region is to be inspected for new position of curve

Traditionally, deformation methods make use of "edge based" gradient information as curve evolution force with expectation to stop at object boundaries. However complex images contain multiple objects & usually borders between different objects are not very strong. As a result, gradient based evolution may not stop & results in leakage. For instance, conformal factor g(x, y) in curve shortening flow equation (2.23) is based on image gradient values. Consequently, weak gradients are not respected & curve evolution does not stop at weak boundaries as conformal factor never approaches to zero. Moreover, Gaussian smoothing in the conformal factor, generally widens the boundary of the objects. This can lead to under segmentation as curve evolution may stop before realistic edges of the objects because of gradient calculation. In contrast, region based image force can be used more effectively for precise segmentation as it is can resist noise while preserving weak edges. Under assumption of intensity homogeneity for gray scale images, Chan & Vese [11] proposed an active contour model based on regional intensity statistics for object segmentation. This method seeks a maximum separation of average intensity value inside & outside the object. If the input image is denoted by I(x, y), then the basic formula for region based active contour proposed by Can & Vese can be written as

$$F(c_1, c_2, C) = \mu \cdot length(C) + \lambda_1 \int_{inside(C)} [I(x) - c_1]^2 dx + \lambda_2 \int_{outside(C)} [I(x) - c_2]^2 dx \qquad (2.33)$$

Where C is the contour to be evolved, c1 & c2 are mean intensities calculated inside & outside of the contour C, $\lambda_1$ & $\lambda 2$ are constant parameters controlling the behavior curve. This model utilizes the regional intensity information of image & successfully delineates objects having weak gradients. Based on global information, it is capable of segmenting all objects irrespective of the initial curve position. This technique efficiently segments classical images but produces poor results for medical image data. For example, this method extracts all objects having similar intensity behavior in an image while objective is to isolate a particular anatomical structure for detailed analysis. Sometimes a combination of different techniques can be useful for extraction particular object. For example, Baillard *et al* [19] combined Bayesian classification method with active contour segmentation (using level set formulation) for isolation of brain structures from MRI data cloud. In their method, contour was derived by posterior probabilities to attract boundaries of gray matter. Another successful combination was used by Pichon *et al* [20] where pixel intensity statistics was combined with region growing method for calculating arrival time of propagating front.. Yezzi *et al* [82] and Xhu & Yuille [62] also proposed regional statistics based segmentation techniques applicable in particular cases.

## 2.6 Enhancement of curvilinear Structures.

Image segmentation process can be aided by performing useful pre-processing. In case of image containing multiple objects, shape prior can be incorporated to help contour identify the objects of interest quickly. This is identical to the classical snake concept which allows user to incorporate image constraints in terms of previous knowledge. However instead of incorporating this knowledge into total energy calculation, this can be used to efficiently threshold the non-concerned objects pixels at beginning. For detecting circular objects in an image, it is suitable to suppress all the voxels that violate the circular shape model & then level set based segmentation contour can detect the objects of interest precisely in very short time.

This idea can be very effective to suppress irrelevant structures in medical images where the objective is detection of a particular structure for clinical analysis. For instance, a radiologist is interested in arterial segmentation i.e. tubular objects isolation for diagnosis of vascular pathologies. Geometric shape features of tubular structures can be used to identify vascular objects in image for quick segmentation. Mathematically, localized feature analysis at a particular point can be performed by using Taylor series expansion. It expresses function value at a point $x_o$ in terms of summation of infinite terms given in Equation 2.34.

$$f(x) = f(a) + \frac{f'(a)}{1!}(x-a) + \frac{f''(a)}{2!}(x-a)^2 + \frac{f'''(a)}{3!}(x-a)^3 + \ldots \qquad (2.34)$$

This expansion model can be applied for feature analysis in 2D or 3D images. If image is represented by L, then 2nd order Taylor expansion at a point $(xo)$ can be represented by Equation 2.35

$$L(x_o + \delta x_o) \approx L(x_o, s) = \delta x_o^T \nabla_{o,s} + \delta x_0^T H_{o,s} \delta x_o \tag{2.35}$$

Where $\nabla_{o,s}$ represents gradient ($1^{st}$ order derivative information) & $H_{o,s}$ denotes Hessian ($2^{nd}$ order derivative information) of the image centered at position $x_o$. According to scale space theory, differentiation is obtained by convolving with derivative of D-dimensional Gaussian that ensures particular scale utilization. In-order to satisfy scale space constraint, differentiation operation is implemented as given in Equation 2.36 (i.e. convolve image with Gaussian of particular scale).

$$\frac{\partial}{\partial x} L(x,s) = s^\gamma L(x) * \frac{\partial}{\partial x} G(x,s) \tag{2.36}$$

Where D-Dimensional Gaussian function is defined as $G(x,s) = \dfrac{1}{\sqrt{(2\pi s^2)}^D} e^{\frac{-\|x\|^2}{2s^2}}$

Default value for parameter $\gamma$ is 1, and it is used as normalizing factor for scale estimation. By applying the derived equations, second order partial derivative for image L(x) can be obtained by using equation 2.37.

$$\delta x_0^T H_{o,s} \delta x_o = (\frac{\partial}{\partial \delta x_o})(\frac{\partial}{\partial \delta x_o}) L(x,s) \tag{2.37}$$

For 3D image, hessian matrix can be calculated by Equation 2.38, Where x=(x, y, z) represents the location of a voxel in the volumetric data set & each element is calculated by using Equation (2.37)

$$H(L(x)) = \begin{bmatrix} L_{xx}(x) \; L_{xy}(x) \; L_{xz}(x) \\ L_{yx}(x) \; L_{yy}(x) \; L_{yz}(x) \\ L_{zx}(x) \; L_{zy}(x) \; L_{zz}(x) \end{bmatrix} \tag{2.38}$$

Eigen value analysis of hessian matrix can reveal the local shape & geometric information of different objects present in the image. Relationship between different combinations of Eigen values ($|\lambda_1| < |\lambda_2| < |\lambda_3|$) and their corresponding 3D pattern is summarized in table 2.1.

Table 2.1 Different combinations of Eigen values representing shape information

| 2D | | 3D | | | Orientation pattern |
|---|---|---|---|---|---|
| $\lambda_1$ | $\lambda_2$ | $\lambda_1$ | $\lambda_2$ | $\lambda_3$ | |
| N | N | N | N | N | Noisy, no preferred direction |
| - | - | L | L | H- | Plate-like structure (bright) |
| - | - | L | L | H+ | Plate-like structure (dark) |
| L | H- | L | H- | H- | Tubular structure (bright) |
| L | H+ | L | H+ | H+ | Tubular structure (dark) |
| H- | H- | H- | H- | H- | Blob-like structure (bright) |
| H+ | H+ | H+ | H+ | H+ | Blob-like structure (dark) |

In medical images, especially CTA & MRI imaging modalities, background appears darker than the objects of interest that are usually bones, calcified depositions, blood filled vascular structures. Moreover, tubular required vessels are thin structures that refer to combination (4) of table1, where $\lambda 1$ is approximately zero & remaining two Eigen values are negative numbers with high magnitude.

$$\begin{cases} 0 \approx |\lambda 1| \ll |\lambda 2| \le |\lambda 3| \\ \quad \lambda 2 < 0, \lambda 3 < 0 \end{cases} \tag{2.39}$$

For all voxels that satisfy the required combination of Eigen values i.e. (L, H-, H-), Frangi *et al* [21] proposed vesselness response measure as defined by equation 2.40.

$$Vo(x,s) = \begin{cases} 0, & if\ \lambda 2\ or\ \lambda 3 > 0 \\ \left(1 - \exp\left(-\frac{R_A^2}{2\alpha^2}\right)\right) \exp\left(-\frac{R_B^2}{2\beta^2}\right)\left(1 - \exp\left(-\frac{S^2}{2c^2}\right)\right) & otherwise \end{cases} \tag{2.40}$$

Where

$$R_A = \frac{(Largest\ CrossSectionArea)/\pi}{(LargestAxisSemiLength)^2} = \frac{|\lambda_2|}{|\lambda_3|}$$

$$R_B = \frac{Volume/4\pi/3}{(LargestCrossSectionArea/\pi)^{3/2}} = \frac{\lambda_1}{\sqrt{\lambda_2\lambda_3}}$$

$$S = \sqrt{\sum_{1 \le i \le 3} \lambda_i^2}$$

Term $R_A$ is used to discriminate plate like structures from tubular vessels. Presence of tubular structure pushes $R_A$ close to Zero. $R_B$ differentiates Blob like structures from other shapes where as S servers a penalty term to suppress the background noise. $\alpha, \beta$ and $c$ are constants that controls the weights in overall vesselness measurement. No standardised way has been established for establishing values of these parameters, and application based user supplied combination is used normally. Response from individual scales is combined & strongest responses are selected to generate final vesselness filter according to equation 2.43.This multi scale approach has been used to respond to vessels of different sizes in image.

$$v_o(\gamma) = \max_{s_{\min} \le s \le s_{\max}} v_0(s, \gamma)$$

(2.41)

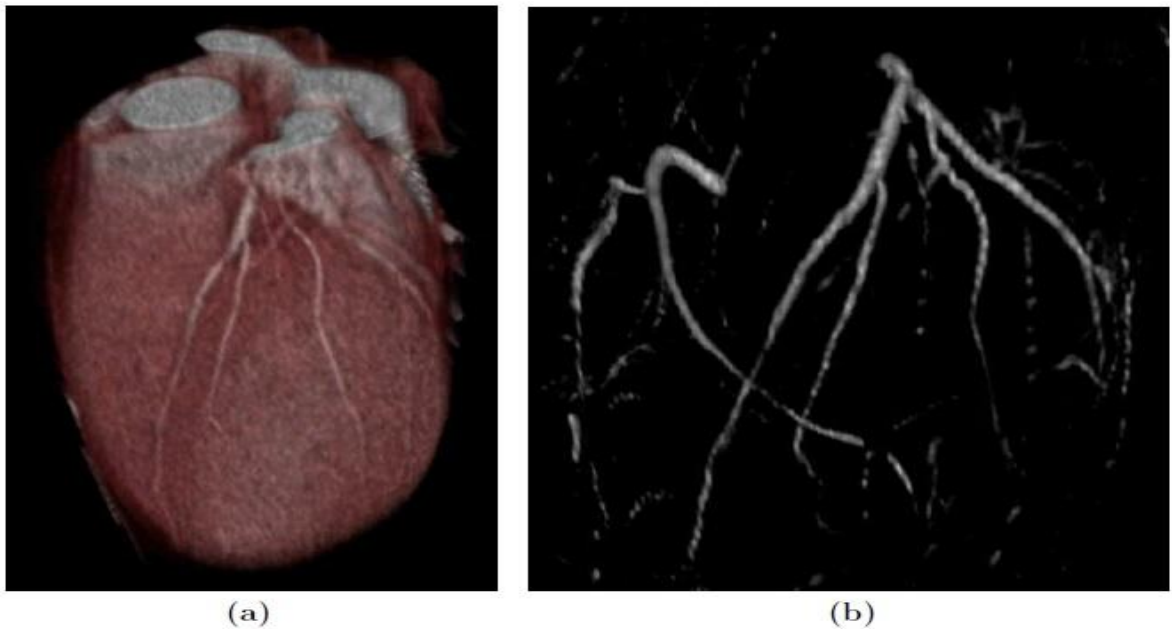

(a)                                        (b)

Figure 2.10 Vascular structure visualization.(a) volume rendering of CT image (b) curvilinear structures based on vessel multiscale enhancement

Figure 2.10 shows the advantage of applying multi scale filter for selection of desired tubular structures. Left side displays volume rendering of heart tissues whereas (b) on right displays volume rendering of vessel filter response.  As vesselness is measured at every voxel, the filter response is another volume of same size. Most of the heart muscle & tissues have been suppressed making shape of coronary arteries are visibly identifiable. Multi scale approach allowed extracting small & large vessels simultaneously. This thresholded volume can be used for precise segmentation using active contour model. Another approach used for enhancing blood vessels in 3D images is anisotropic diffusion method. This technique performs smoothing operation while the edge information is preserved. According to [22, 23] application on CT liver images suppresses image noise significantly whereas small vessel details were retained in comparison with application of traditional Gaussian filter.

# Chapter 3        Medical Image segmentation

With state of the art imaging equipment capable of recording sub millimeter details of internal body organs, segmentation algorithms have gained extraordinary concern of research community in recent years. Latest imaging modalities including CTA & MRA generates bulk amount of 3D data due to high spatial & temporal resolution. Manual analysis by a clinician is time consuming as well as interpretation depends upon the previous knowledge and expertise of the radiologist. Computer aided surgery, vascular abnormalities detection & quantification, tumor delineation and anatomical structures research are few examples demanding precise segmentation of anatomical structures. For instance, in cardiac CTA image appropriate segmentation of blood filled vessels is very important, since their geometrical features can reveal useful information regarding vessel remodeling and obstructed blood flow. Therefore fast & robust automated application for specific diagnosis task is the need of time.

By definition, image segmentation is defined as division of image domain into constituent regions that are substantial & easy to process. Generally intensity or texture homogeneity is used as dividing criteria that produces non overlapping regions with meaningful shapes [5, 24, 25]. If the image domain is represented by $\Omega$ then segmentation process produces sets $S_k \in \Omega$, whose union is the image domain & intersection is null set. Mathematically, the constituent (sets) must satisfy the equation

$$\Omega = \cup_{k=1}^{K} S_k \tag{3.1}$$

As constituents are non-overlapping, therefore $S_k \cap S_j = \{\emptyset\}\ for\ k \neq j$, and homogeneity represents that each region is connected. In case of image having multiple objects, segmentation algorithm aims to establish individual set $S_k$ for respective organ /object. For medical image data, pixel segmentation overtakes connected region segmentation. Pixel segmentation is flexible version of classical segmentation where connectivity constraint between regions is removed. In consequent segmentation, multiple disconnected regions actually belong to the same class of objects. Depending upon the complexity of image data, it is quite difficult to find total number of intended classes (sets) in pixel segmentation at start. Some prior knowledge can be employed to estimate the number of classes at start for efficient segmentation. For example, in cardiac CTA images, pixel classification can be done into three classes including lungs, heart tissues myocardium & blood filled regions [26]. Different segmentation mechanisms have been used for extraction of region of interest from medical images. Individual or combined application of clustering, Markov random field, artificial neural networks, PDE based deformable models, region growing and the simplest thresholding have been reported in literature. Comprehensive details regarding medical image segmentation algorithms can be found in [27, 28, 29]. In this chapter a brief review of CTA data analysis techniques is presented at start due to the fact that this work intends to use CTA data. In the following section vessel enhancement techniques specific to anatomical structures in medical images are discussed. Core problem of vessel segmentation is discussed in section 3.3 which will be consequently used in this research.

### 3.1    CTA Image Analysis Methods

3D image analysis is a complex problem and several techniques have been proposed for meaningful evaluation. Advance visualization techniques like contrast adjustment & window leveling are applied for visual emphasis of specific features. Tracking objects of interest is done in 3D data by establishing centreline/skeleton and surface behavioral analysis is performed using object segmentation. These techniques can be used as stand alone or in combination to explore 3D image behavior in substantial manner. CTA modality provides non invasive way of producing detailed 3D view of internal organs of human body but the amount of data generated is much more than what is required. Precise delineation of specific anatomical structure becomes challenging due to the intensity distribution in CTA. In DSA angiograms & MRA data, blood pool can be traced due to distinctive behavior in global histogram as blood voxels are assigned very low or high intensities in two methods. Direct visualization of DSA data is possible with maximum intensity projection (MIP) or volume rendering. In contrast, blood vessels in CTA data have intensity values that fall in middle of the global histogram (between intensity of lungs [very low HU] & bones [very high HU]). Consequently, application of MIP or direct volume rendering does not yield reasonable visualization in computed tomography images. For instance, MIP view of cardiac CTA is obstructed by bones & large blood filled regions (left ventricle of heart) resulting in hidden tubular vessels (region of interest for radiologist); however, a limited field of view based MIP can be used to visualize a local segment of the vessel. This leads to the conclusion that CTA data demands more processing for meaningful visual representation & segmentation of arterial structure in comparison of DSA or MRA data sets. Level set framework, deformable contours, Hessian based analysis, minimal cost path are some techniques often used for addressing CTA segmentation problem. Few examples from literature are presented here that reported accomplished precise segmentation specifically for CTA data.

Majority of the CTA segmentation studies in literature are related to carotid artery segmentation. In contrast to coronaries, carotid arteries are larger vessels & their static behavior leads to good image quality in CTA data set that makes processing easy. Antiga [30] proposed centreline detection mechanism for carotid arteries by using sparse field representation of geodesic active contour model. Andel *et al* [31] proposed a medial axis extraction method for carotid arteries based of gradient & second order information of image. Canny edge filter was used in combination with Hessian filter for design of cost function. In the following stage, minimal cost path search mechanism was used for extracting final skeleton of lumen. A semi automatic approach reported by Scherl et al [32] performs delineation based on Chan & Vese [11] energy model. Stenosis calculation efficiency of the model was evaluated by investigating cross sectional CTA images. Close approximation to manual expert markers was reported with mean error of 8%. Harnandez *et al* [33] addressed the challenging task of separating blood filled vessels from bones (both having high intensity values). He used a sequential combination of pixel segmentation & curve deformation. Pixels were classified into bone & vessel based on probability density function (achieved by using KNN) in the first stage. Consequently active contour deformation was applied for extraction of vessel boundaries. Similar hierarchical segmentation approach was reported in [34] where input CTA image was segmented into (bones & vessels) in the first stage. In the following stage, algorithm delineates blood filled vessels from bones successfully. Olabarriaga *et al* [35] reported successful delineation of abdominal aorta aneurysms. In this work double layer deformable model was proposed for capturing inside & outside vessel wall of aorta. As aorta is a large vessel, inner wall represents a notable contrast with blood pool whereas out wall was delineated based on nearest neighbor based probability, however this method is not suitable for segmentation of small tubular vessels like coronary arteries. Blondel *et al* [36] proposed a method for true reconstruction of 3D Coronary structures from 2D angiographic projection sequence but modern scanners are directly capable to generate 3D data representing true coronary behavior. Another method implemented on pig cardiac CTA cast, was proposed by Chen & Molloi [37]. They reconstructed vascular tree successfully by applying 3D thinning techniques & skeleton pruning methods. In [38], Szymczak used topological approach to track coronary artery. Proposed method was capable of locating distal segments; however validation of the complete coronary vessels was not reported by authors. Besides, it also required interactive aid from user for coronary tree identification. It is notable that very few resources exist in literature addressing the question of fully automatic coronary segmentation from CTA data. Almost all the existing method requires some extent of user interaction which makes them prone to the user expertise & prior knowledge.

## 3.2 Vessel Enhancement Review

Blood filled vessels resemble with cylindrical / tubular structures in 3D CTA data. For detecting abnormalities in vasculatures, a detailed diagnosis demands precisely segmented tubular structures. Several algorithms have been reported for enhancement of vascular structures from images & majority of them relies on second order Hessian matrix analysis. Vessel enhancement filter based on Hessian analysis responds at each pixel by calculating likelihood of such pixel to be vascular structures. For handling different vessel sizes, the enhancement response is calculated at different scales & pixel is assigned the strongest scale likelihood values. Seminal work for vascular enhancement was reported by Lorenz [39], Sato [40] and Frangi [21]. With small variations the core idea is analysis of Eigen value system to calculate vesselness measure at a particular pixel point (Eigen values represent geometrical shape information). However most Hessian matrix based vesselness methods reported in the literature share a common prejudice that they produce strong false response along the sharp edges in the image. These false responses imposes computational burden during segmentation process. In order to resolve the false edge problem, Bennink *et al.* [41] proposed a 'lineness' filter which is able to produce an intensity independent response to the center of the line structure while suppressing the unwanted response to the sharp edges. The proposed filter consists of three components, including gradient operator, Canny optimized second order lineness filter and the Gaussian derivatives operator, which enables the filter to respond only to the center of the vessels. Wu *et al.* [42] presented a hybrid filtering approach to detect the vascular structure from the image dataset. In their work, they proposed to combine the Hessian based filter with a matched filter capable to enhance the tubular structure while suppressing step like edges efficiently. The method was tested on retina angiography images. As illustrated in Fig. 3.1, the edge effect is significantly suppressed by the proposed algorithm.

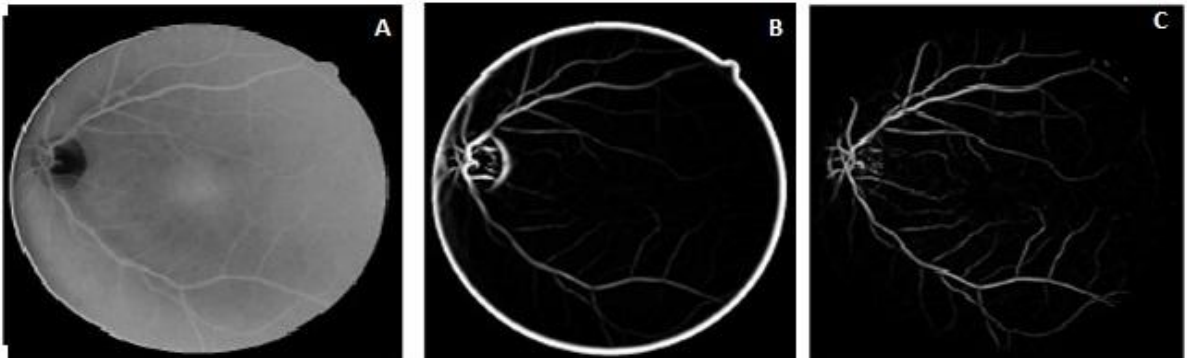

Figure 3.1: Vessel enhancement effects (A) The input image, (B) Frangi measure applied, (C) Edge suppression filter applied

Enhancement filter is mathematically modeled to complement a specific structure of interest. For instance vessel filter is designed to enhance the tubular structures & suppress blobs. Interestingly, bifurcation points in vessel exhibits a blob-like structure and traditional vesselness filter will suppress those particular pixels. Zhou *et al.* [43] proposed a new response function (likeliness measure) that can enhance both tubular & blob structures simultaneously. They used metric for vesselness proposed by Frangi with a slight modification. Additional constant "*c*" was introduced in the vesselness calculation which plays an important role in enhancement of both structures. Equation (3.2) represents vesselness measure proposed by Zhou.

$$
R(x) = \begin{cases} \dfrac{\lambda_1 + \lambda_2}{2} \exp(-\left| \dfrac{|\lambda_1|}{\sqrt{\lambda_1^2 + \lambda_2^2 + \lambda_3^2}} \right| - c) & \lambda_1, \lambda_2, \lambda_3 < 0 \\ 0 & others \end{cases}
$$

(3.2)

An appropriate selection of the constant *c* allows enhancement of two mentioned structures while suppressing background noise. However in context of the coronary artery segmentation, the amended filters can be avoided. Based on traditional Frangi vesselness measure, region based active contour deformation discards all non-vessel pixels efficiently.

### 3.3 3D Vessel Segmentation Review

Despite of active research in the last decade, 3D vascular segmentation remains a challenging task due to several reasons. In context of medical images, Intra patient variability is the most important factor. Even for a particular subject, vessels are generally surrounded by complex anatomical organs that make delineation much difficult. A number of techniques can be found in literature addressing 3D vascular segmentation including deformable models, minimal path cost analysis & levelset based deformation; however automatic frameworks are least reported because of the complexity of problem.

Boskamp *et al* [45] reported the use of region growing algorithm for vessel segmentation in CT and MR images. Initialized with one or more **manual seed points** within the vessel to be segmented, region growing segmentation was used. As illustrated in Fig. 3.1, neighboring voxels that satisfy the similarity criteria becomes part of vessel. To minimize the impact of nearby nonvascular objects, an optional 'pre-mask' procedure was used. The criteria used for region growing mechanism was traditional "intensity thresholding" that is sensitive to noise. This also leads to leakage at the point of weak gradients.

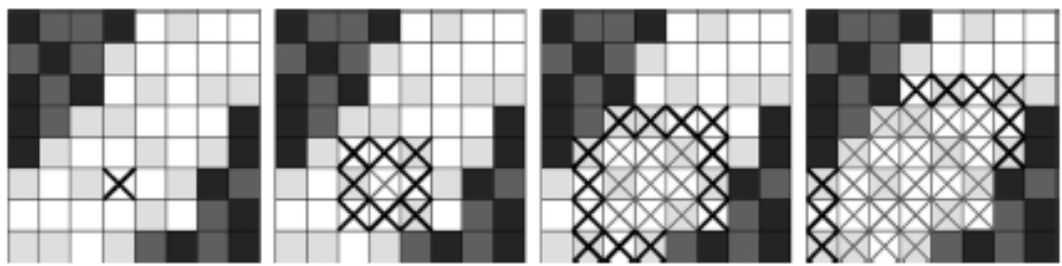

Figure 3.2: Region growing algorithm for vessel segmentation

Yi and Ra [46] proposed solution of leakage problem by localized region growing algorithm. Region growing was performed in a local cube where the size of the cube is determined dynamically with the help of estimated diameter. Similar scheme of controlled region growing was adopted by Tschirren *et al* [47] as they incorporated fuzzy connectivity criteria to minimize the leakage. Metz *et al.* [48] described a semi-automatic approach for tracking vessel centreline in CT data. The proposed region growing method was used and special measures were taken to handle bifurcation points. Consequently, their method successfully minimizes the leakage problem in segmentation that usually occurs at bifurcation points.

Deformable models have been used for effective vascular segmentation by many authors. Basic idea of contour fitting method is to model variations in basic features (either the shape or texture) of the objects so that curve deforms itself to detect the structures of interest in new images. Authors in [49] proposed to model the axis and cross-sectional radius variations independently. However, these methods were not able to fit the entire vascular structures since they only were capable to address simple bifurcations. Feng *et al.* [50] applied a two-phase modeling approach to segment the entire vascular structures from volumetric data. In the first stage, model deformed to fit the medial axis of the vessels & radius of vessel was estimated in second stage of deformation. By interpreting the tubular structures as the assembled cylindrical branches, their method segmented the vessel with multiple bifurcations. Florez *et al.* [51] presented a deformable cylindrical-simplex-model based algorithm to extract the vascular lumen form 3D MR images. Initially they applied a fast skeleton method to estimate the centreline of the vessels in a coarse resolution. In the second stage deformable model was initialized near the medial axis of the vessels. The model evolved in terms of the geometric constrains & image forces to capture the boundary of the vessels. Deformable model based tubular reconstruction approach was also reported by Yim *et el* [52]. In MRA images they used tubular coordinate system that controls surface deforming process; however, re-parameterization required to avoid self intersection makes this approach very complex. Worz *et al* [53] used famous cylindrical method for vessel segmentation. In order to capture vessels of different sizes, parametric intensity model was proposed in their work. Main limitation of this work is approximation of vessel cross section with circle shape, which is not always valid especially for abnormal vessels showing extent of vessel remodeling. Model based methods simplify the vessel extraction and representation problem by fitting the shape of the vessel to a certain geometric model. These can be fast & intuitive but the model usually have limitations in representing all possible shapes such as bifurcations & irregular cross sections, which is often the case for diseased vessels. The construction of such models remains a difficult task as it is quite difficult to obtain much required training data representing all possible variations.

A number of minimal cost path based studies proves this methodology as a valuable alternate for vessel segmentation. Implementation of this technique is reported in [54, 55] where authors proposed to extract vessels in 2D & 3D images respectively. Cost function was based on image gradient (intensity based information). Centreline extracted using wave front propagation and back tracking techniques respectively was not very accurate for aforementioned methods. By integrating the vessel size as $4^{th}$ dimension, Li and Yezzi [56] adjusted the centreline position in augmented space. The additional constraint leads to aligned & more accurate centre points. The intrinsic nature of minimal path technique that requires the specification of at least two points for path calculation makes it less automatic & initialization depends upon the user input. For instance, in case of left coronary artery, multiple seed points are required according to this approach for detection of complete arterial tree.

According to literature, levelset formulation (geometric active contour model) has been used frequently for delineating anatomical structures from medical images. Ability of level set formulation to handle topology changes makes them ideal choice for vascular segmentation as the vessel often exhibits complex topology. The effective utilization of geometric active contour to detect object boundaries in medical image can be found in [57], where authors reported a detailed experimentation of geometric contours for different imaging modalities. Wink *et al* [58] reported a simple vessel extraction approach based on 2D contour segmentation. Repetitive process of determining centre point followed with corresponding vessel boundary detection was performed to generate final segmented surface. Axial slices along z-axis were used in 2D segmentation where boundaries were based on gradient changes. After 3D surface construction, re-sampling was performed in 3D volume to obtain 2D orthogonal cross sectional slice in the direction of the vessel. The achieved segmentation was not very effective due to inherent limitations of gradient based approaches i.e. sensitivity to noise. Geodesic active contour curve evolution was also used by Antiga et al [30] for carotid arterial segmentation from CTA images. For efficient implementation, sparse field representation proposed by Whitaker [59] was used. Based on initial seed points, surface was allowed to grow like balloon as curve evolves iteratively. Artifacts in segmented surface due to collateral vessels were removed by applying smoothing operation. Chen & Amini [60] proposed a hybrid framework for quantification of 3D structures. Level set evolution was applied on vessel enhanced image for arterial segmentation, but unfortunately CTA data was not tested in this work.

Mostly geometric active contour models rely on image gradient information as the curve driving force. As a result contour may leak into adjacent structures at the ambiguous boundaries. To address this problem, Nain *et al.* [61] proposed to integrate shape prior into the geometric active contours framework. They proposed to apply a local shape filter representing geometric constraints as shown in figure 3.3. The shape filter was defined as a ball structure centered at each point along the contour to be evolved with radius "*r*". The filter calculates the percentage of the voxels that are both within the ball and the region inside the contour. Depending upon the current point location, the output becomes high or low value. High values signals that point lies inside widening region & vice versa.By combining the filter response with the level set formulation, the proposed method was able to penalize leaks during curve evolution [88].

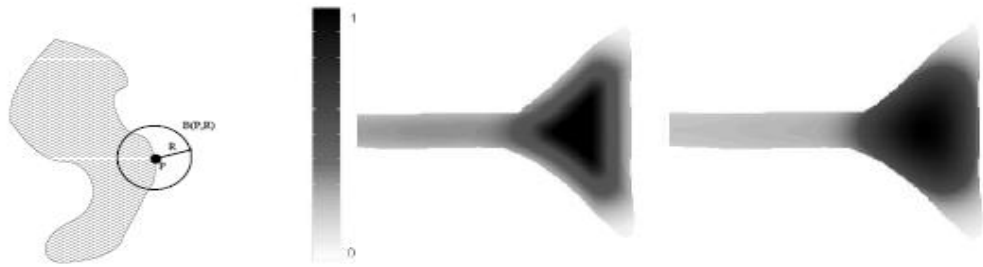

Figure 3.3: Application of shape filter to prevent segmentation leakage (a) Intersection of the shape filter & region R inside contour. (b) Response of shape filter for synthetic image

Yang *et al.* [62] described a hybrid level set approach to segment the coronary arteries in CTA images. They started with the pixel classification into three classes namely air, cardiac muscle and blood filled regions. In the following stage, posterior probabilities were calculated for each voxel of image that served as image force for driving contour. An adaptive inflation term was introduced that allows curve motion in multi directions. As illustrated in Fig. 3.4 the boundaries of vessels defined by the posterior probabilities are clearer than conventional gradients based approach.

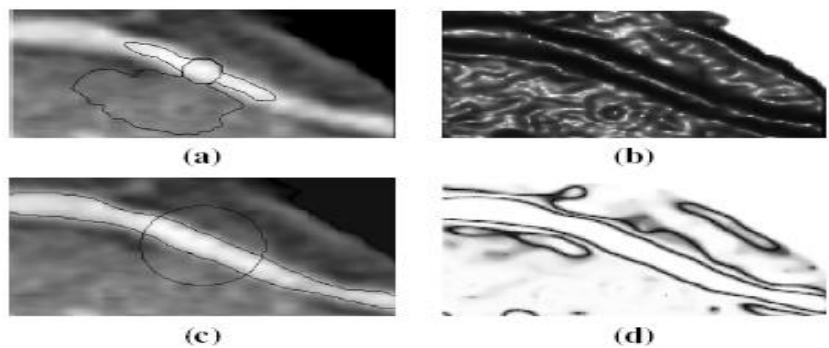

Figure 3.4: conventional vs posterior probability based active contour segmentation (a) Resultant vessel based on gradient energy, (b) Curve stopping term obtained by intensity gradient. (c) Resultant vessel based on posterior probability energy (d) Curve stopping term obtained by posterior probabilities.

Active contours model discussed so far are based on edge information, i.e., making use of the edge/gradient information as the stop criterion. The major disadvantage of edge based active contour is that the initial curve should be placed near the object boundaries as well as the segmentation quality is prone to the image noise. In recent years, region based active contour models have been widely used to resolve leakage problem at object boundaries. Rather than using the edge information as the stopping criterion, these methods attempt to model different regions by intensity statistics such that total energy is minimized at optimal region separation. Conventional region based active contour models reported in [11, 63, 64] fails to handle medical image data due to the intensity in-homogeneity problem. Straightforward application to medical images leads to erroneous segmentation due to global image behavior approximation. Li *et al.* [65] proposed to employ localized / regional intensity information in segmentation process for intensity inhomogeneous images. They extracted local intensity information from image on the basis of Gaussian kernel function that defines the scope of region. Main limitation of their work is extreme sensitivity to initialization mask that results in huge leakage due to small variation of mask. A more generalized & robust method for handling intensity inhomogeneity problem was proposed by Lancton in [66]. Authors illustrated localization of three different energy models for successful segmentation of tubular structures from 3D CTA data & the impact of localization kernel was fully evaluated.

# Chapter 4. State of the art on Soft Plaque Segmentation

Precise vessel tree is extracted from data cloud to detect pathological abnormalities. The geometric shape information of vessels can help in overall assessment of disease in terms of plaque burden & remodeling index. Major segments of coronary artery tree are generally inspected for lesion points where the blood flow is obstructed as this hindrance leads to severe consequences including angina, heart attack & myocardial infarction. The obstructions are categorized in two classes termed as calcified & non-calcified plaques (NCP). Calcified plaques are easy discernible due to their composition as they are reflected clearly in CTA image because of intensity properties. These plaques can be identified by clinicians easily, and depending upon the total plaque burden necessary measures are taken to restore the adequate blood flow. NCP results from a build-up of atherosclerotic deposits (lipid & fat deposition) and generally resides within the walls of coronary arteries. In contrast to calcified plaques, NCP composition makes their visibility very ambiguous and challenging. Isolating vulnerable plaques in CTA is a challenging problem because soft plaques generally have similar appearance to nearby blood and muscle tissues. Non-calcified plaques are potentially more dangerous in terms of clinical risk, as it is difficult to mark their existence before they results in fatal consequences. Therefore segmentation of vulnerable plaques is essential given that non-calcified plaques are much more likely than calcified plaques to rupture and cause a variety of acute coronary syndromes [90].

The main limitation of traditional diagnosis methods is the associated inability to provide information about vessel-walls during visualization [67]. Consequently there is ultimate need of advance imaging algorithms for early identification and shadowing of NCP lesions in patients. Recent developments in medical imaging including magnetic resonance imaging (MRI) & multi-detector CT (MDCT) has emerged as promising tool for description of atherosclerotic plaques, both in terms of shape quantification and composition. As calcified plaques are easily separable in CTA, numerous methods have been proposed to automatically detect calcified plaques with stable accuracy [68, 69]. In context of the main focus of this research, a detailed review of state of the art NCP plaque segmentation methods is presented in this chapter. It is notable that frequent technical literature is not available addressing the problem of vulnerable plaque detection. However, several case studies are available on the internet but these represent controlled environment experimentation & used for medical diagnosis in particular medical centers rather than general computerized framework design for detection & quantification.

## 4.1 Coronary Artery Extraction & Analysis for Detection of Soft Plaques in MDCT Images.

The focus of this study [81] was detection of non calcified plaques in coronary arteries. Two CTA data sets were investigated for the plaques existence. The application of the proposed technique identified the plaque location correctly in the coronary arteries & visual results were produced with statistical measures/graph data. In the first step of two fold methodology, authors obtained the centerline of the vessel in 3D volume by using technique proposed in [71] where vessel is tracked by local Eigen values of the Hessian matrix. To ensure the minimal impact of image local features on centerline extraction process, pre-processing was applied to isolates myocardial cavities & calcified plaques etc. In pre-processing stage associated voxels were regularized by assigning low intensity value. After centerline generation statistical modeling was used for lumen & arterial-wall segmentation. Gaussian mixture model was used to represent vessel & its surrounding tissues initially followed with application of Expectation Maximization algorithm for optimized probability map. To handle the variations in the intensity values along the vessel (usually occurs in medical images), a cylindrical model based on the local neighborhood of centerline point was used with radius value equal to 10mm. This model extracts segment between two consecutive points of the vessel along with surrounding myocardium tissue. Extracted cylinder was modeled by three class Gaussian mixture model to obtain distribution parameters ($\mu$, $\sigma$) for three classes namely lumen, wall and Myocardium. After building probability map for three classes, lumen and vessel wall voxels were identified for each segment by investigating the (high) probability for every class.

The existence of plaque was perceived by investigating geometric features of the vessel. The luminal narrowing was measured by calculating cross section area of lumen & wall represented by ($A_L$) & ($A_w$) respectively. Area between two consecutive points P(i) & P(i+1) was calculated as ratio of volume to length, i.e. Cross section Area=Volume between points / Length between consecutive points. By comparing two area measures $A_{(L)}$ and $A_{(w)}$ , A metric was obtained that signals the presence of either Calcified or non-calcified plaque as shown below in the Figure 4.1.

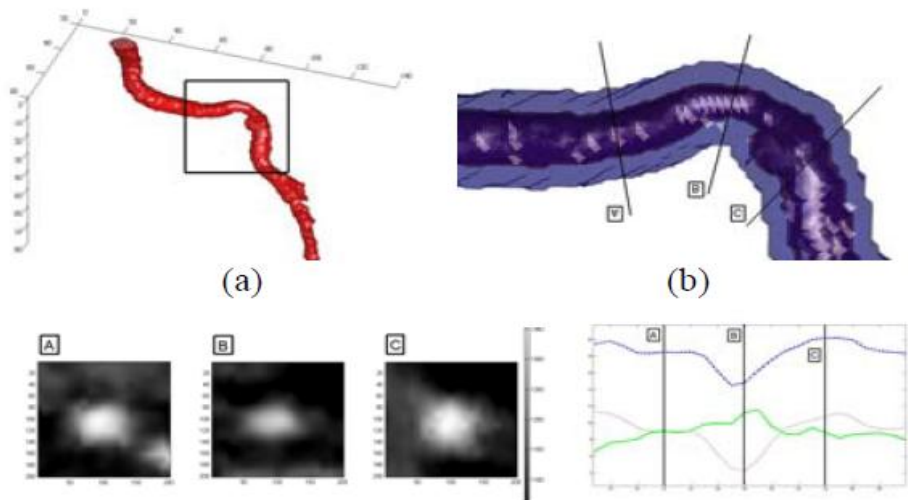

Figure 4.1: Geometrical analysis for soft plaque detection

A computationally efficient procedure developed for detecting soft plaques in CTA but no clinical validation of the results was discussed in the paper. Although author claims to detect the soft plaques but no detailed quantitative analysis of the detected plaques was reported in paper. Another limitation of this method associated bulk pre-and post-processing that makes this approach user knowledge dependent.

## 4.2    Soft Plaque Detection and Automatic Vessel Segmentation

This work [90] aimed to detect vulnerable lesions in coronary arteries. 8 CTA data sets were evaluated and success rate of 88% was reported. Detected plaque locations were validated by expert clinicians as reported by the authors. Twofold methodology was used by authors starting with region based coronary tree segmentation whereas plaques were identified in subsequent stage. Based on single framework, "simultaneous segmentations" was the fundamental notion of this novel work. In the first stage arterial tree was segmented from volumetric data based on universal modelling energy (Chan-Vese energy) as the driving force of evolving contour in region based segmentation. In the successive step two surfaces (interior & exterior contours) were evolved simultaneously based on mean separation energy. Overlapping extent of two contours was used as metric for soft plaque detection. All locations where two evolving contours did not override were identified as regions with **non-calcified plaques**. Detection process is illustrated in figure 4.2

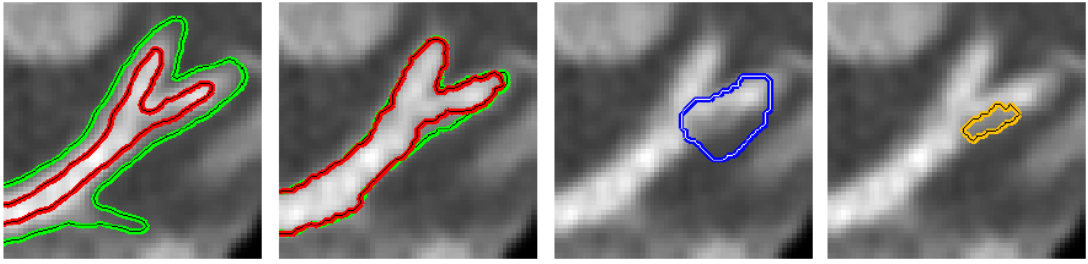

Fig.4.2 Plaque detection Results on CTA imagery.(a) Initial surface (b) Evolution result. (c) Expert marking. (d) Detected plaque

In contrast to approach of Renard [81], proposed method do not required any pre or post processing. The level set formulation for curve evolution allowed to capture all branching & bifurcation points successfully. Mathematical modeling used in this method is briefly reviewed here as it will be used in our subsequent research.

Segmentation of the arterial tree was achieved using deformable active contour model by posing it as an energy minimization problem represented by Equation 4.1.

$$\int_{\Omega x} \delta\emptyset(x) \int_{\Omega y} \beta(x,y).F(I,\emptyset,x,y)dy + \lambda\emptyset(x)|\nabla\emptyset(x)|dx \qquad (4.1)$$

Where F represents the **driving force** of the active contour and it can be selected according to the application. For vessel segmentation in the CTA volume, universal modelling energy based on Chan-Vese model was used. Mathematical representation for universal modelling energy is given in equation 4.2.

$$Fum = H\emptyset(y).(I(y)-\mu in(x))^2 + (1-H\emptyset(y)).(I(y)-\mu out(x))^2 \qquad (4.2)$$

Localization was used during segmentation as it accommodates inhomogenity caused by the fluctuating intensity values along the length of vessel. By substituting the driving force into energy functional, the segmentation curve evolution was defined according to equation 4.3.

$$\frac{d\emptyset(x)}{dt} = \delta\emptyset(x).\int_{\Omega y} \beta(x,y).\delta\emptyset(y).((I(y)-\mu in(x))^2 - (I(y)-\mu out(x))^2)dy + \alpha.div(\frac{\nabla\emptyset(x)}{|\nabla\emptyset(x)|})|\nabla\emptyset(x)|$$

$$(4.3)$$

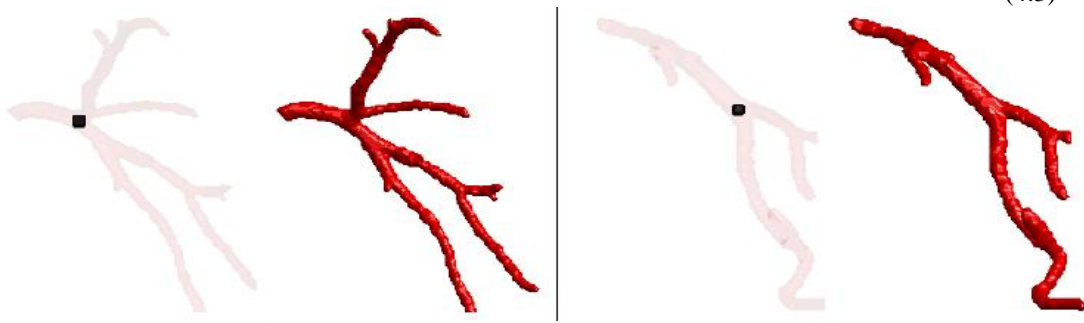

Fig.4.3Vessel Segmentation achieved using universal modeling energy

Arterial Segmentation was followed by plaque detection where two explicit surfaces were initialized using morphological operations. These surfaces were initialized inside & outside the original segmentation so that non-calcified plaques that reside within the wall could be located between two surfaces. These explicitly generated contours were evolved based on Means Separation energy [17] that pulls two contours towards each other. By substituting the driving force into energy functional, curve evolution was defined by equation 4.4.

$$\frac{d\emptyset(x)}{dt} = \delta\emptyset(x). \int\limits_{\Omega y} \beta(x,y). \delta\emptyset(y). (\frac{(I(y) - \mu out(x))^2}{Aout(x)} - \frac{(I(y) - \mu in(x))^2}{Ain(x)})dy + \alpha. div(\frac{\nabla\emptyset(x)}{|\nabla\emptyset(x)|}) |\nabla\emptyset(x)|$$

$$(4.4)$$

As illustrated in figure 4.4, initially the local interior region of inside surface contains only the bright voxels (red). As the contour is allowed to deform, it expands to capture more voxels containing blood but does not expand into a bit darker (changed HU values) soft plaque voxels. Similarly external contour initially contains initially the Myocardium voxels (green), and it does not contract to accommodate the soft plaque voxels from the boundary. All NCPs were isolated between two contours as neither moved into plaque voxels when driven by localized Means separation energy. In case of absence of soft plaque (no inhomogeneity in intensity values) these two evolving contours met on the vessel wall.

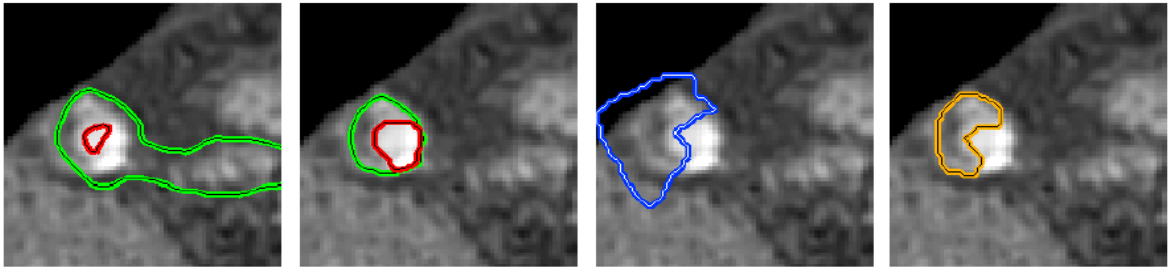

Fig.4.4 Plaque segmentation achieved using Mean Separation energy (a) Initial surfaces. (b) Result of evolution. (c) Expert marking (d) Detected plaques

## 4.3    Automatic Soft Plaque Detection From CTA

The focus of this work [91] was the detection of soft plaques in coronary arteries. 4 CT data sets were investigated for presence of soft plaques and the results were validated by a qualified cardiac expert manually. Like earlier methods twofold methodology was adopted in this work comprising of segmentation followed with plaque detection. Interestingly two different methods were evaluated for plaque detection including water shed & energy minimization segmentation. According to results water shed segmentation method has limited applicability & only captures stable plaques which develops a notable concavity in blood lumen. Majority of the unstable plaques were missed by this method. Detection scheme that employs energy minimization based deformation showed improved efficiency & detected plaque results were validated by manual endorsement of an expert radiologist.

In context of plaque detection, all blobs (heart chambers filled with blood and lung regions filled with air) were regularized by assigning intensity values of cardiac muscle in preprocessing step. This intensity allocation guarantees the minimal impact of non-vascular structures during segmentation. For arterial tree extraction, vesselness was obtained based on Eigen values of Hessian matrix as proposed by Frangi metrics. Initial probability map based on vesselness measure was generated for Bayesian network & then expectation maximization (EM) algorithm was used for construction of optimized probability map. In the segmentation stage edge based active contour evolution was used for detecting vessel boundaries. Traditional edge based cure evolution relies on gradient information however in this method curve was driven by Bayesian internal energy i.e. optimized probability map. After segmentation of the coronary arteries, surface dilation was performed to ensure that any possible plaques in the walls must be encompassed.

Plaque detection was accomplished by employing a hybrid scheme that combines geodesic model with region based active contours. Simply authors used geodesic model (curvature flow) but the energy used for curve shortening was derived from voxel intensity statistics of neighborhood (region based energy) around the contour points. The Energy function used for hybrid active contour (similar to conventional geodesic active contour) is defined by equation (4.5) as proposed by Casselles in [10].

$$\oint_{C(s)} g(I,s) ds \qquad (4.5)$$

Novelty of this method lies in the interpretation of the conformal factor g(I,s), as it depends upon the intensity values of voxels in localized neighborhood around the contour instead of traditional gradient values. Lancton derived the gradient descent equation for hybrid segmentation in [72] according to which curve shortening speed is defined by equation (4.6).

$$\frac{dC}{dt} = g(I,s)k.N + T1(I,s) \qquad (4.6)$$

In this work, authors used famous region based energy functional proposed by Chan-Vese (also termed as universal modeling energy) defined by equation (4.7).

$$g(I,s) = \int_{x \in w(s)} (I(x) - \mu in(s))^2 dx + \int_{x \in w'(s)} (I(x) - \mu out(s))^2 dx \qquad (4.7)$$

By replacing the driving image force $T_1(I, s)$ of equation 4.6, with the corresponding uniform modeling energy from equation(4.7), curve evolution equation was defined by 4.8.

$$\frac{\partial C(s)}{\partial t} = \left\{ \int_{x \in w(s)} (I(x) - \mu in(s))^2 dx + \int_{x \in w'(s)} (I(x) - \mu out(s))^2 dx \right\} k.N - \left\{ \oint_{C(r) \cap Br(C(s))} (\mu in(r) - \right.$$

$$\mu out(r)2(I) - \mu in r - \mu out r dr N \qquad (4.8)$$

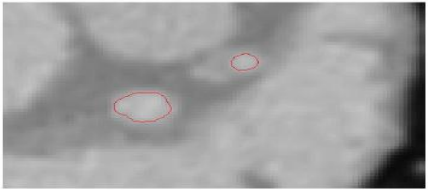

**Figure 23:** Dataset1: A 2D Slice of the dataset

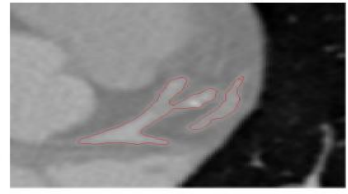

**Figure 26:** Dataset2: A 2D Slice of the dataset

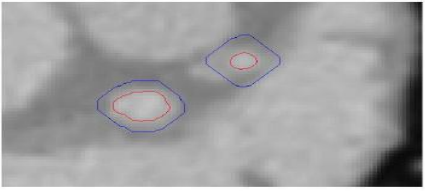

**Figure 24:** Dataset1: Initial Contour for Hybrid scheme

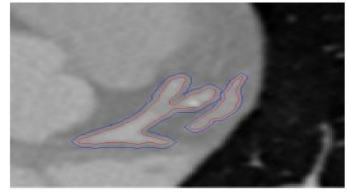

**Figure 27:** Dataset2: Initial Contour for Hybrid scheme

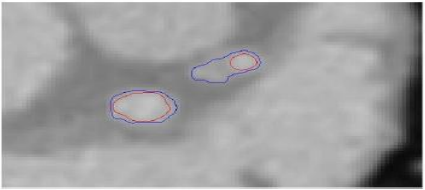

**Figure 25:** Dataset1: Final Contour after evolution

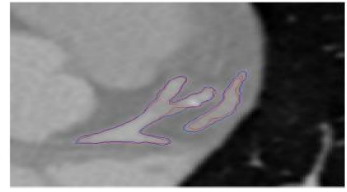

**Figure 28:** Dataset2: Final Contour after evolution

Fig.4.5 Plaques isolation achieved after hybrid segmentation. (Left) Dataset 1. Right(Dataset2)

## 4.4 Computerized Detection of Non-Calcified Plaques In Coronary CT Angiography: Evolution of Topological Soft Gradient Pre-Screening Method & Luminal Analysis

Main focus of this work [92] was detection of the soft plaques from CTA coronary vessels. CTA data of 83 patients was investigated that contained a total of 120 soft plaques. A dedicated pre-screening algorithm was developed to minimize the false positives. Accordingly authors reported a sensitivity of 92.5%. In this multi stage work, segmented arteries were investigated for soft plaques through a sequential geometrical analysis. According to the results reported, prescreening filter proposed by the authors significantly reduces false positive rate in plaque detection. A brief layout of the proposed approach is presented here.

Total 83 CTA data sets were investigated in this study. Initial segmentation of arterial tress (shown in figure 6) was achieved by using algorithm MSCAR-RBG[1] proposed by Zhou [73]. The used algorithms extracted about 86% correct arteries when compared with standard 17 segment coronary artery model so the incorrect arterial branches were eliminated/ inserted interactively, to ensure that accurate coronary arteries are to be passed to plaque detection phase.

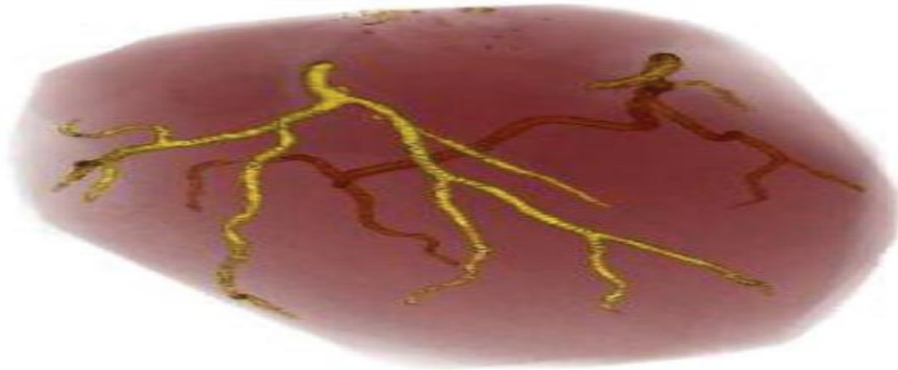

Figure 4.6. Extracted Left & Right Arterial tree

By applying curve planer reformation, different branches of the arterial tree were transformed into straightened volume as it allows detailed analysis including diameter variations, wall behavior & surrounding tissues. [81 * 81] rectangular 2D cross sectional planes were extracted across the length of the vessel centreline. Medial axis points control the orthogonal slice extraction process from 3D volume as shown in figure 4.7.

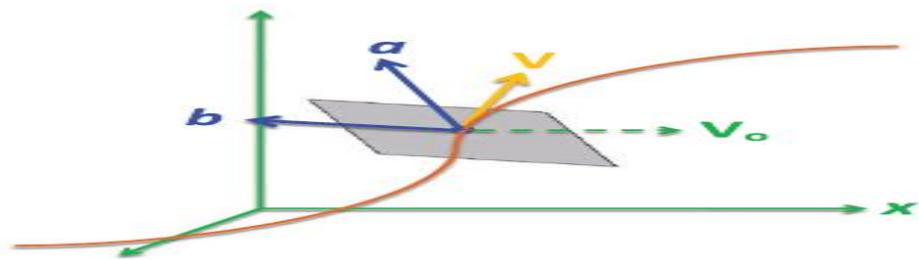

Figure 4.7.Vessel centerline controls the Cross sections to be used in CPR

Anisotropic diffusion was applied to minimize the impact of noise inducted because of motion & numerical re-sampling in CPR volume. After filtration localized cylindrical analysis was performed for examining vessel wall & lumen details. At every point of centreline, horizontal intensity gradient from the centre to the vessel periphery was calculated in slice using ray casting. Location of the maximum radial gradient was treated as vessel wall (It was assumed that all the voxels inside the vessel have similar values giving a small gradient inside lumen). Radius (distance between centre & wall) for every point of the centerline was obtained that defines the vessel wall in terms of 'Radius Profile'. A novel approach named topological soft gradient (TSG) was proposed for pre-screening of NCP candidates along the vessel centerline. Figure 4.8 represents the schematic layout for TSG method.

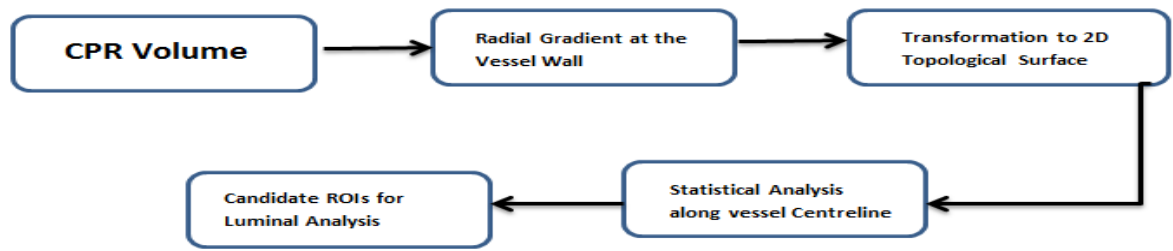

Figure 4.8 showing the TSG pre-screening method

For TSG screening, gradient in the radial direction from vessel centerline to the outward wall is defined as:

**(Average CT value at half radius from vessel centre to the vessel wall) --- ( Average CT value at half radius from vessel wall to outwards).**

After obtaining the radial gradient at all locations of wall, 2-D surface characterizing the radial gradient field of vessel wall was constructed. This radial gradient field was treated as 2D image and evaluated to identify the regions having soft gradients. A running window of 1.5mm centered at each voxel of the centerline was used to map corresponding values from gradient field.

Histogram was generated for mapped values & upper boundary of lowest quartile was selected as soft gradient value for that particular centerline point. Successively obtained soft gradient values along the vessel centerline forms the soft gradient profile for the vessel. The soft gradient profile was traversed for local minima and every local minimum was labeled as NCP candidate and a 2mm vessel segment centered at the candidate voxel was defined as ROI for luminal analysis. Plaque related voxels were detected from NCP candidate voxels via quantitative analysis. In quantitative analysis, geometric features and gray level characteristics were weighed as geometric features corroborate the shape information & gray level values confirm the voxel intensity information. Intensity value statistics were obtained from CPR whereas for geometric features, two additional transformations were applied on the CPR volume namely VSI & GDM.

1. Volumetric shape Indexing (VSI) to capture intuitive notion of local shape of surface.
2. Gradient direction mapping (GDM) to characterize the local direction of gradient vector.

Four measures including mean, standard deviation and skewness were calculated for each voxel in all three transformed volumes. One geometric measure termed as radius differential was obtained by calculating the first derivative of the radius along the vessel. In total, 13 statistical measures were used for detecting existence of the soft plaques in the coronary arteries. According to the authors, reported sensitivity of this TSG based method was 92.5%. Figure 4.9 shows the detected plaque by applying the proposed method.

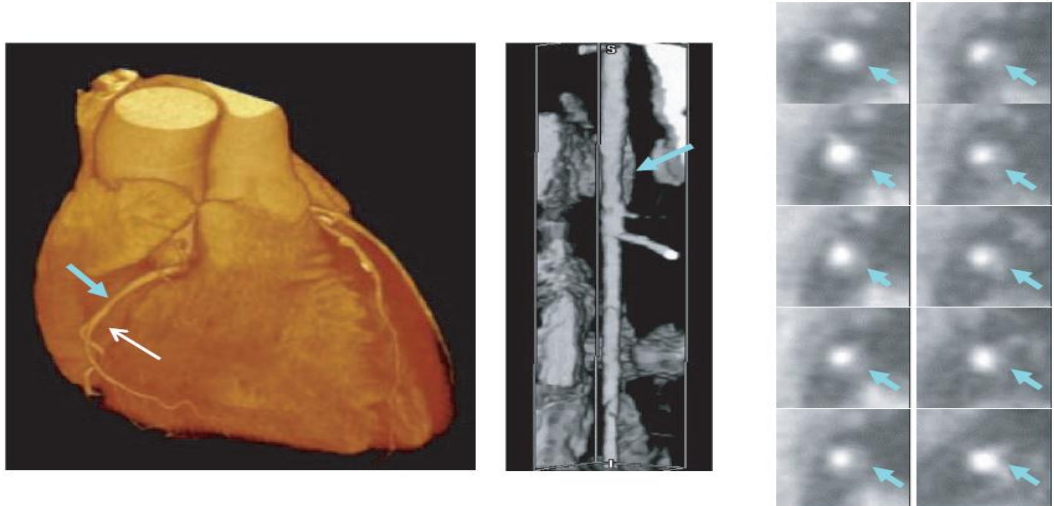

Figure 4.9 Plaque identified by applying TSG pre-screening method

## 4.5 Automatic Transfer Function Specification for Visual Emphasis Coronary Artery Plaque

Direct volume rendering (DVR) represents vascular structures more realistically [74], and automated transfer function can aid accurate interpretation. Generally soft plaques reside in vessel walls & HU value difference is insufficient to discriminate them from blood or cardiac muscles. Transfer function(s) were developed in this work [93] with the aim of facilitating clinicians to identify and track the vulnerable plaque present in the arteries. The main emphasis of this work was to improve the unique appearance of the clogged area for better visual analysis. Novelty in this work was TF based mapping of CT values to color & opacity that ensures different color coding for every dataset. This incorporated the inconsistent diffusion of contrast agent in different patients. In contrast to traditional methods of highlighting the vessel lumen, vessel wall was focused in this work. Total 63 CTA datasets were evaluated in this work & detected soft plaques were in correspondence with expert's manual identification.

Coronary arteries constitute approximately 2.5 % of the total CTA volumetric data, so global histogram does not represent the vascular behavior & pathological changes. Therefore segmentation was performed in first step to focus on the region of interest. In this study, coronary arteries were delineated using method proposed in [75] & the under/over segmented coronaries were adjusted manually under the guidance of expert. Local histogram analysis was applied to approximate the blood intensity distribution ($\mu$, $\sigma$) in the segmented arterial tree. Blood intensity values were estimated with Gaussian distribution & an optimal fit to the intensity distribution to the local histogram was calculated (using least squares). Blood Intensity parameters ($\mu$, $\sigma$) were obtained as ($\mu_{blood} = 356 \pm 136$) & ($\sigma_{blood} = 46 \pm 16$). This indicated that the average intensity varies strongly for different data sets, so setting a static threshold for hard plaques applicable to all data sets [76] is not realistic. Threshold value (350HU) defined by Agatston [77] shows an over estimation in the hard plaque separation. Accordingly authors defined a new threshold for separating hard plaques as defined by Equation (4.10).

$$T = \mu_{blood} + 3\sigma_{blood} \qquad (4.10)$$

Vessel branches were analyzed individually with the help of arterial centreline. Intensity profile volume (IPV) was generated for every branch by processing centreline voxels. For each centreline point, (n) rays perpendicular to the centreline were casted in outwards directions. These rays were sampled in dataset up to radius (3mm to ensure that whole arterial cross section is covered, since 2.5 mm is maximum radius of coronary arteries). The sampled intensity values were stored in a slice of IPV & this process was repeated for all centreline voxels of the branch.

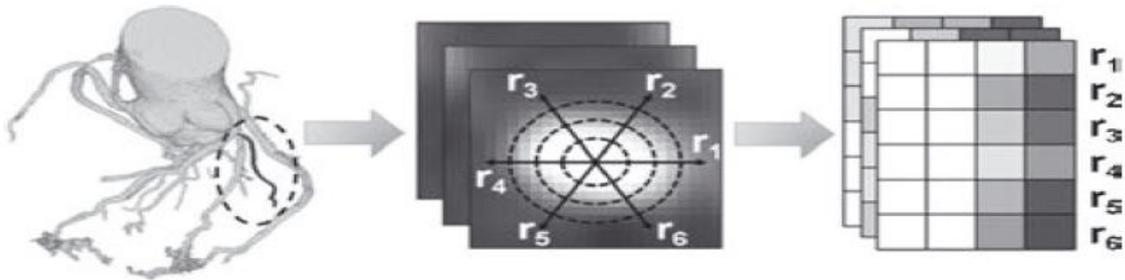

Figure 4.10. Intensity profile volume generation for a vessel branch

After building IPV, vessel wall intensities were detected. Vessel wall intensity was expected to be a vertical structure since all the values at a particular distance makes vertical line. A slice wise search mechanism was employed for locating vertical structures in IPV however these vertical structures are sometimes distorted because of artery remodeling which can be improved by applying Laplacian of Gaussian filter. So wall intensity distribution parameters ($\mu$, $\sigma$) were obtained for every branch as shown in figure 4.10. Finally, the longest centreline branch satisfying the condition ($\mu_{wall}$) < ($\mu_{blood}$ - $2\sigma_{blood}$) was extracted as best candidate for global vessel wall approximation.

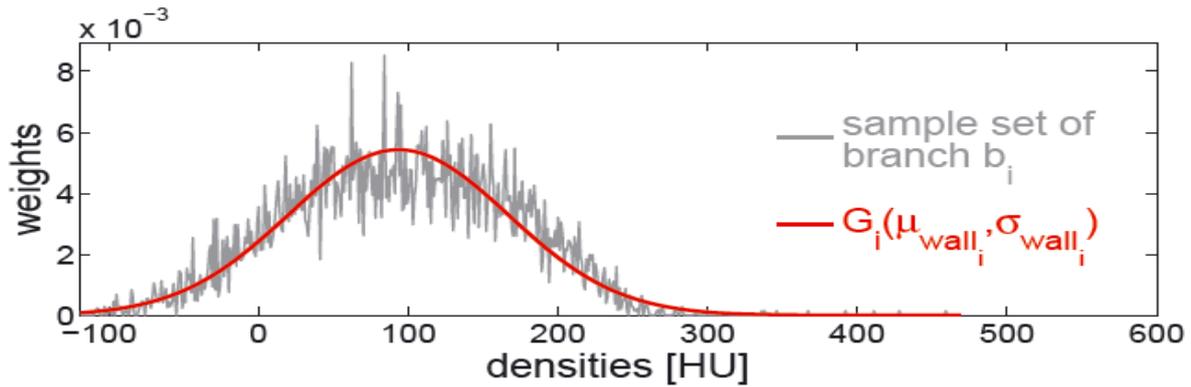

Figure 4.11. Estimation of Vessel Wall intensity value distribution

Transfer functions emphasizing the visualization of pathological changes were based upon supporting points which requires approximation of vessel wall intensity & blood intensity i.e ($\mu_{wall}$, $\sigma_{wall}$, $\mu_{blood}$, $\sigma_{blood}$). Supporting points were related with different opacities and colors and intermediate values were linearly interpolated. Color association to the supporting points targets high contrasts for the vessel wall visualization. For the vessel wall, a color scale from blue over red to green was applied, yielding high contrasts for the visualization of different vessel wall intensities & plaque deposits.

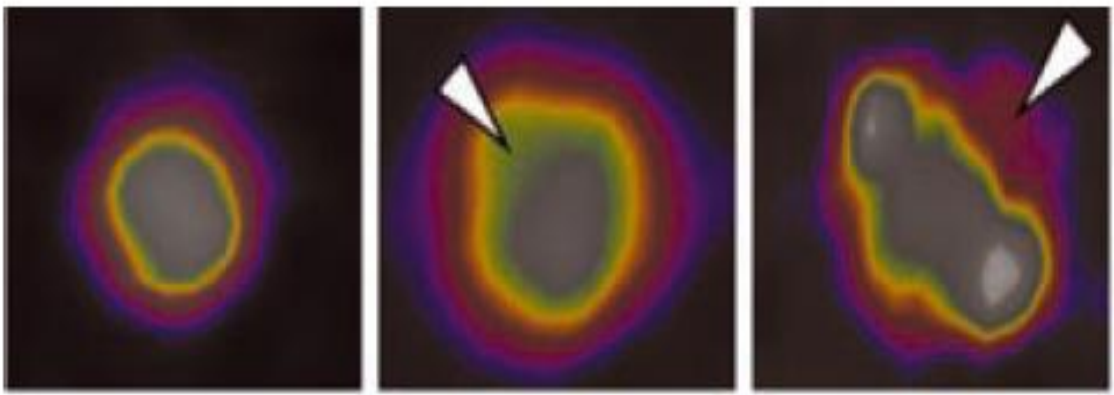

Figure 4.12. Visualisation enhancing the pathological changes in vessel.
(Left) without plaques (Middle) greenish color denser structures (Right) Pinkish color soft plaque

## 4.6    Automatic Detection of Calcified Coronary Plaques in Computer Tomography Datasets

The main focus of this work [94] was to design an automated framework for detection of calcified coronary plaques in CT images. In contrast to the avant-garde, both native and angio - data sets were processed in this technique for detection and assessment of calcified plaques. Authors reported the success rate of the proposed method as 85%. The study focused on the calcified plaques specially, NCPs were not addressed explicitly in this work.

The proposed experimentation was carried out in 6 steps. First stage was the localization of aorta that leads to the segmentation of the coronary arterial tree. After extracting coronary artery, the potential plaque candidates based on HU (defined threshold) values were identified. In order to eliminate false positives (included because of CT artifacts) from the plaque candidates, correspondence between two scans was accomplished via registration process between angio and native datasets as shown in the figure 4.13. Finally rule based approach was applied to maximize the sensitivity by minimizing false positives. Throughout the pipeline, state of the plaque detection system at any stage was represented by four sets.

1. A     (Unvarified plaque candidates from angio set)
2. N130(Unvarified plaque candidates from native set)
3. N200(Unvarified plaques candidates from native set but highly calcified)
4. V     (Varified plaques)

The ground truth for comparison of the obtained results was obtained with the help of a radiologist who marked the degree of the stenosis and the proximal & distal end positions of each plaque. Combination of native & angio data sets in the detection process achieved 85.5% detection rate according to the authors but this method leads to exposed radiation in terms of two scans. The computational load is also increased due to the required registration process between two independent scans.

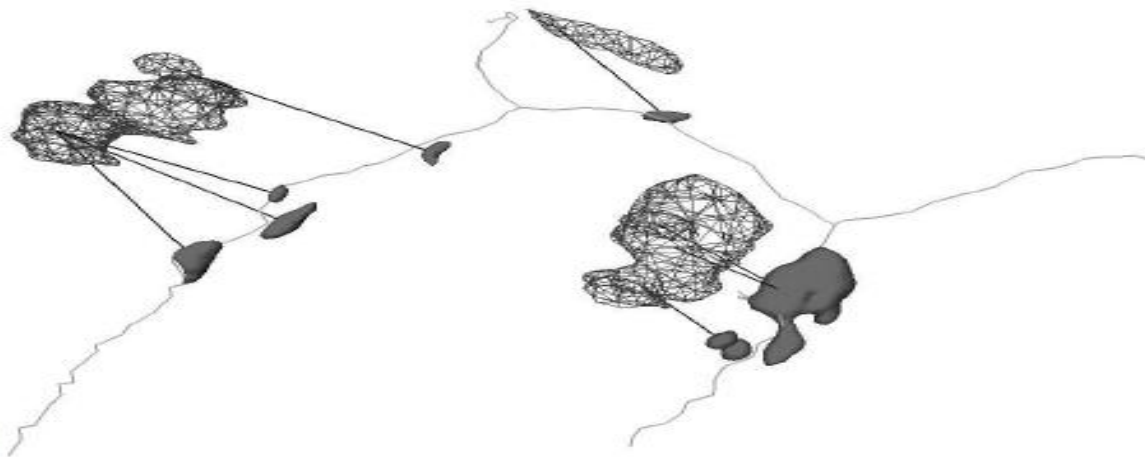

Figure 4.13.Correspondance between Angio (solid) and native (mesh) plaque candidates

## 4.7     Automatic Segmentation of Coronary Arteries and Detection of Stenosis

The focus of this work [95] was to design a fully automated framework for identification of coronary artery plaques by highlighting the discontinuities in the vessel. The performance of the proposed approach achieved 97% success rate as reported by the authors. Authors did not specifically included or excluded the scope of NCPs but referred to all plaques in results section. In the pre-processing step Input image was convolved with Gaussian filter to minimize possible CTA artifacts. After filtering, aorta was localized by applying connected component analysis. Vessel enhancement mechanism (based on Sobel operator) was applied to improve the connectedness between the branches of the coronary arteries and finally arteries were delineated by subtracting the localized aorta from the vessel enhance volume. Stenosis / calcification was detected by generating the centreline of segmented coronaries through skeletonization process. After generating the skeleton, discontinuities in the centreline were marked as clogged points as it symbolizes the presence of calcium/fats at corresponding location. The Intensity and diameter of the vessel at suspected points were evaluated and decision was made regarding degree of stenosis burden. Although the authors reported 97% success rate of this approach but it was very limited and based upon several manually selected thresholds. Figure 4.14 shows the detected stenosis points but no quantitative assessment was done in this study. Along with this, the proposed method was not tuned specifically for detecting soft plaques.

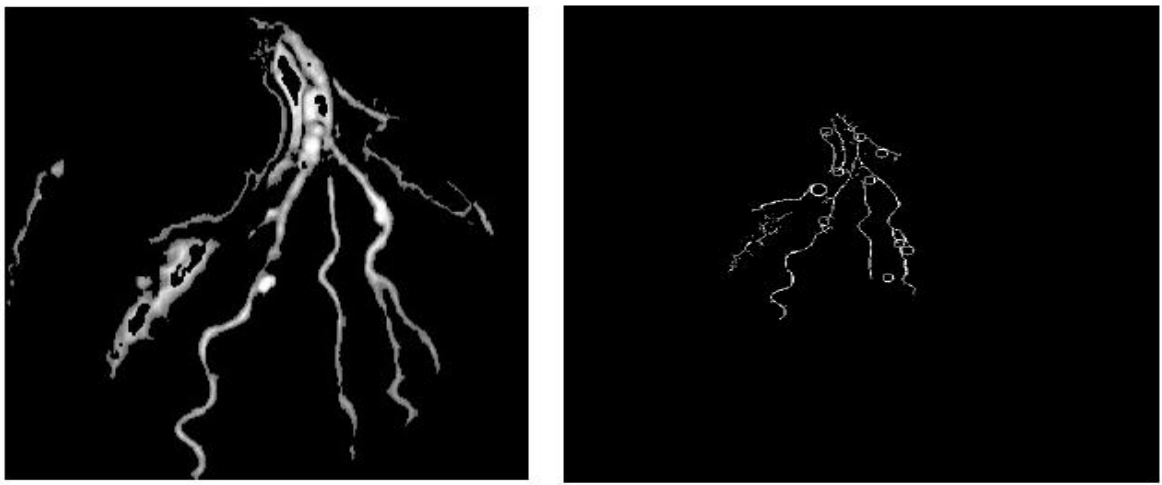

Figure 4.14.Arterial tree with marked Plaque points

## 4.8 Measuring Non-Calcified Coronary Atherosclerotic Plaque Using Voxel Analysis With MDCT Angiography: A Pilot Clinical Study

The main focus of this work [96] was to design a voxel analysis framework for quantification of non-calcified plaque in coronary arteries. Quantification was performed in terms of diameter & volume of the plaque. Total 49 arterial cross sections (41 Normal & 8 abnormal with non-calcified plaque) were chosen from a set of 40 patient CTA data. According to the reported results voxel analysis technique appears to be robust method for measuring the vessel wall thickness, vulnerable plaques & resultant stenosis burden. It is notable that abnormal arterial cross section refers to NCP plane where lesion did not occlude or narrow the arterial lumen > 70%. The cine projection was chosen to maximize the visual appearance & avoid the shortening or overlapping of branches. Specification for data used in the research is given in table4.1.

Table 4.1 Cross sections under observation for different segments of coronary vessel

| Artery segment | Normal Arterial sections chosen | Arterial section with plaques |
|---|---|---|
| Proximal right coronary | 10 | 0 |
| Mid right coronary | 7 | 1 |
| Left Main | 4 | 1 |
| Proximal left anterior descending | 8 | 4 |
| Mid left anterior descending | 4 | 2 |
| Proximal left circumflex | 8 | 0 |

Voxel Analysis was performed by plotting 8 radial lines at 45° in the arterial cross sections. Each line starts from epicardial fat & terminated inside lumen to ensure that the wall surface has been well traversed as shown in figure 4.15.

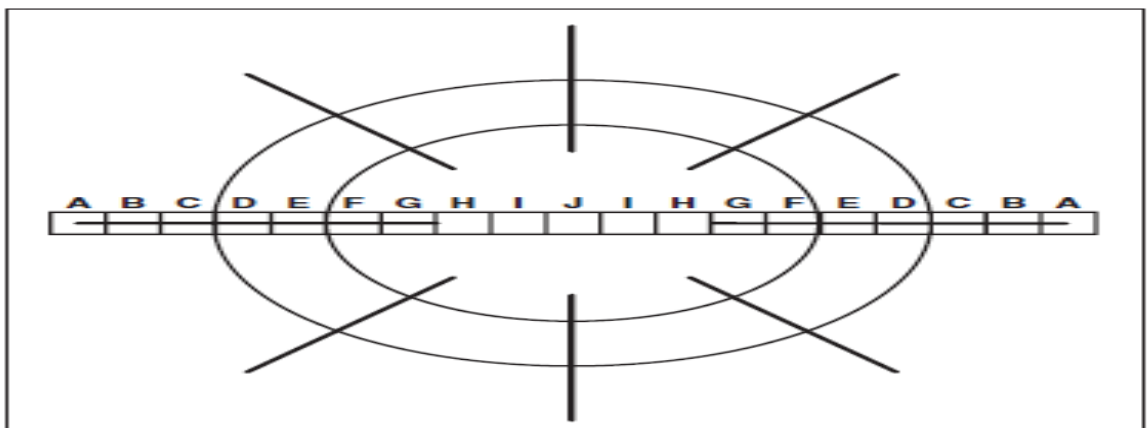

Figure 4.15. Passing radial lines across wall to record intensity values

For every line CT attenuation value was recorded at seven (7) locations A to G. (**A, B**) representing epicardial fat, followed by interface of epicardial fat & vessel wall (voxel **C**). (**D, E**) represents vessel wall itself followed by (**F, G**) that represents the inner lumen. Plaque detection / identification mechanism used by authors is illustrated in figure 4.16 (b) where the attenuation values for plane passing through point3 (having plaque inside wall) are plotted. The density of wall voxels (E=66) is less than normal segment.

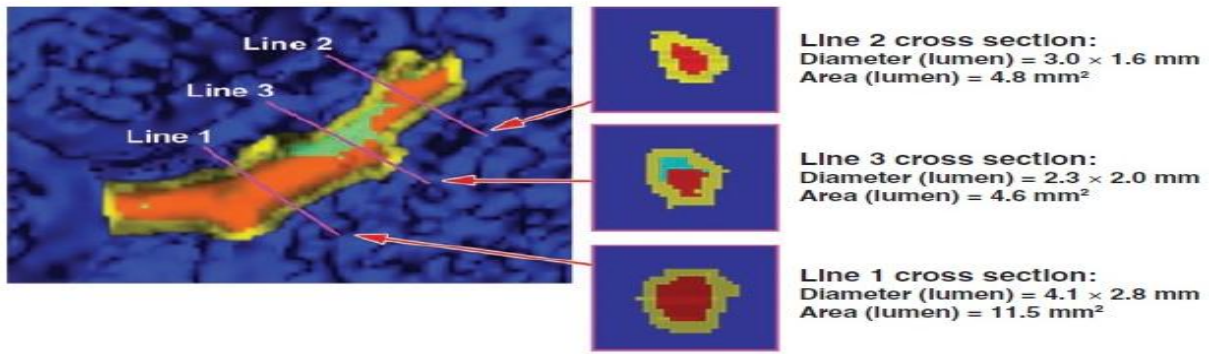

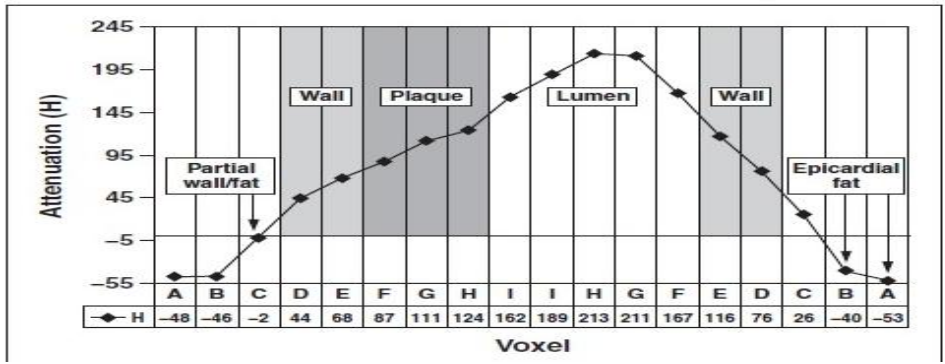

Figure 4.16. Intensity value plot for radial lines(A-G....G-A) passing through point3 (NCP present)

Total 2296 voxel intensity values (41planes * 8lines * 7voxels) for normal arterial sections were recorded. Moreover 448 voxel intensity values (8planes * 8lines * 7voxels) for plaqued arterial sections were recorded and the statistical analysis validated the method by spotting plaqued regions.

As shown in the figure 4.17, the mean attenuation values of wall voxels (E–G) are significantly lower than their counterparts in the non plaqued sections. This indicates the presence of lower density structure (non-calcified plaque) compared with higher-density material (contrast medium and blood) in the normal cross sections.

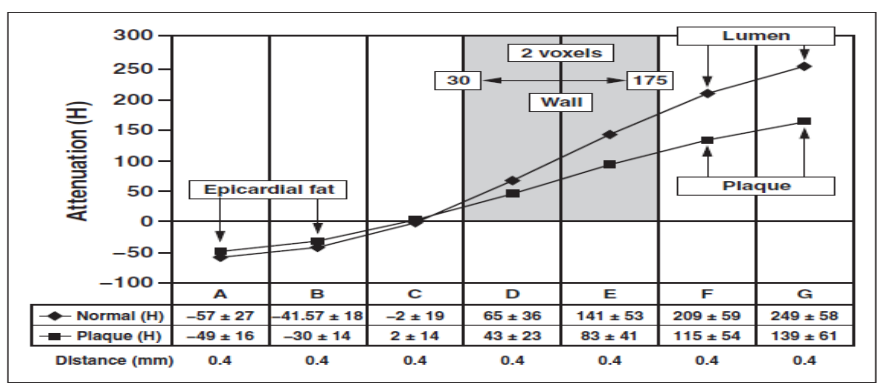

Figure 4.17. Mean Intensity values for voxels in Normal & Clogged arteries (Vessel wall voxels "E" have notably less value for plaqued voxels in comparison with normal)

### 4.9 A Voxel-Map Quantitative Analysis Approach for Atherosclerotic Noncalcified Plaques of the Coronary Artery Tree

Main focus of this work [97] was to develop a quantitative analysis framework for detection and quantification of soft plaques in coronary arteries. Test CTA data for this research was obtained in a controlled environment at a medical centre. According to the work of [78 & 79] pixel having CT attenuation number greater than 160Hu was considered as first voxel of the lumen. Consequently all the lumen voxels were supposed to have value greater than 160 whereas the voxels having value less than 160 were assumed as external voxels. This cut-off value was used to delineate all the lumen voxels with the help of region growing algorithm. This segmented coronary artery mask was used to extract arterial tree from CTA volume & skeleton centreline was generated using proprietary software Amira (v.5.4).

Voxel map was generated by applying morphological operations dilation and erosion. Dilation reflects the voxel changes outwards (boundary layers are termed as $B_1$, $B_2$, $B_3$…) whereas erosion mirrors voxel changes inside lumen (boundary layers are termed as $B_{-1}$, $B_{-2}$, $B_{-3…}$). Figure 4.18 (a) represents a cross sectional plane intersecting the arteries orthogonally, whereas (b) shows a coarser view.

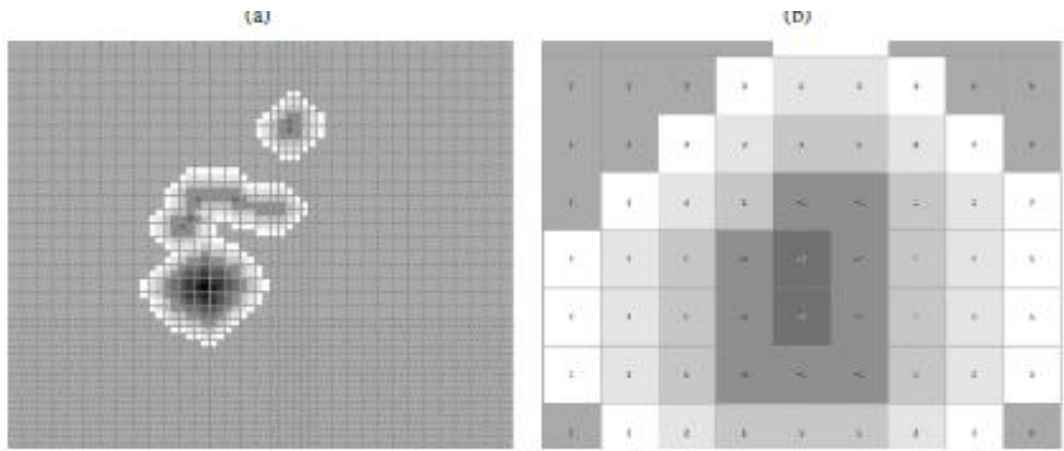

Figure 4.18. Voxel map after dilation & erosion

After generation of voxel map, the vessel wall (from outer border of **lumen** to the outer border of **wall**) was divided into four layers namely -1, 1, 2, 3. The attenuation values on the wall were divided into 6 groups to define the severity of the plaque composition and assigned different colors. These distinct colors were associated for better visual experience. Figure 4.19 represents the application of the voxel map based method for identifying the vascular lesion present in arteries.

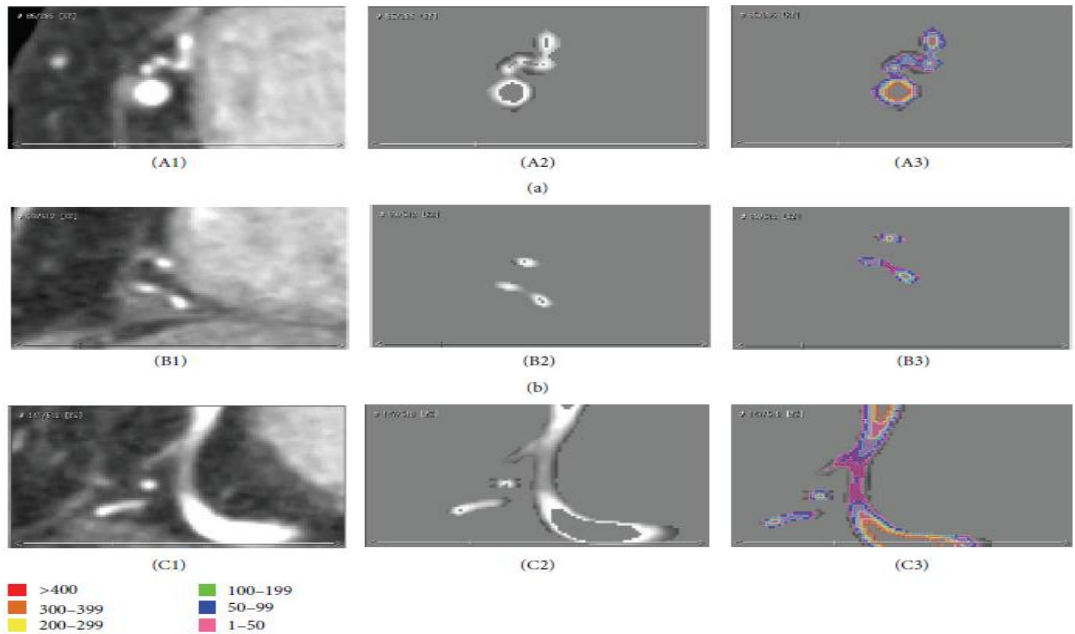

Figure 4.19.(Left) CTA planes with different views, (Middle). Arterial wall model after erosion/dilation. (Right) Visually enhanced view based on CT attenuation values

The Inner lumen intensity value increases sharply as approaches close to the aorta due to increased concentration of the contrast agent whereas the CT value remains stable for boundary of vessel wall adjacent to lumen as shown in figure 4.20(a). Afterwards change in HU values (gradient) was recorded at four defined layers that shows that mean CT value decreases from inside to outside of vessel. The abnormal behavior of gradient is related to plaque existence.

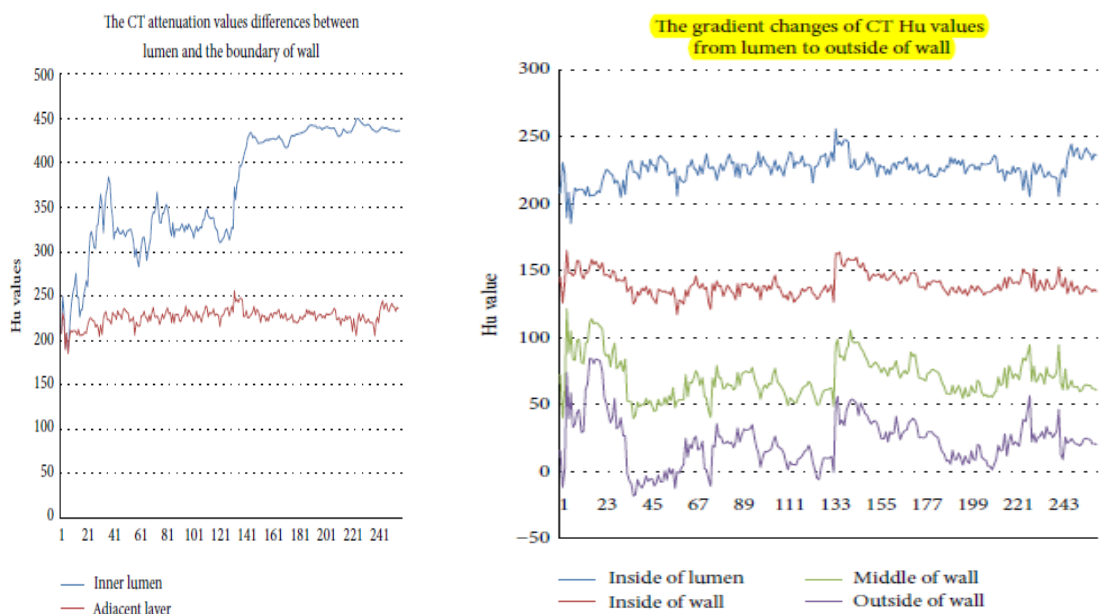

Figure 4.20. Left Mean CT value of lumen voxels versus boundary voxels. (Right) Gradient of CT value at four layers of vessel wall from Inner lumen to outer vessel boundary.

# Chapter 5 Proposed Framework

This chapter presents the proposed framework for automatic segmentation & soft plaque detection. This is a two stage process as represented in Figure 5.1(a) where phase-I comprises of automatic coronary tree segmentation based on single seed point. In the subsequent stage the detected tree will be used in vessel wall analysis for detection & quantification of soft plaques. A simple yet effective approach has been used in phase-I for accomplishing the task of coronary tree segmentation. Figure 5.1 (b) represents work flow for phase-I of this project which has been completed. Process of seed detection has been automated completely by incorporating the shape features of coronary vessel in CTA volume. In the consequent stage, vesselness is calculated for CTA voxels based on Hessian matrix analysis which gives strong markers to extract shapes. The potential candidate voxels having significant vesselness measure are further thresholded to integrate the impact of the contrast agent in respective CTA volume. In the final stage, region growing segmentation is implemented to extract complete arterial tree. To ensure the optimal segmentation for medical images, localized intensity information is used in active contour based evolution. Two- way segmentation utilizes seed information in both directions (forward & backward direction with respect to axial planes as seed lies in the mid of CTA volume) for extraction of complete arterial tree. The last stage of phase-I is "skeleton generation" based on fast marching method as it is required for computing oblique planes in lumen & wall quantification. Extracted arterial tree and centreline will be used in phase- II where a comprehensive intensity & geometry analysis will be performed for segmentation & quantification of soft plaques. Implementation details and accomplished results for phase-I are presented in this chapter.

## 5.1 Automatic Seed Detection in CTA volume

Automatic seed detection process is based on the work of Han *et al* [87] where Hessian based filtering was combined with a local geometric measure to effectively track coronary arteries. Due to the elongated nature of the coronary arteries, a cylindrical model based geometrical analysis was applied for detecting seed point value for coronary arteries. According to authors obtained seed points(values for individual branch) were used in particle based filtering for centreline extraction however complete arterial tree segmentation has not been reported yet. We propose a modification to improve the efficiency of the seed detection method & consequently seed points are used in region growing based segmentation for extracting complete 3D arterial surfaces from CTA. Main strength of this method is fully automatic selection of initial seed points, which leads to complete automatic coronary segmentation in a robust manner. No pre-processing is required for an adequate quality CTA image; however this method incorporates prior anatomical knowledge about coronary arteries in terms of shape information.

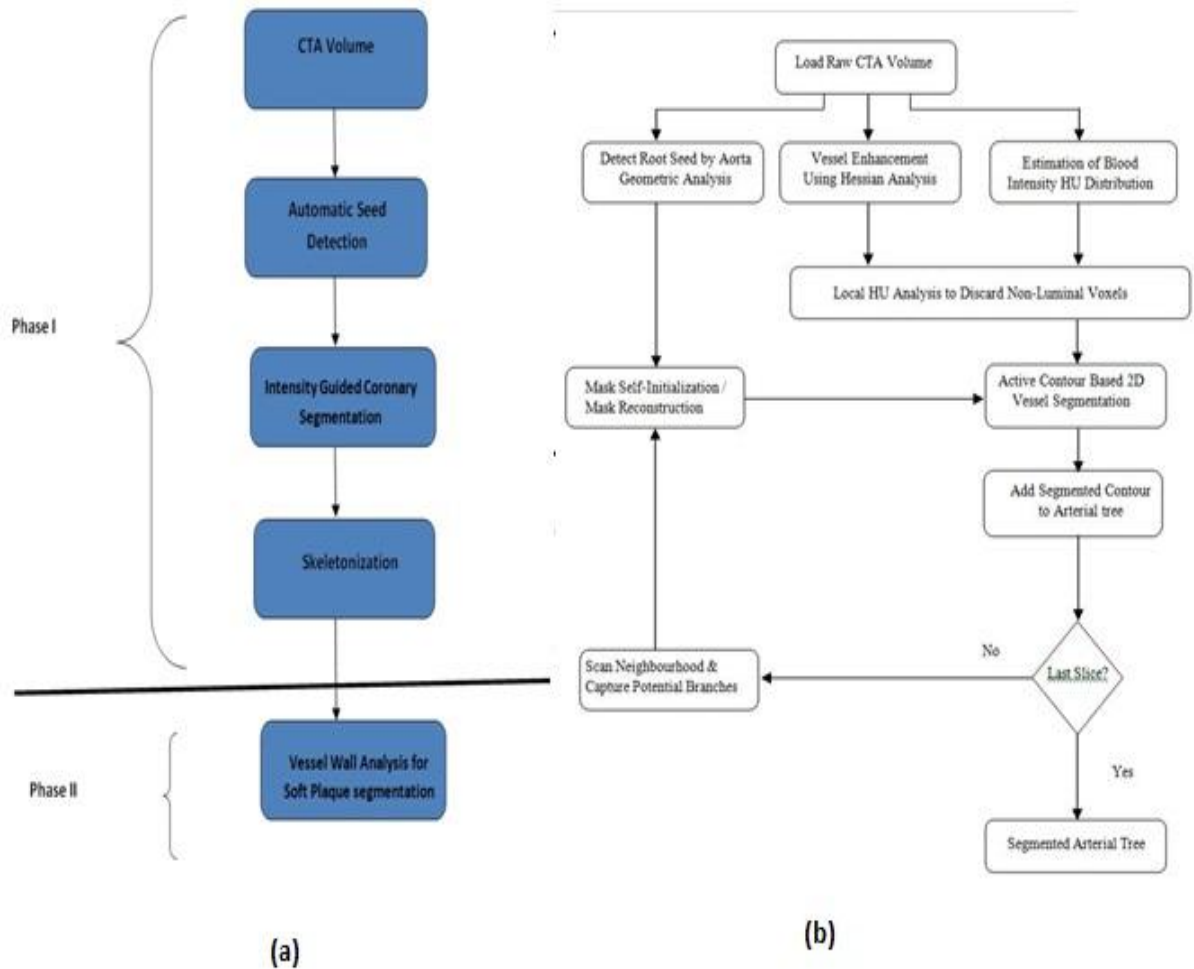

Figure 5.1. Work flow for project (a) Summary for two phase plaque detection project (b) Expansion for phase-I activities addressing automatic arterial tree segmentation

In CTA volume, 2D axial slices represent coronary arteries as tubular structures along z-axis where arterial geometry remains collateral through consecutive slices for major branches i.e. (LCA, LCX, LAD & RCA) appear as bright circular /elliptic objects in axial slices from aorta to distal end points). This clue can be used in mathematical analysis to discard all non vascular structures found in 2D axial slice. Although Frangi vesselness effectively detects vascular components but CTA data is sometimes misinterpreted due to inability to handles edges. As a result, edge voxels are also assigned high vessel measure and increased false positive rate degrades computational efficiency. Yang et al [86] proposed to use local geometric feature WF(x) based on sphere ray-casting but processing time is increased exponentially due to connected component analysis. Work by [87] proposed a new localized feature ($GF(x)$) based on consecutive orthogonal cross sections analysis that measures the similarity index of geometric shape in consecutive planes. High similarity value leads to vessel presence whereas low value reflects non-vasculature shapes.

Seed detection process starts with selection of reference slice from middle of CTA volume to ensure that major coronary arteries are captured. In the following stage, edge detection is applied to isolate different objects present in reference axial slice. Potential region of interest (ROIs) are detected by discarding open boundary objects as it is assumed that coronary arteries will appear as circular or elliptical cross sections in axial slices having closed boundary. Potential seed points are determined by obtaining centroid values of all detected ROIs. Final seed values are selected by investigating localized geometric feature strength that is based on shape similarity through consecutive slices. Consequently all selected ROIs (representing coronaries) are identified & respective centroid values are interpreted as final seed point(s). Depending upon the reference slice, multiple seed points are obtained simultaneously; however only one seed point is required for extraction of complete arterial tree. We integrated the local contrast behavior in terms of HU information to select the strongest seed point for respective coronary arteries.

### 5.1.1   Region of Interest (ROI) selection in an axial slice

General behavior of cardiac CTA reveals that coronary arteries appear as bright closed structures (circular or elliptic shape) & run almost perpendicular to the axial slice( along z-axis) in the mean of scanned volume. First step in automatic seed selection is to select appropriate reference slice that containing left & right coronary arteries. This can be selected from a wide range of axial slices centered on middle of the CTA volume. Reference slice index P can be can be modeled according to Equation 5.1

$$P = Cr \times N \qquad\qquad (5.1)$$

Where Cr is the constant controlling the reference slice index & N represents the total number of slices in given CTA image. Generally value of Cr ranging (0.4—0.6) ensures that selected slice contain all major branches of Coronary tree. After selection of reference slice, computational load can be reduced by extracting the heart region only, since coronary arteries lies on heart surface. Irrelevant background information is discarded by using fast heart isolation tecnhique proposed by Wang [88] to retain only relevant information.

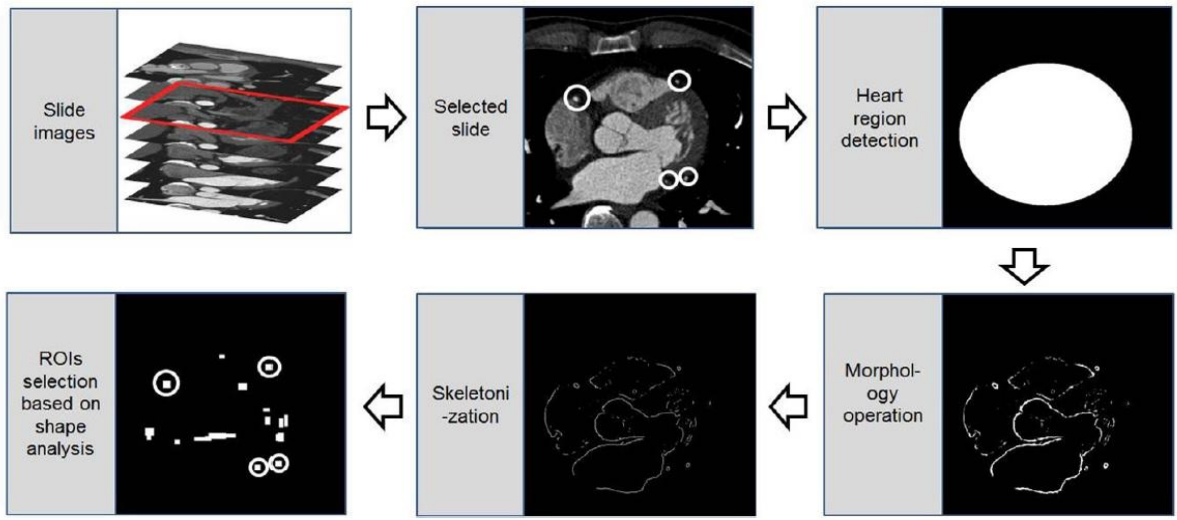

Fig.5.2 process explaining selection of ROIs for seed point detection

Based on the heart region obtained potential edges are detected using Sobel edge detection. Contour tracing operation based on length & curvature analysis of object boundary is applied to remove open boundaries (according to assumption that coronary arteries are close boundary objects). Figure 5.2 represents step wise implementation of ROI selection process whereas contour tracing process for discarding open boundary objects is illustrated in figure 5.3. Closed boundary objects are extracted from CTA volume using bounding box & hessian based vesselness measure is calculated as proposed by Frangi metric. Removal of open boundary objects speed up the process of vessel identification since it is useless to evaluate remaining                                                                       pixels.

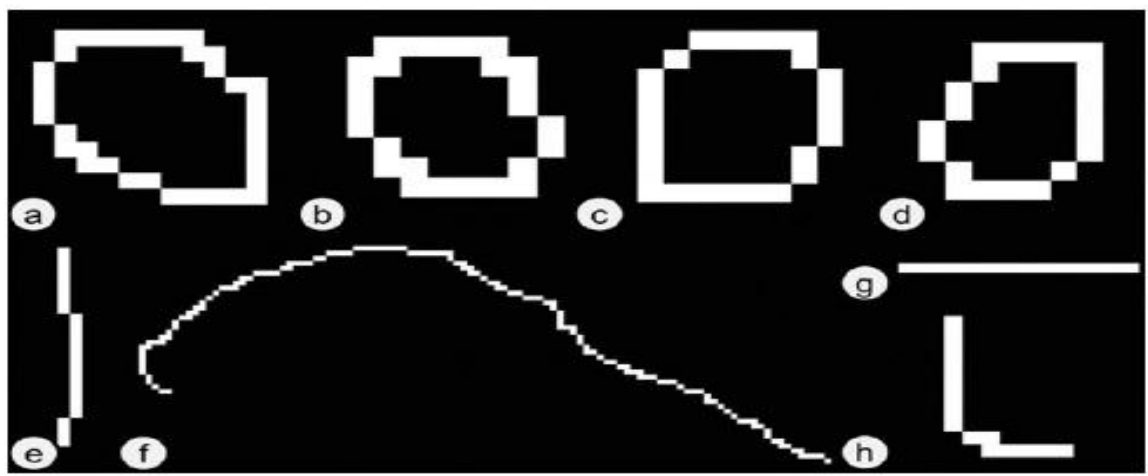

Fig.5.3 Contour tracing for elimination of open boundary objects. (Objects labeled a-e are retained due to close boundary, whereas objects labeled f-h are discarded due to open boundary)

### 5.1.2 Geometric Shape Analysis for individual ROI

Inherent inability to handle step edge response in CTA data makes Frangi vesselness a bit vulnerable in terms of increased false positives. For seed detection, it is not acceptable as the subsequent arterial tree segmentation relies on detected seed. To address this limitation, a localized geometric feature GF(x) is proposed in [86] that is calculated by performing shape analysis of consecutive orthogonal cross sections. At a vessel point directional information is obtained with the help of Eigen values. Eigen vector of Hessian matrix corresponding to minimal Eigen value $\lambda_1 (|\lambda_1| \leq |\lambda_2| \leq |\lambda_3|)$ of the Hessian matrix represents vessel direction at voxel $A$. Three consecutive oblique planes that are orthogonal to the vessel direction are extracted. Planes are assigned index values [-1, 0, 1] as shown in Figure 5.4.

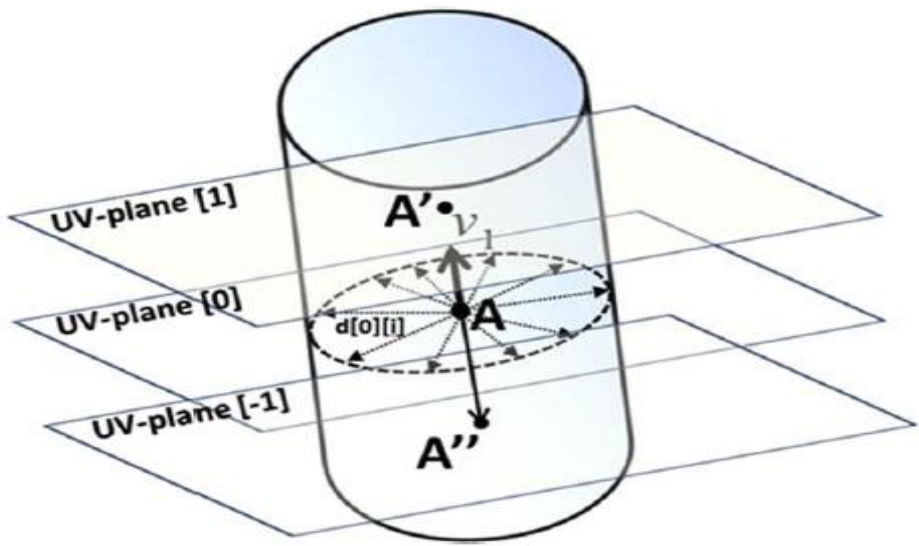

Fig.5.4 Consecutive planes orthogonal to vessel direction used in cylindrical modeling of vessel

Centre plane UV [0] passes through point "A" i.e. the centroid of the region of interest & it is orthogonal to the direction of vessel represented by eigen vector. Two consecutive planes (forward UV [1] & backward UV [-1]) are parallel to plane UV [0] at parametric distance of D units. Shape similarity is measured by correlating shapes through these three consecutive planes. For establishing shape correlation, ray casting is performed on each plane in 16 uniformly sampled directions based on respective centre points, as shown in figure 5.4. Boundary is detected along each ray by examining the radial gradient, where a significant change represents border point. Distance between border point & centre of the plane is interpreted as ray length in corresponding direction. After sorting 16 ray lengths, highest & lowest three are discarded to avoid abnormalities. Remaining 10 ray lengths are arranged for every plane in a 2D data structure representing [plane][ray] index. For each ray index ray = [1, 2, . . . , 10], the minimum values $B_{min}$ [ray] and the maximum values $B_{max}$ [ray] among the three ray lengths plane[−1, 0, 1 ][ray] are calculated & finally the proposed local geometric feature is defined as follows:

$$GF(x) = \prod_{j=1}^{10} \frac{k}{B_{max}[ray] - B_{min}[ray] + 1} \qquad (5.2)$$

Where $k$ is some constant. Fig. 5.5 shows how this measure can differentiate vessels from non vascular structures by investigating minimum & maximum value profiles for two objects. In case of a vessel point (A), border on three consecutive cross sections does not show a significant change in radius in different directions i.e. minimum values remain closer to maximum values. This small difference will lead to high value assigned to geometric feature GF(x) in Equation 5.2. In contrast non-vessel points will exhibit high difference in minimal & maximal ray lengths, resulting in small value of GF(x) that remain insignificant. In combination with Frangi vesselness, new information that ensures the shape similarity is respected as given in Equation 5.3.

$$vesselness(x) = \begin{cases} 1 & if\ F(x) \geq T_f\ \&\ GF(x) \geq T_{Gf} \\ 0 & otherwise \end{cases} \qquad (5.3)$$

Where T represents threshold values for local geometric feature & Hessian based vesselness for controlled experimentation. An effective amendment to ensure only coronary based points is integration of contrast behavior in vesselness calculation. Additional threshold ($V_T$) representing the contrast medium diffusion in respective CTA volume is imposed as additional constraint in vesselness measure of equation 5.3. A particular pixel (centroid of the ROI) will be assigned final vesselness measure equal to 1 when Hessian, geometric and intensity thresholds are surpassed as defined in equation 5.4.

$$vesselness(x) = \begin{cases} 1 & if\ F(x) \geq T_f\ \&\ GF(x) \geq T_{Gf}\ \&\ I(x) \geq V_T \\ 0 & otherwise \end{cases} \qquad (5.4)$$

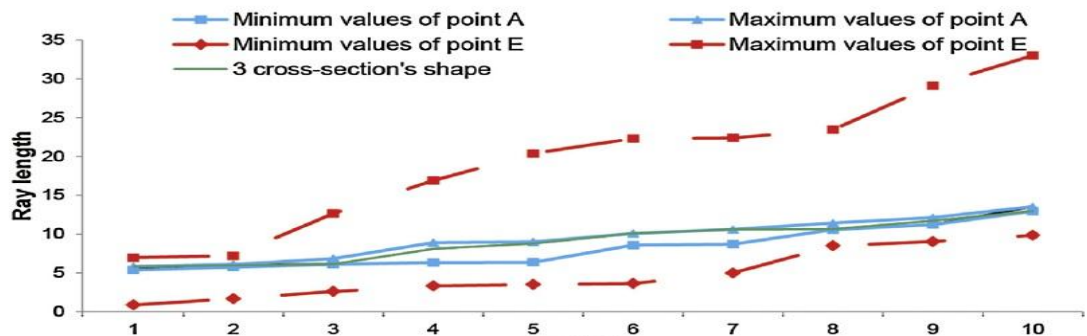

Fig.5.5 Separating Vessel based ROI from Non vascular regions.

Figure 5.6 –figure 5.9 illustrates the process of automatic seed detection for CTA volume 3.

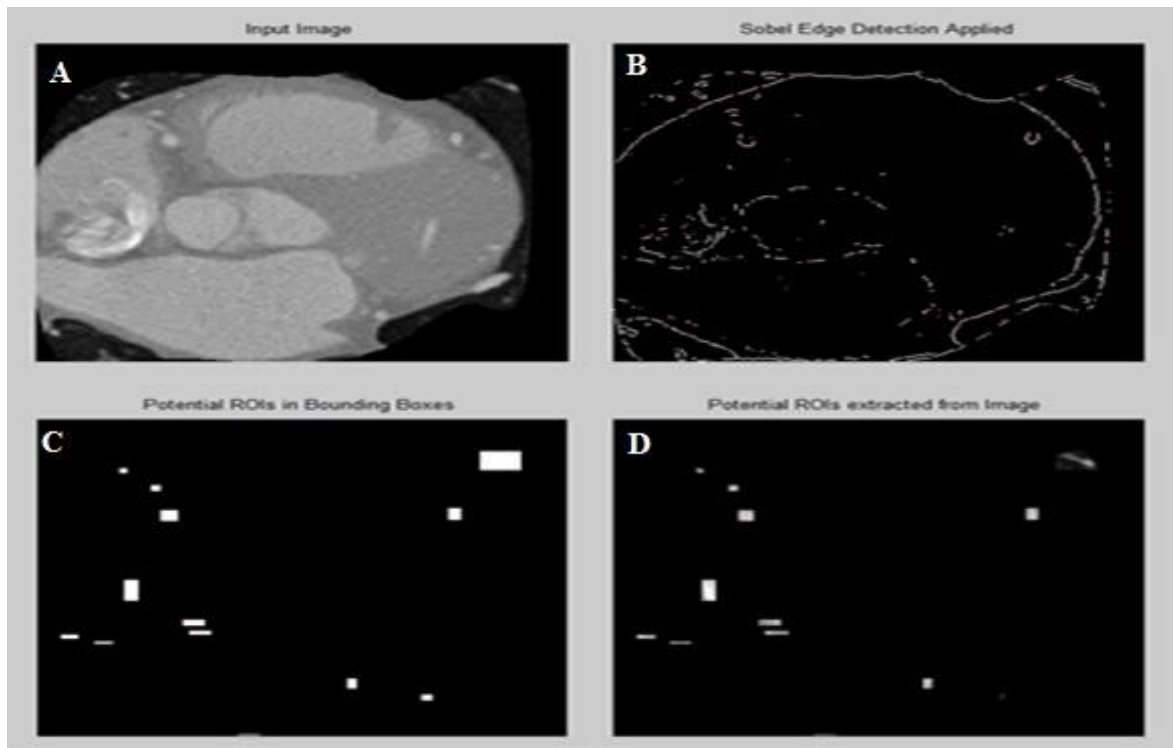

Fig.5.6 Region of Interest (ROI) selection for CTA Volume 3 with Cr=0.27. (A) Axial Image from CTA image. (B) Sobel detection applied (C) Bounding boxes for potential ROIs (D) ROIs extracted from CTA Volume

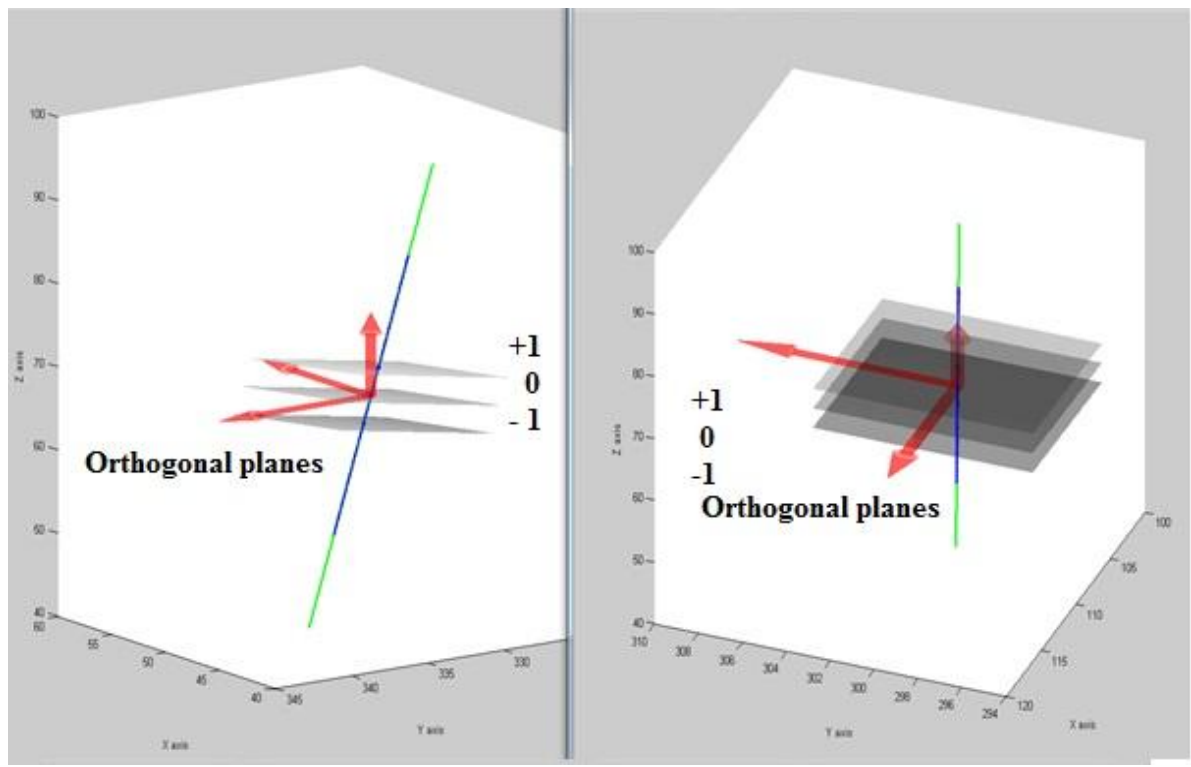

Fig.5.7 Illustration of consecutive orthogonal planes analysis for different ROIs of CTA Volume3

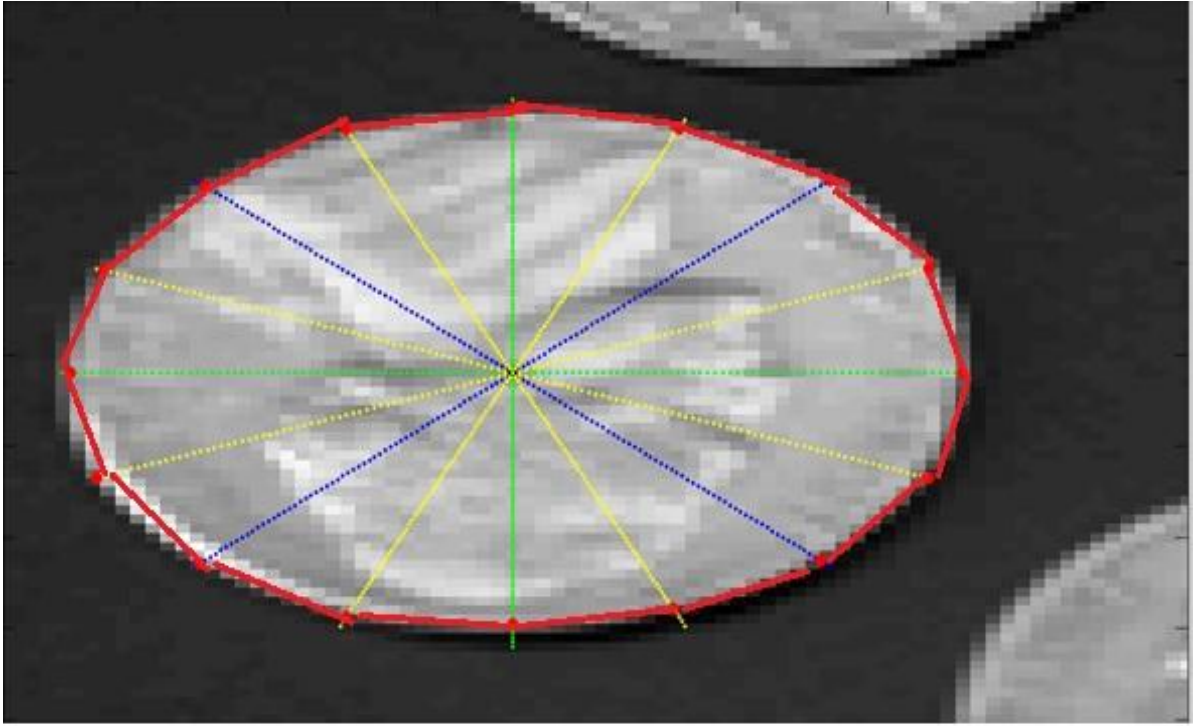

Fig.5.8 Ray Casting illustrated for CTA Volume3 to obtain Geometrical parameter GFx

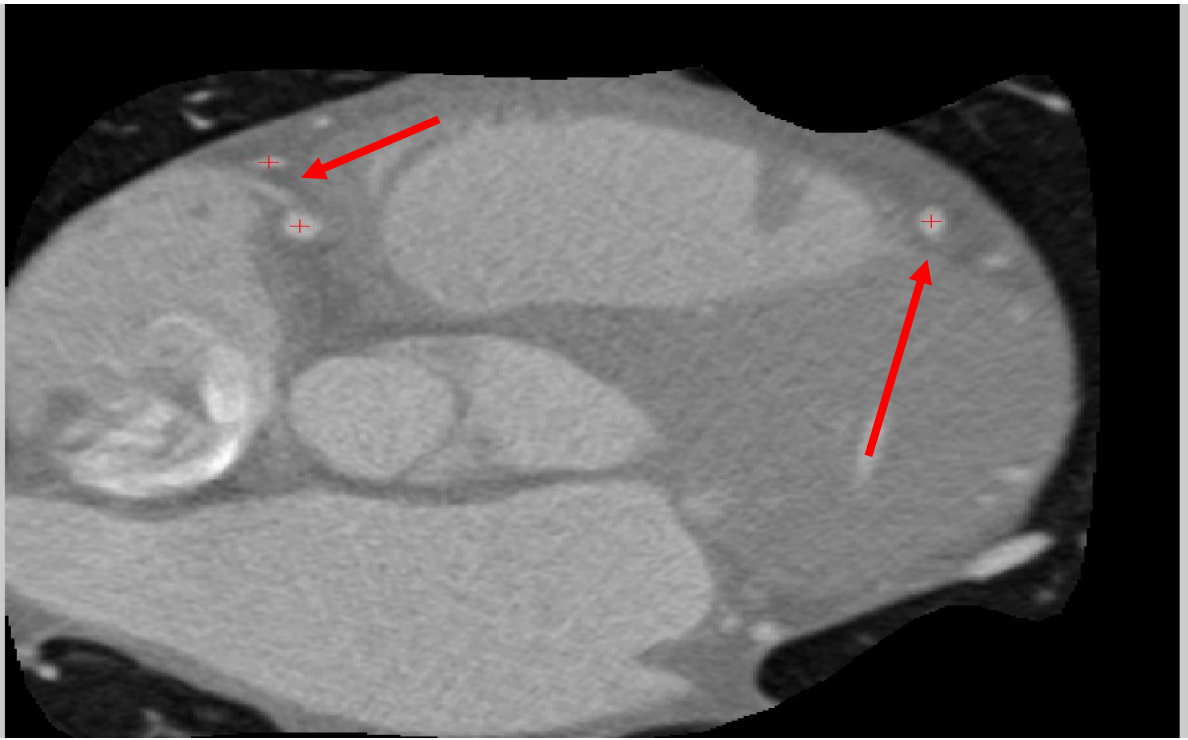

Fig.5.9 Automatic detected seed points for Left & Right Coronary arteries in CTA volume 3. Seed points are marked with "+"

Figure 5.10 –figure 5.13 illustrates the process of automatic seed detection for CTA volume 2.

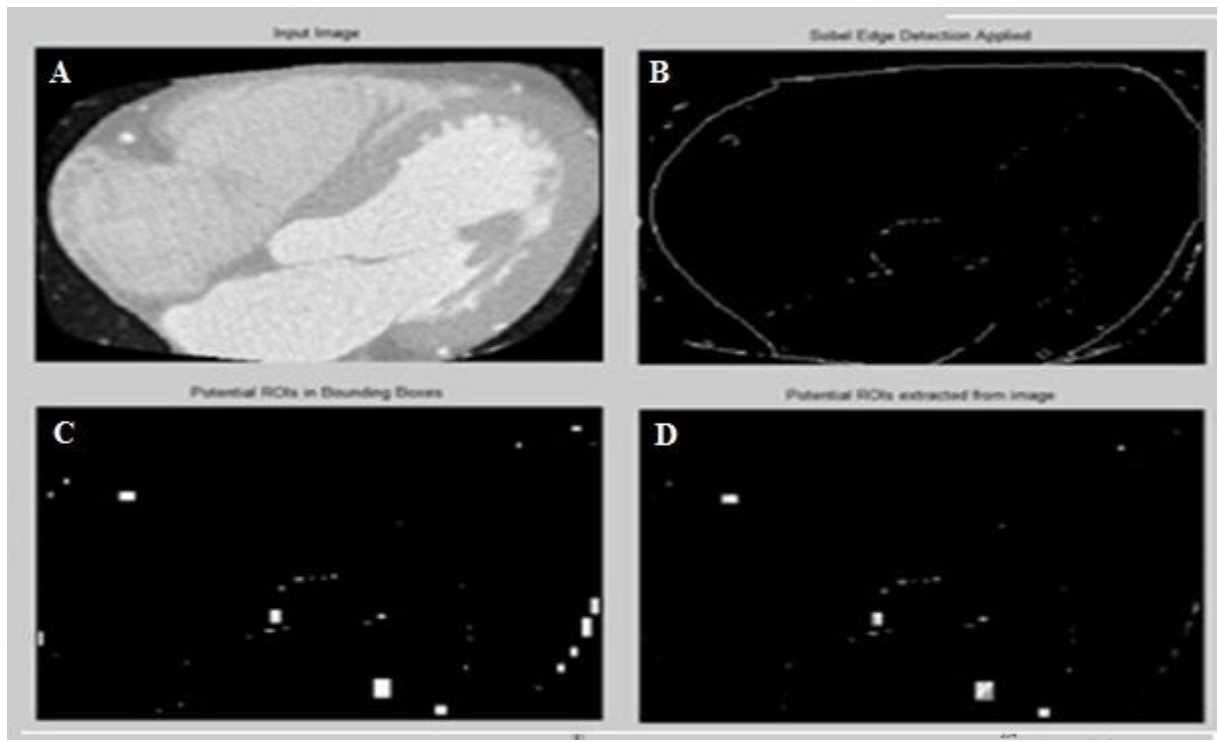

Fig.5.10 Region of Interest (ROI) selection for CTA Volume 2 with Cr=0.55. (A) Axial Image from CTA image. (B) Sobel detection applied (C) Bounding boxes for potential ROIs (D) ROIs extracted from CTA Volume

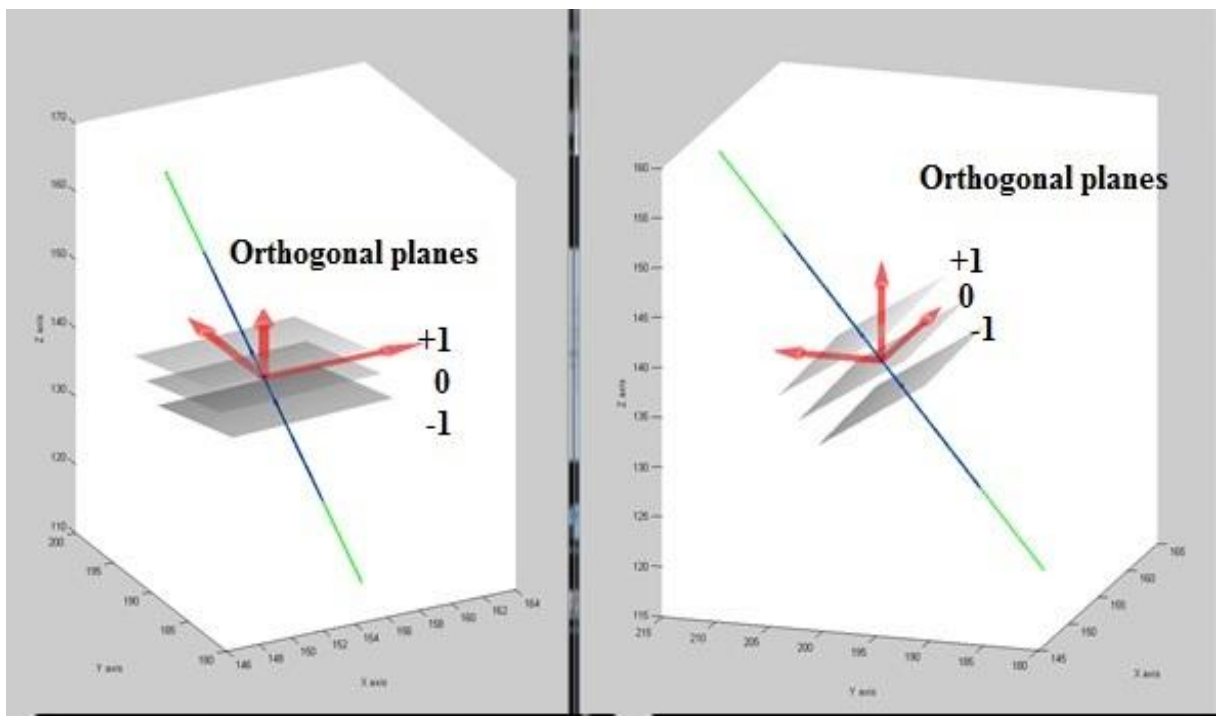

Fig.5.11 Illustration of consecutive orthogonal planes analysis for different ROIs of CTA Volume 2

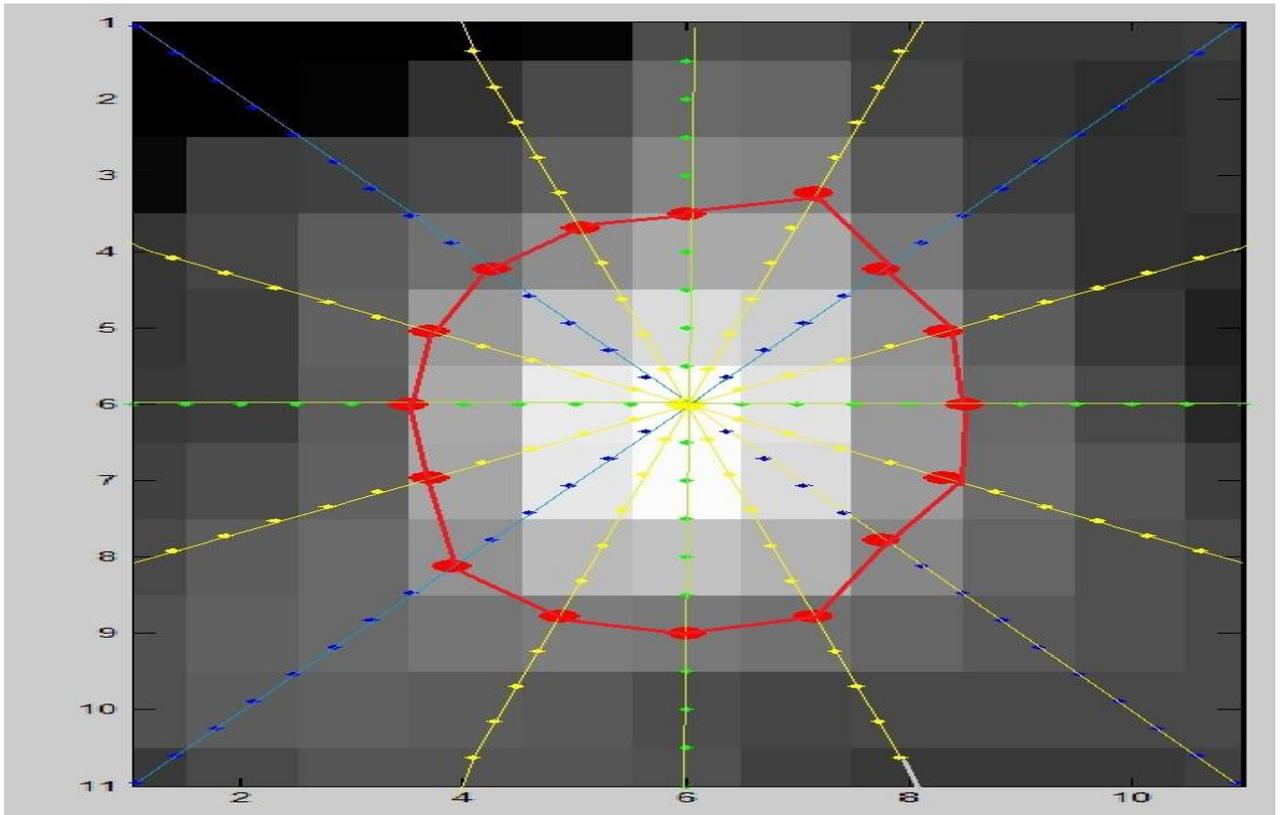

Fig.5.12 Ray Casting illustrated for CTA Volume2 to obtain Geometrical l parameter GFx (Zoom-in view)

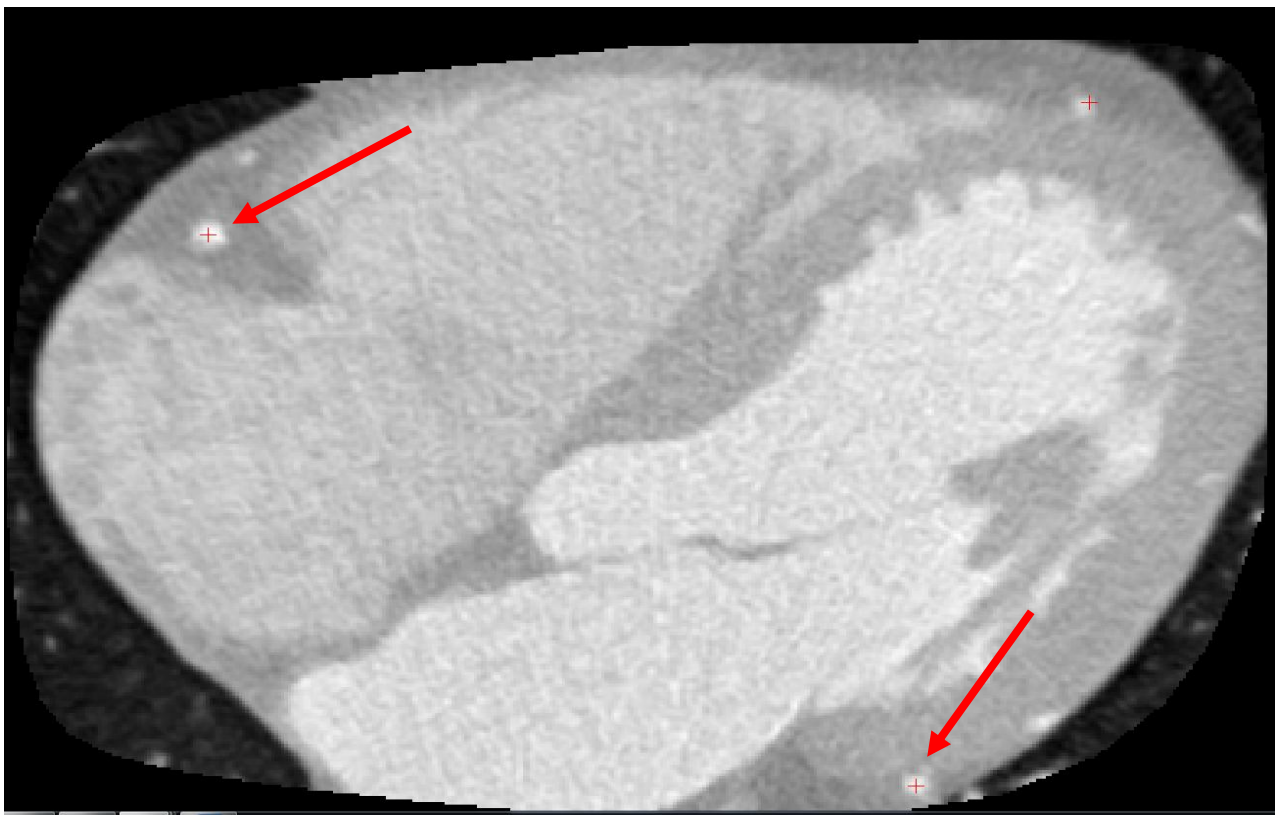

Fig.5.13 Automatic detected seed points for Left & Right Coronary arteries in CTA volume 2. Seed points are marked with "+"

Figure 5.14 –figure 5.17 illustrates the process of automatic seed detection for CTA volume 1.

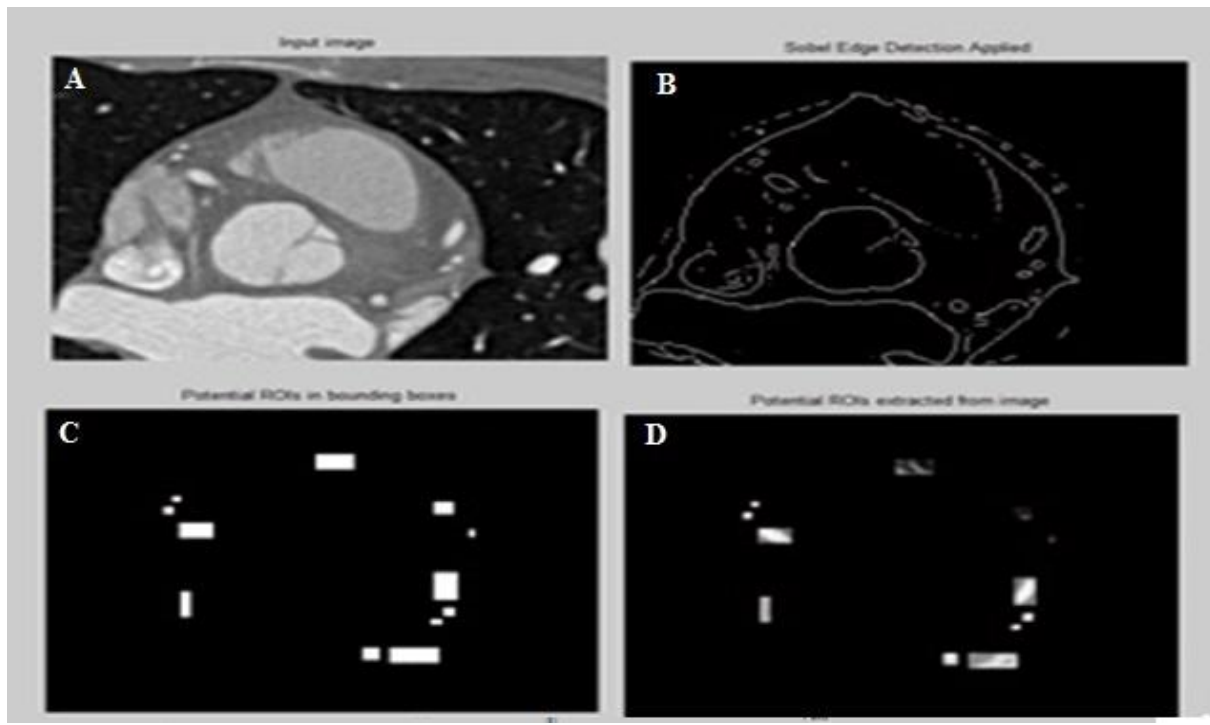

Fig.5.14 Region of Interest (ROI) selection for CTA Volume 1 with Cr=0.46. (A) Axial Image from CTA image. (B) Sobel detection applied (C) Bounding boxes for potential ROIs (D) ROIs extracted from CTA Volume

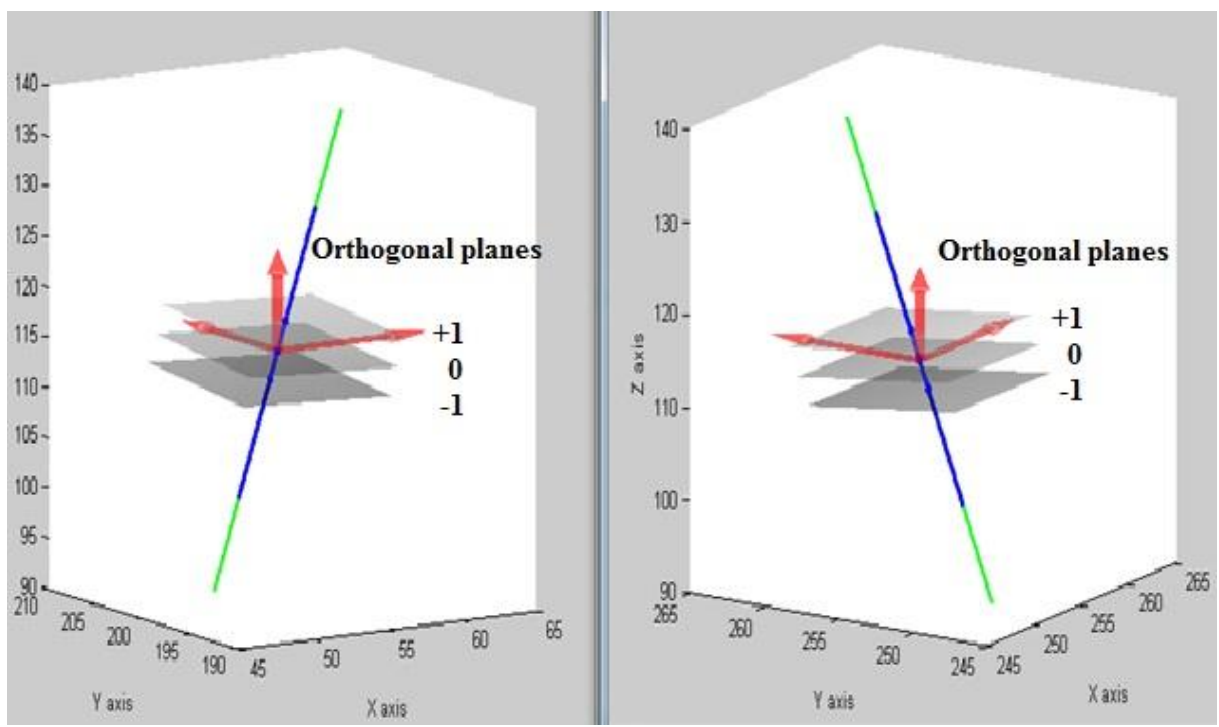

Fig.5.15 Illustration of consecutive orthogonal planes analysis for different ROIs of CTA Volume 1

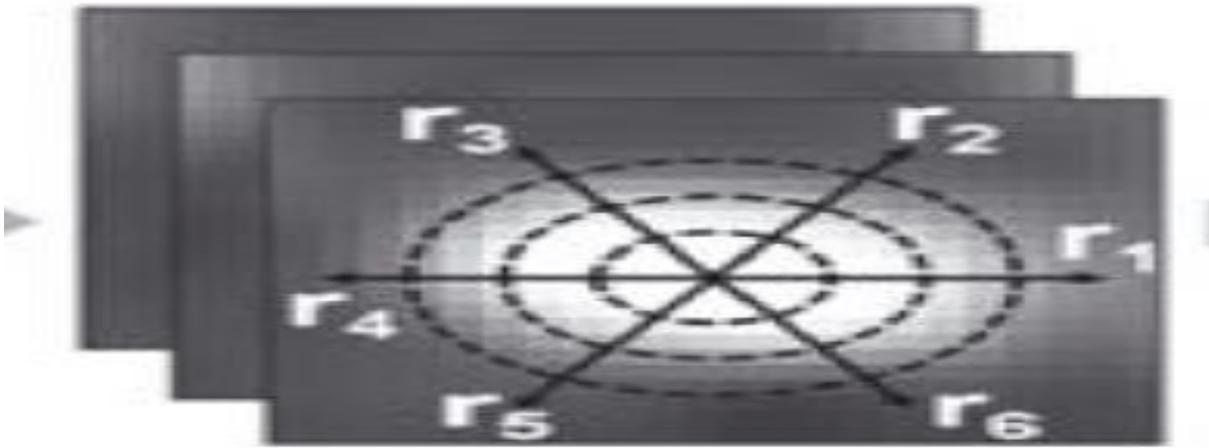

Fig.5.16 Ray Casting illustrated for CTA Volume1 to obtain Geometrical parameter GFx

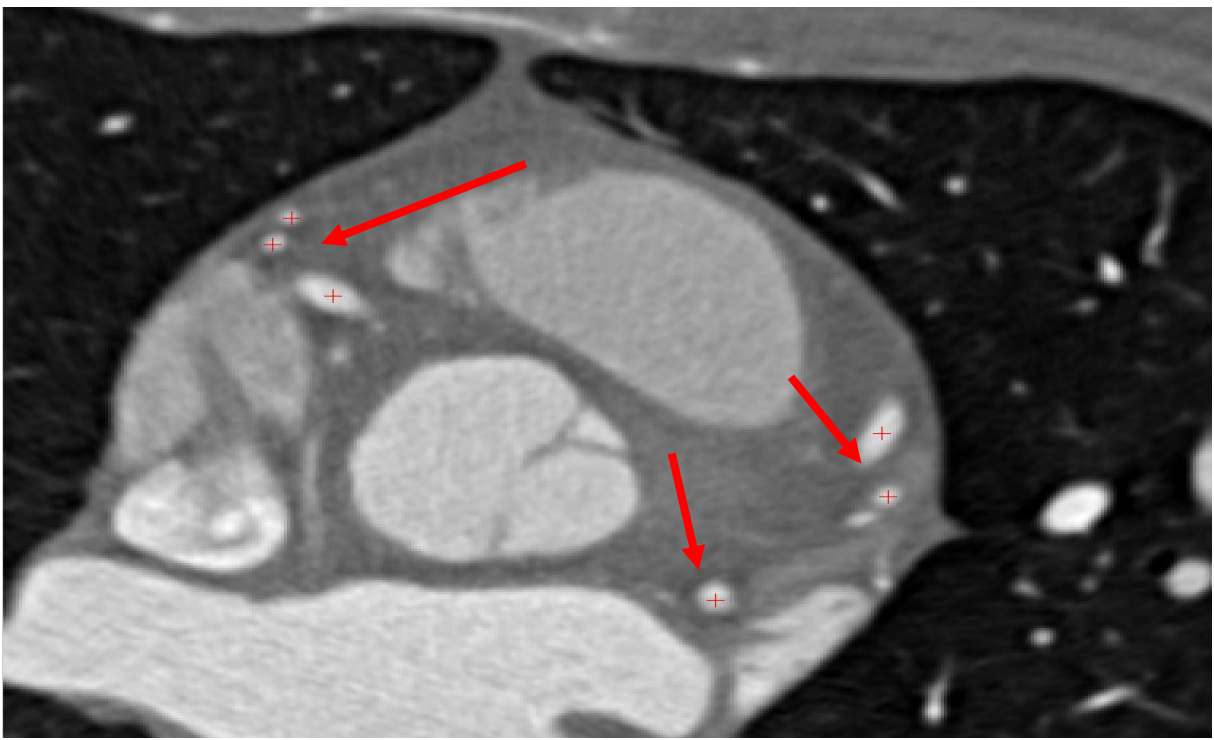

Fig.5.17 Automatic detected seed points for Left & Right Coronary arteries in CTA volume 2. Seed points are marked with "+"

## 5.2    Hessian Based Vessel Enhancement for CTA Volume

Hessian based Frangi vesselness measure is frequently used for extracting structures of interest from complex medical images. However some inherent limitations of this method make computational process passive. It becomes difficult to distinguish line & step edges using Frangi measure resulting in increased false positives. In CTA data, step edges represents heart chamber boundaries or edges separating heart muscles from lungs. So these edges must be assigned very low vesselness measure. Lorenz also proposed edge suppression for increased efficiency by applying edge indication given in equation 5.1.

$$E(x) = \frac{2|\nabla f(x,s)|}{s(\lambda_1(x,s) + \lambda_2(x,s))}$$

(5.4)

Where $x$ is the location of the current voxel in the volume image, $s$ represents the scale and $\lambda$ represents respective Eigen values of Hessian matrix. According to equation 5.4, strength of edge E(x) incorporates local gradient & Eigen values. Within a vessel, strong Eigen values will minimize the edge measure E(x) whereas on object boundaries resultant edge measure attains higher value due to strong gradient. Generally for CTA images, edges produced by heart chambers are relative weak, so the measurement *E(x)* will be small in magnitude & Lorenz edge detector fails to capture these edges. Based on the observation that surrounding voxels of heart chamber have high intensity, local intensity utilization was proposed by Yin that suppresses non-vascular voxels despite of weak edges.  Equation 5.2 shows the modified Lorenz edge detector that can reduce the false edges in vessel enhancement process of CTA data.

$$E(x) = \frac{k \cdot f(x) \cdot |\nabla f(x,s)|}{s(\lambda_1(x,s) + \lambda_2(x,s))}$$

(5.5)

Where *f(x)* is the intensity value at position ***x***.  Figure 5.18 represents the effectiveness of localized        information        integration,        used        in        this        work.

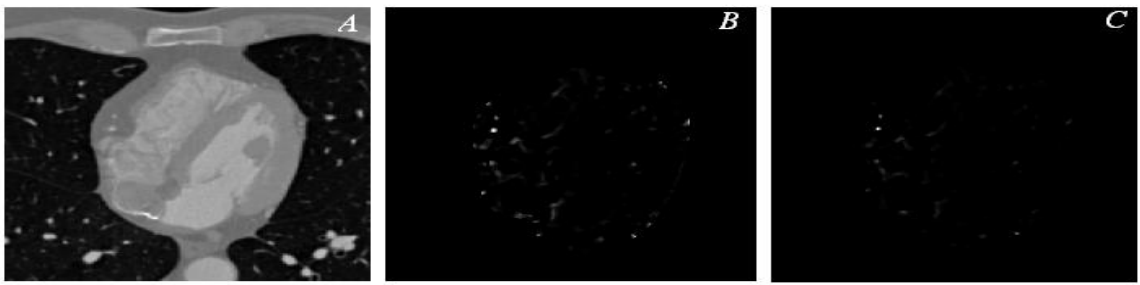

Fig.5.18 Vesselness measure. (A) Original Image, (B) Frangi metrics application, (C) Edge suppressed output

In our work every voxel is assigned a vesselness strength that describes the likelihood that voxel belongs to vasculature structure. Edge suppression technique has been tested for different CTA volumes & only a marginal improvement in computational efficiency has been observed. Segmented tree does not show significant change as our active contour evolution algorithm already incorporates the contrast agent behavior in terms of HU value range, so heart chamber based weak edges are automatically discarded during segmentation. The effectiveness of the vessel enhancement filter for CTA volume1 & Volume4 is demonstrated in Fig 5.19 & 5.20 respectively.

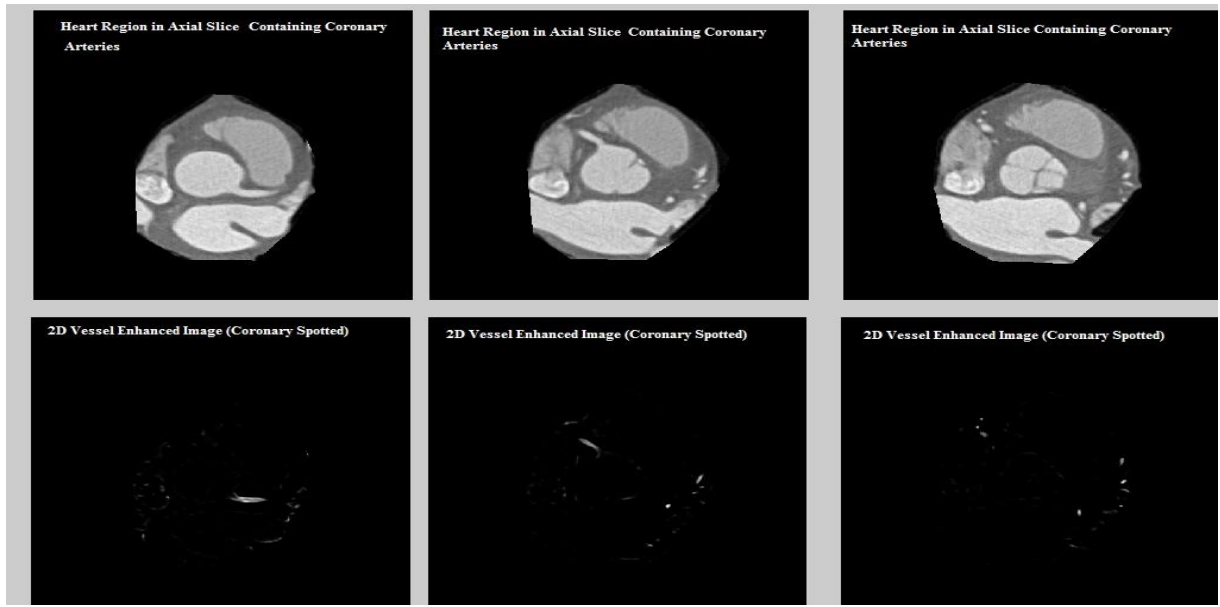

Fig.5.19 Coronary arteries spotted in different axial slices using vessel enhancement for CTA Volume1

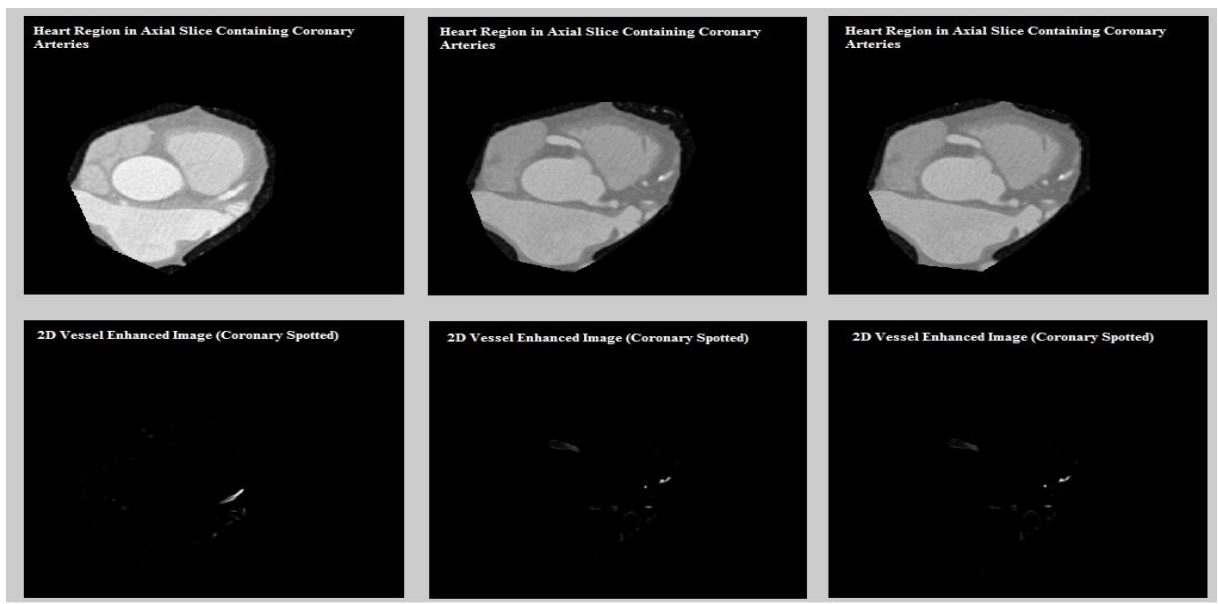

Fig.5.20 Coronary arteries spotted in different axial slices using vessel enhancement for CTA Volume4

In the vessel enhancement step all voxels of 3D volume are assigned vesselness measure according to their local geometric features. Results obtained after application of vessel enhancement filter on CTA volume 1, 2 and 3 are graphically presented in figure 5.21-5.23. Next stage is to derive HU based threshold value representing contrast agent diffusion behavior in respective CTA volume. Established threshold value will be used to retain only potential coronary voxels showing significance impact of contrast agent. In the final stage these candidates voxels will be passed to active contour model segmentation for precise boundary estimation of coronary arteries.

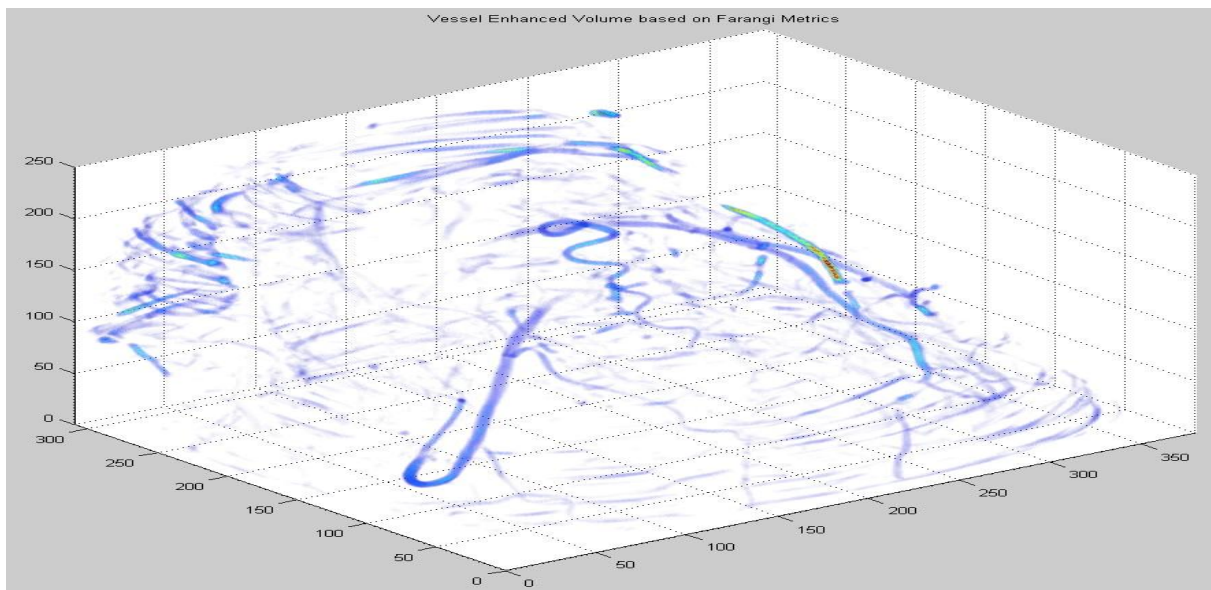

Fig.5.21 Construction of 3D vascular structures for CTA Volume1 after calculating vesselness

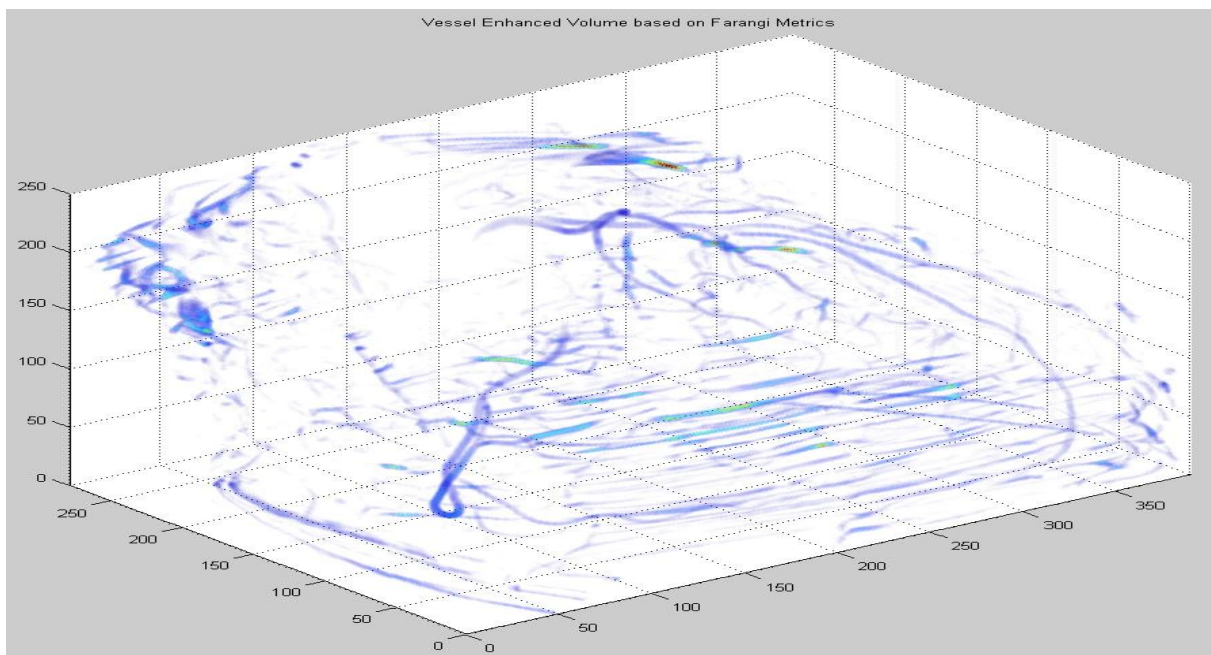

Fig.5.22 Construction of 3D vascular structures for CTA Volume4 after calculating vesselness

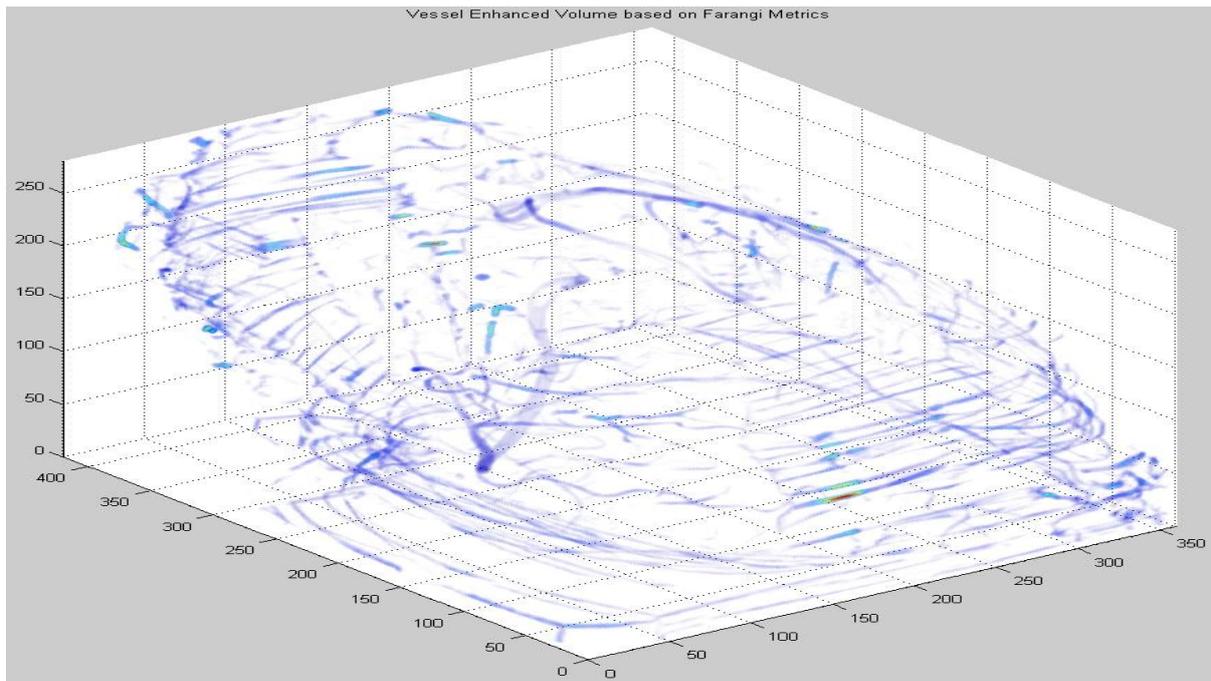

Fig.5.23 Construction of 3D vascular structures for CTA Volume8 after calculating vesselness

## 5.3 Blood Intensity Approximation for CTA volumes

Before CTA scan, a contrast medium is injected in the patient through intravenous passage. Contrast agents are used to improve pictures of internal organs of body in MRI & CT imaging. Often, contrast materials allow the radiologist to distinguish abnormalities in blood filled structures. Different CTA volumes (contrast enhanced) apparently exhibit similar visual behaviour as shown in Fig. 5.24 (coronaries appearing high-flying than the surrounding muscles), but statistical analysis of HU intensity values for blood voxels shows significant difference. After obtaining vesselness measures for 3D volumes, thresholding is performed to retain the strongest vessel candidate voxels. Local thresholding at this stage is performed on the basis of intensity values for respective contrast enhanced CTA image. Previously, a pre-defined HU cutoff value is chosen from literature & suppression of irrelevant voxels is accomplished. However, choosing a hard threshold to suppress non-blood voxels for all CTA volumes often leads to inaccurate results as fixed threshold lacks local image characteristics. Isgum *et al* [83] presented an automated system for calcification detection in aorta in CTA images where all connected components of intensity value greater than 220 HU were interpreted as potential calcified plaques. Hong *et al* [85] defined fixed threshold of 350HU for separation of calcified plaques in contrast enhanced CTA volumes. Accordingly images having strong concentration of dye will lead to exponential raise in false positives.

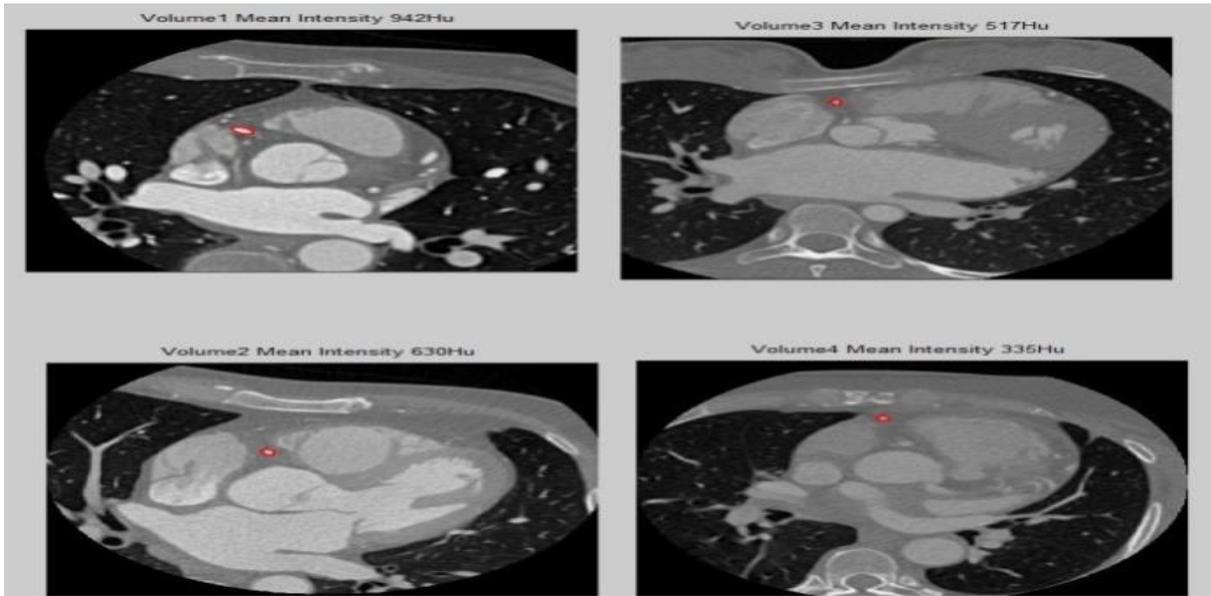

Fig.5.24 Similar Visual appearance of Different CTA images

To incorporate the local behavior of contrast agent in respective CTA volume, we propose an approximation method to establish optimal HU threshold value for accurate arterial tree. Based on the Bayesian probabilistic framework blood voxels are separated from muscular regions in first stage as suggested by Yin *et al* [86]. Followed with the application of Hough estimation aorta is detected slice-wise for isolation of aorta voxels. Intensity distribution of extracted aorta voxels is modeled with histogram & least square histogram fitting is performed to obtain Gaussian distribution parameters G($\mu,\sigma$ ) reflecting precisely contrast agent behavior for respective volume. Gaussian fitting for CTA volume 1, 4 & 10 is presented in figure 5.25-5.27.

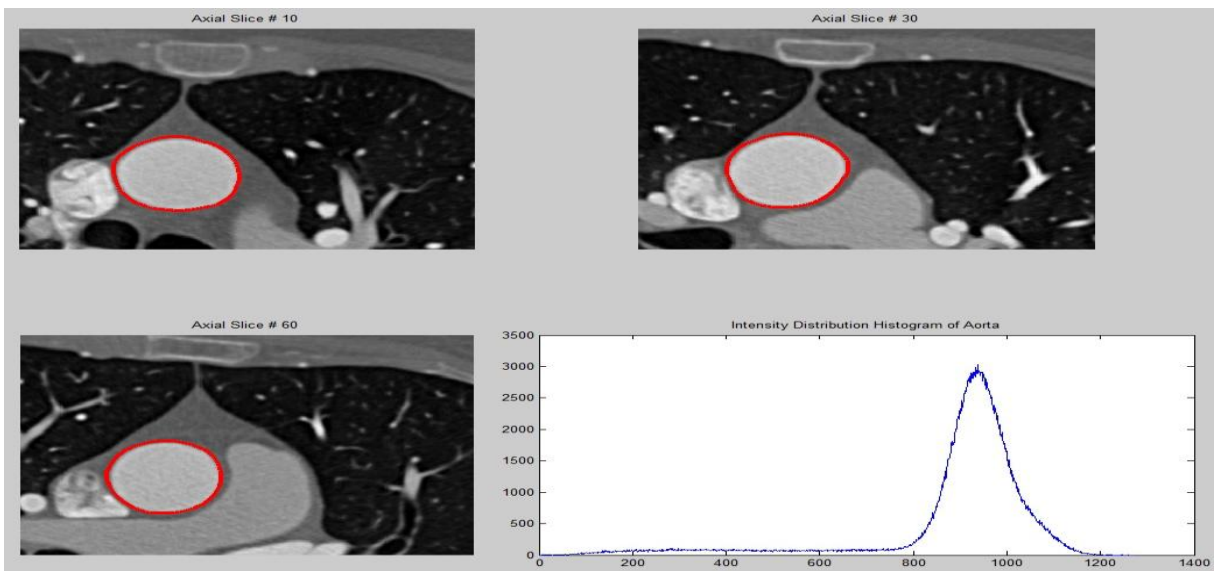

Fig.5.25 Aorta segmentation & intensity distribution histogram for CTA volume1

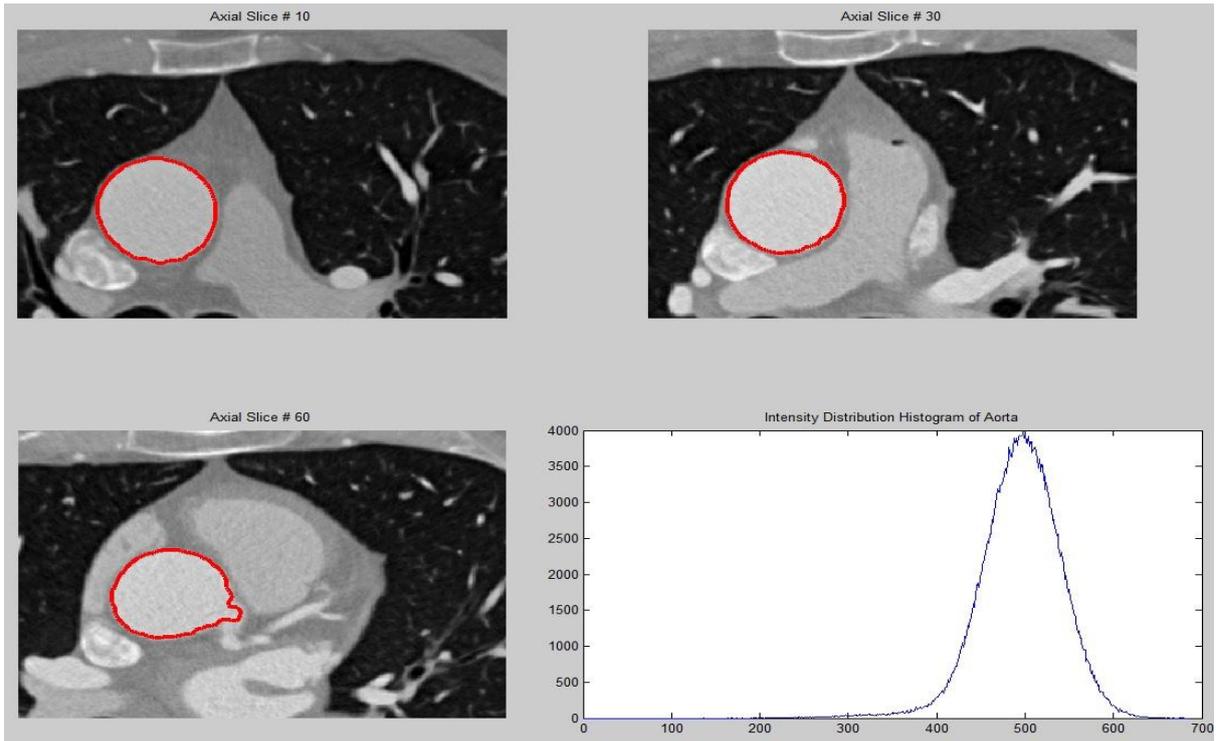

Fig.5.26 Aorta segmentation & intensity distribution histogram for CTA volume4

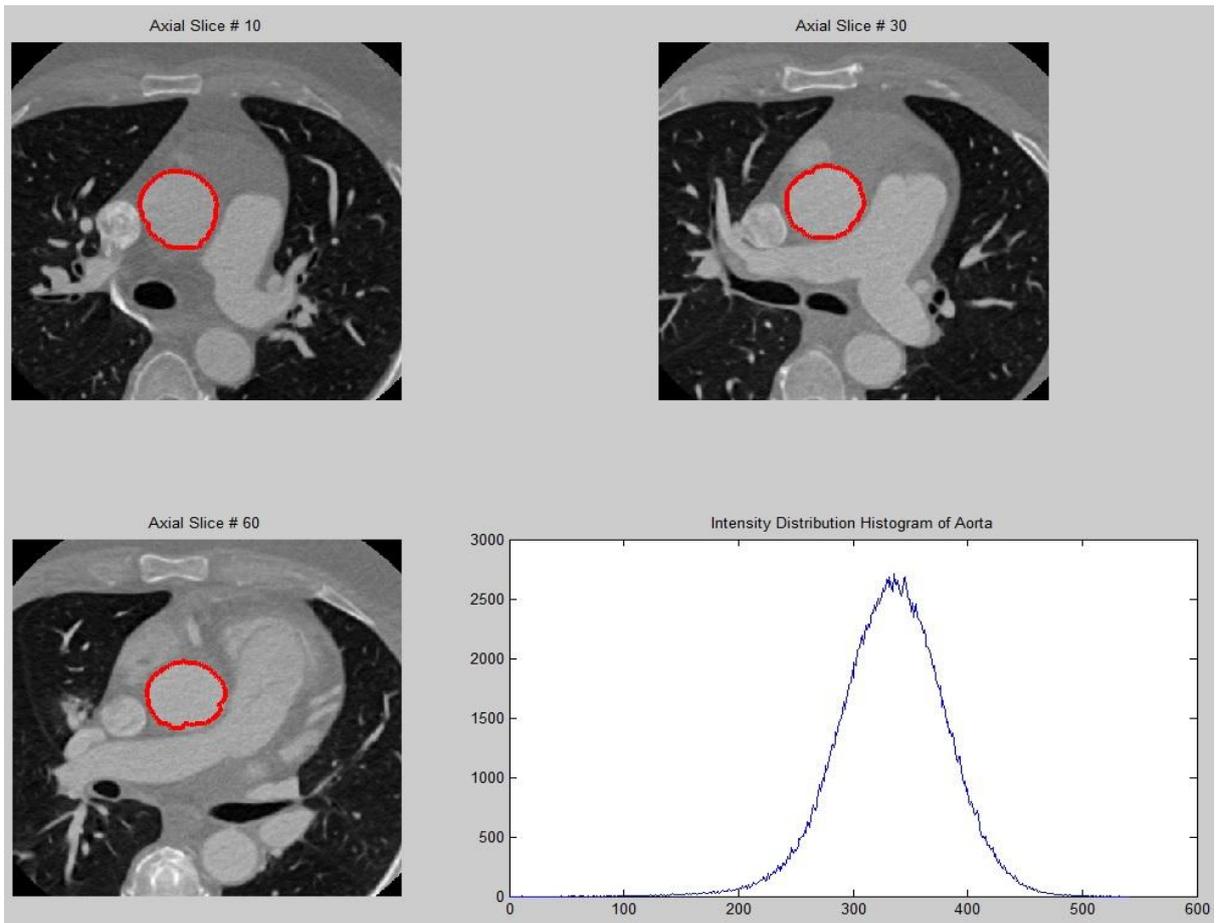

Fig.5.27 Aorta segmentation & intensity distribution histogram for CTA volume10

A significant variation in the HU intensity distribution parameters stimulates to choose different HU threshold value for every volume. Distribution parameters are used to obtain a valid HU range for accommodating uneven diffusion of the contrast agent through arterial coarse as defined by equation by (12);

Blood Voxel HU Intensity Range= [μ - 3σμ + 3σ]                    (5.6)

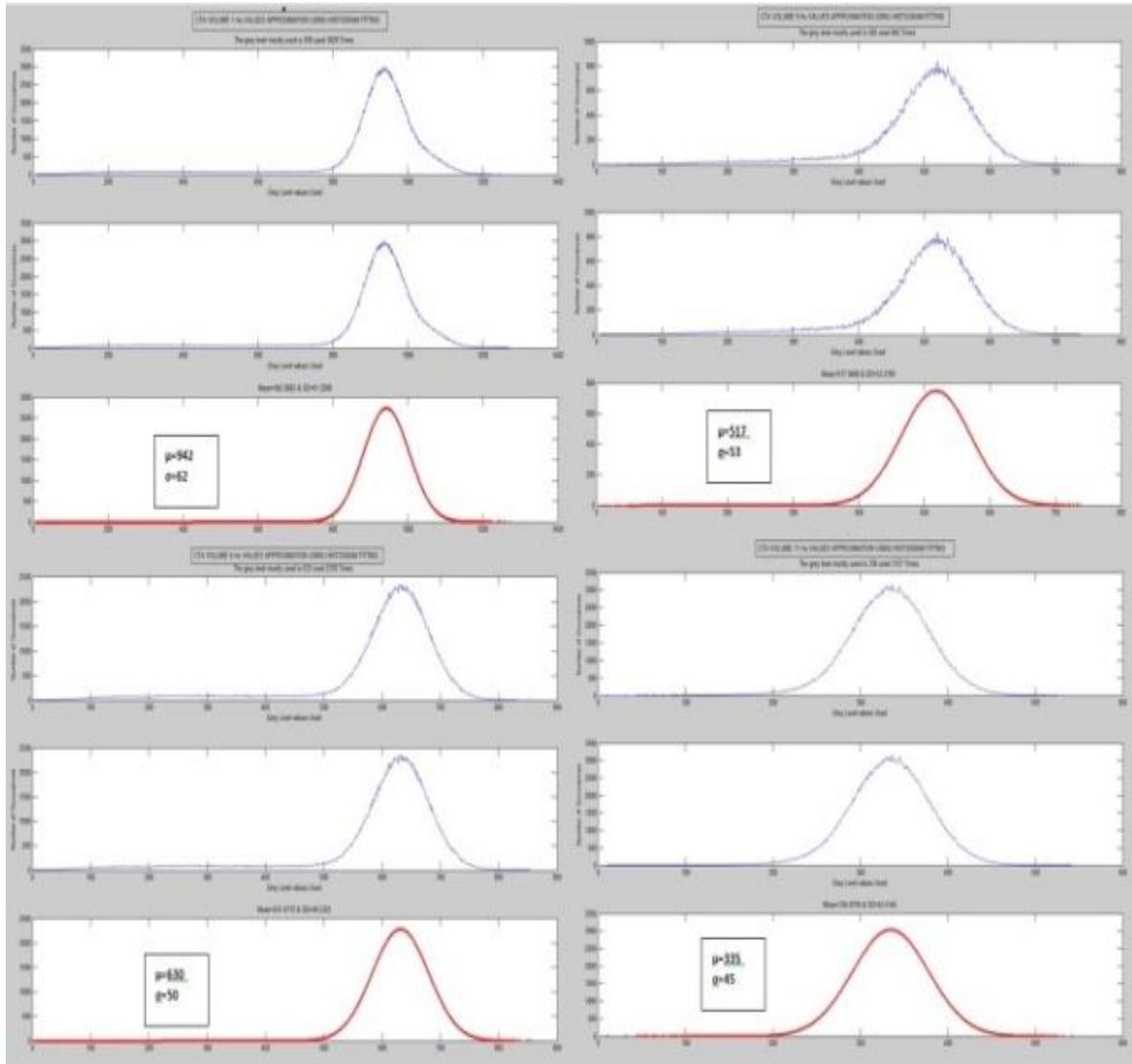

Fig.5.28 Histogram analysis & Gaussian fitting for contrast enhanced CTA volumes

Table 5.1 represents HU value range derived by Gaussian fitting of histogram for 12 different CTA volumes. The lower boundary is meant for suppressing all the voxels which does not have significant impact of contrast agent. As blood filled coronary arteries arise from aorta, the distribution remains valid for coronary arterial tree except few cases where unexpected drop of HU values is observed. This will be investigated for presence of soft / plaque in phase II.

Table 5.1. Allowed HU Range Vs Real Coronary Sampled Values

| CTA_VOLUME # | Histogram Fitting Parameters | | Hu Expected Range (μ +/- 3σ) | | Hu Obtained Values by Coronary Analysis Coronary Arteries Hu Recorded at 50 slices | | |
|---|---|---|---|---|---|---|---|
| | Mean | S.D | Min (μ -3σ) | Max (μ + 3σ) | Min obtained | Mean obtained | Max obtained |
| 01 | 942 | 62 | 756 | 1128 | 790 | 867 | 930 |
| 02 | 495 | 42 | 369 | 621 | 362 | 414 | 510 |
| 03 | 436 | 45 | 301 | 571 | 368 | 428 | 486 |
| 04 | 485 | 38 | 371 | 599 | 362 | 425 | 485 |
| 05 | 542 | 60 | 362 | 722 | 369 | 457 | 526 |
| 06 | 630 | 50 | 480 | 780 | 433 | 499 | 635 |
| 07 | 663 | 53 | 504 | 822 | 503 | 552 | 645 |
| 08 | 463 | 62 | 277 | 650 | 325 | 420 | 500 |
| 09 | 517 | 53 | 358 | 676 | 369 | 442 | 505 |
| 10 | 543 | 55 | 378 | 708 | 385 | 460 | 548 |
| 11 | 335 | 45 | 200 | 470 | 238 | 320 | 395 |
| 12 | 425 | 53 | 296 | 554 | 302 | 403 | 501 |

Figure 5.29 (a) shows segmented right coronary artery of CTA volume 1 based on predefined threshold value of 300HU. It contains several side branches in the way from aorta to distal endpoints whereas proposed histogram-generated-threshold based segmentation is shown in figure 5.29(b). It is clearly visible that progression of the right coronary artery is well tracked in this method & non coronary/weak side branches are eliminated. Effective validation of proposed technique is illustrated in the figure 5.30 (A – E). For investigating any missed periphery in proposed segmentation, curve planar reformation (CPR) is performed to visualize true structure of right coronary artery. Skeleton for right coronary artery is generated in first step using fast marching implementation of sub-voxel skeletonization algorithm. Figure 5.30 (A-B) shows extracted coronary surface & corresponding skeleton overlain with red dotted line.   Three different CPR images are constructed by re-sampling CTA volume as shown in figure 5.30 (C-E). Distinct views (collinear with x-axis, y-axis & z-axis respectively) are produced for validation of segmented coronary structure. Different CPR images confirm that RCA is well tracked from aorta to distal points by proposed histogram based segmentation. Potential side branches which appear to be a part of the coronary in figure 5.29 (a) are not coronaries indeed, but vascular structures in close proximity which are grabbed by active contour during evolution.

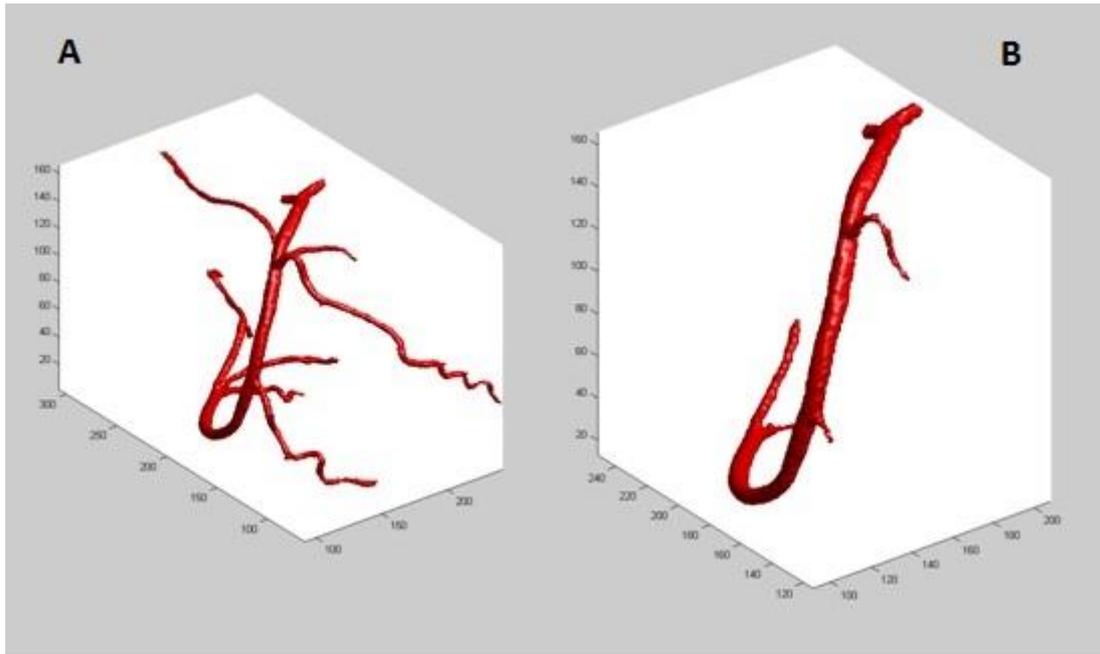

Fig.5.29 Right coronary artery (a) without using intensity distribution approximation (b) after using intensity distribution approximation

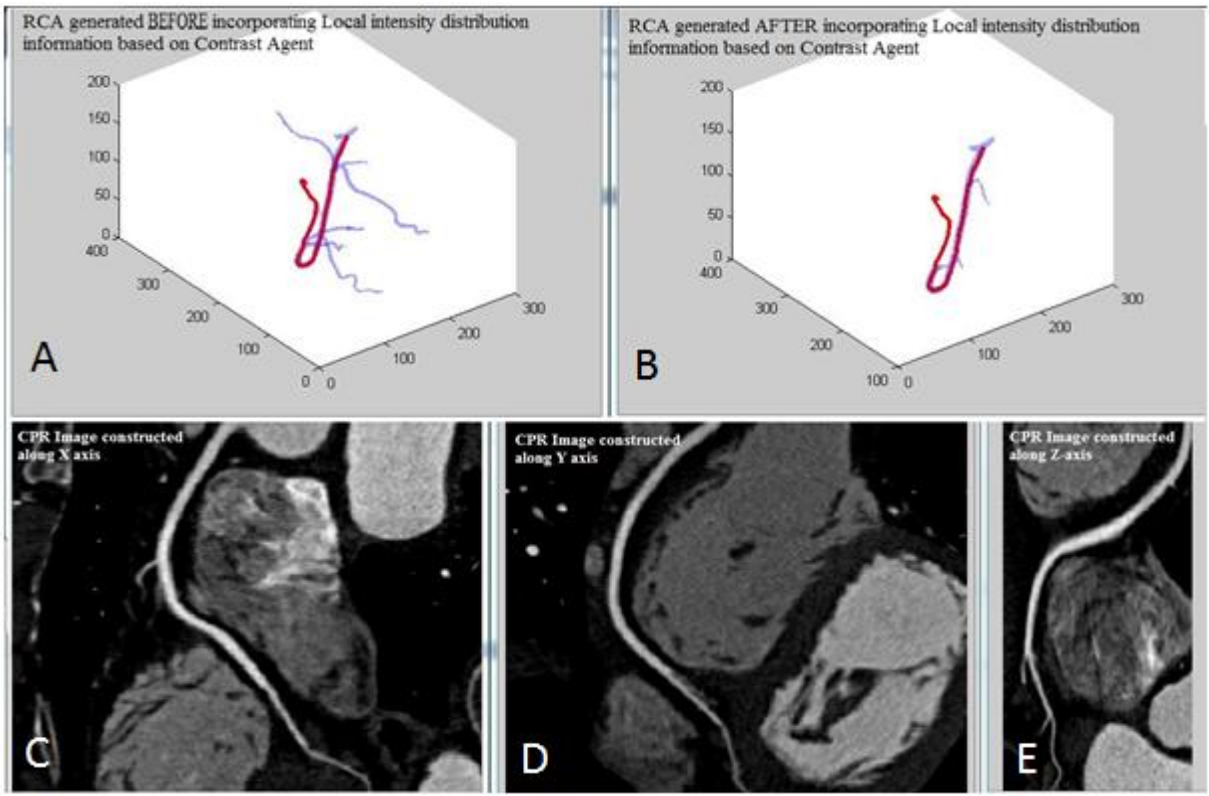

Fig.5.30 CPR visualization for Right coronary artery of CTA volume1 showing main progression of vessel

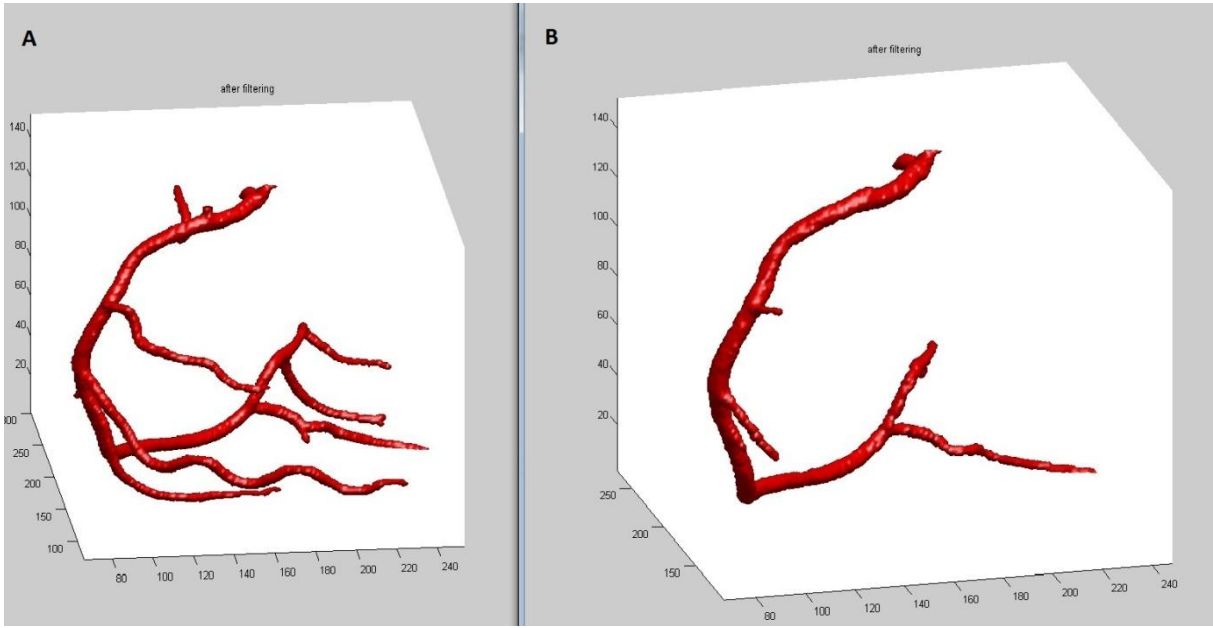

Fig.5.31 Right coronary artery (a) without using intensity distribution approximation (b) after using intensity distribution approximation

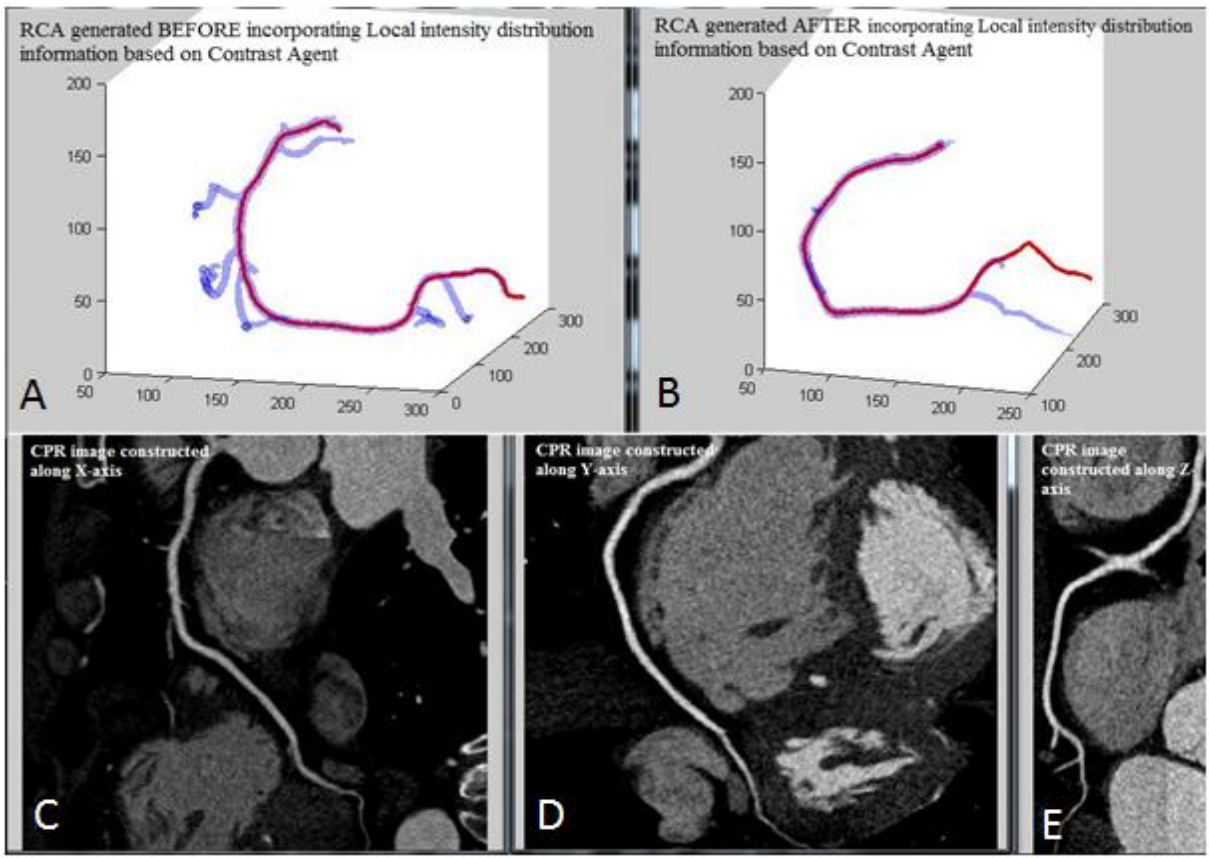

Fig.5.32 CPR visualization for Right coronary artery of CTA volume2 showing main progression of vessel

## 5.4 Arterial Tree Segmentation

In the vessel extraction stage, complete coronary tree is segmented based upon information obtained in previous stages (Automatic seed point plus Contrast agent behavior in terms of HU value). Detected seed point is used to initialize a constrained mask (curve) in vessel enhanced volume that evolves iteratively until the complete arterial tree is segmented. In contrast to edge based energy, contour evolution is based on region based energy. Region based models have shown more potential for coronary segmentation as it relies on regional intensity statistics to attain minimum energy representing the object boundary. Region based active contours proposed by Chan & Vese [11] and Yezzi *et al* [62] successfully detected objects with weak gradient in general but show poor performance for medical image segmentation due to the underlying assumption of intensity smoothness. Based on the seminal work of Mumford & Shah, universal modeling energy based active contour segmentation proposed by Chan & Vese is defined by Equation 5.7.

$$F(c1, c2, C) = \mu. length(C) + \int_{inside\,(C)} [I(x) - c1]2dx$$

$$+ \int_{outside\,(C)} [I(x) - c2]^2 dx \qquad (5.7)$$

Where C is the contour to be evolved, c1 & c2 represents interior & exterior average intensities. A robust framework for addressing in-homogeneity problem of medical images was proposed by Lancton *et al* [66] by introducing radius based ball kernel for selecting localized intensity statistics in a pre-defined neighborhood as expressed in (5.8). According to this approach, a restricted region is evaluated for attaining minimum energy instead of modeling whole image behavior. This focused information allows more accurate & precise segmentation realizing the local neighborhood behavior.

$$B(x, y) = \begin{cases} 1, & ||x - y|| < r \\ 0, & otherwise \end{cases} \qquad (5.8)$$

Kernel defining ball interacts every point of the contour, hence defines the local interior and exterior region around respective point as shown in figure 5.9. The integration of the localized information yield equation (5.9) that represents driving force for evolving contour.

$$E(\emptyset) = \int_{\pi x} \delta \emptyset(x) \int_{\pi y} B(x, y). F\big(I(y), \emptyset(y)\big) dy \, dx \qquad (5.9)$$

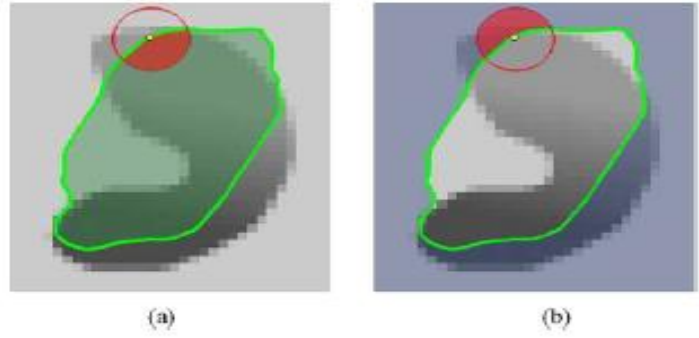

Figure 5.33 Localization kernel for curve points to yield local interior/exterior of curve

Variational level set method is used for implementation of active contour models to handles complex nature of coronary arteries. After a series of differential operations, evolution equation for contour can be written as Equation 5.10. Readers are referred to [66] for complete mathematical derivation.

$$\frac{\partial \emptyset}{\partial t}(x) = \delta \emptyset(x) \int_{\pi y} B(x,y). \nabla \emptyset(y). F\big(I(y), \emptyset(y)\big) dy + \alpha. \delta \emptyset(x) div \left( \frac{\nabla \emptyset(x)}{|\nabla \emptyset(x)|} \right)$$

(5.10)

Substituting the Chan & Vese region based energy into the generalized equation, the driving force of the contour becomes;

$$F = H\emptyset(y)(I(y) - c1)^2 + \big(1 - H\emptyset(y)\big)(I(y) - c2)^2$$

(5.11)

Where Heaviside function is used to differentiate interior & exterior of the curve. Finally level set evolution based on localized regional intensity statistics used in this work is defined by Equation 5.12.

$$\frac{\partial \emptyset}{\partial t}(x) = \delta \emptyset(x) \int_{\Omega y} B(x,y) \delta \emptyset(y). \big((I(y) - c1)^2 - (I(y) - c2)^2\big) dy + \lambda. \delta \emptyset(x) div \left( \frac{\nabla \emptyset(x)}{|\nabla \emptyset(x)|} \right)$$

(5.12)

Proposed approach is self-adjusting algorithm that reconstructs contour after every iteration to follow the arterial progression more precisely. Figure 5.34 shows auto adjustment capability of mask that allows grabbing side branches. In CTA images, arterial tree splits into branches as distance increases from aorta in terms of axial planes but in some cases, side branches originate away from the lumen and joins surface as slices are navigated. To grasp the potential side branches of coronary tree, algorithm reconstructs the mask after every iteration by scanning the neighborhood of the centerline in a restricted region. It captures all the prospective branches that satisfy geometric and intensity constraints as shown in Fig. 5.34 - 5.36. This self- adjustment feature of mask offers computational robustness without increasing processing load ensuring all side branches are captured. Non-connected components are discarded by applying morphological operations.

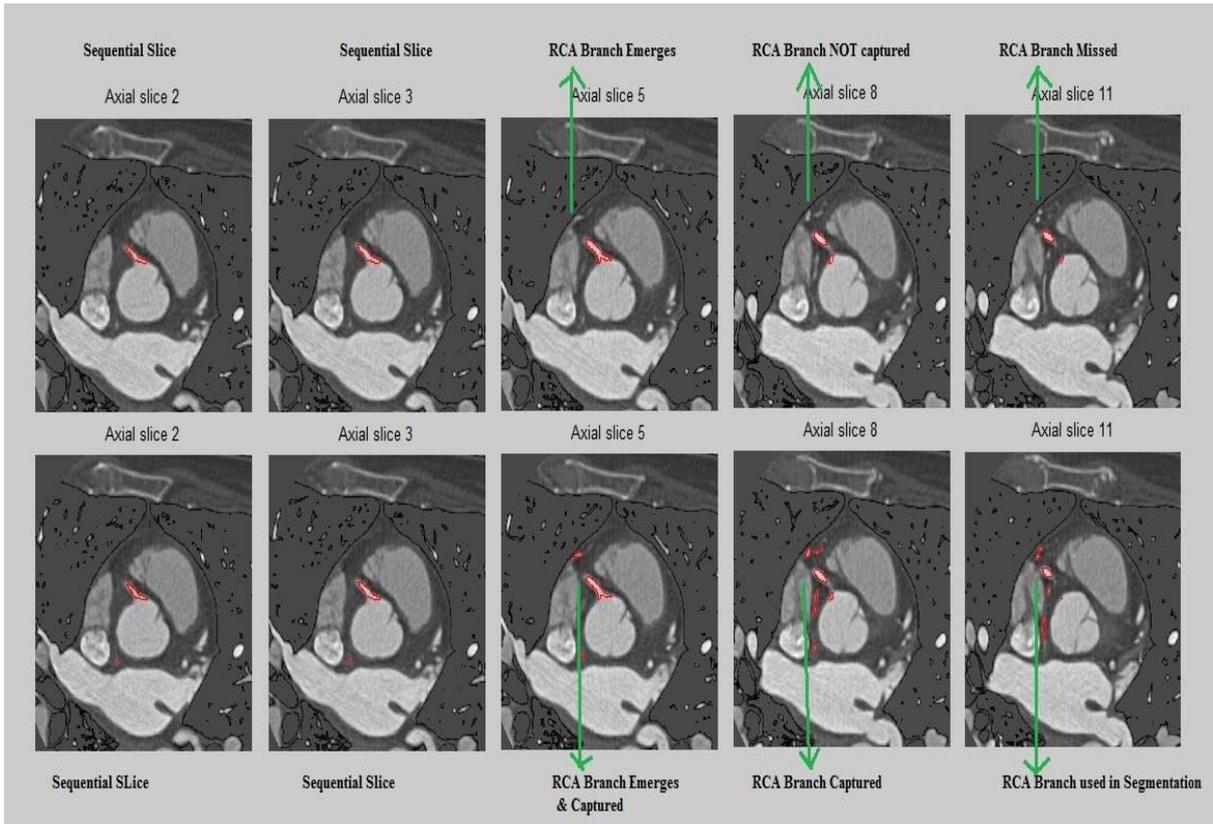

Fig.5.34 Auto capturing of Emerging side branches for CTA Volume 1

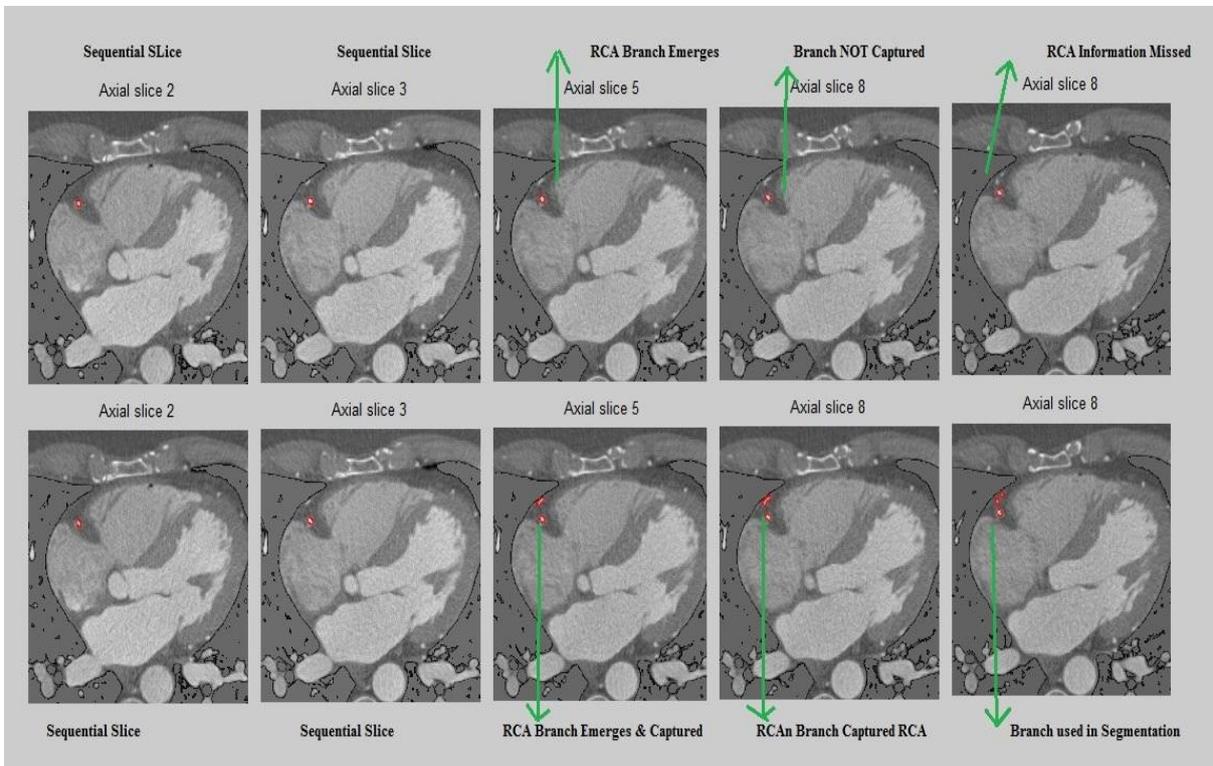

Fig.5.35 Auto capturing of Emerging side branches for CTA Volume 2

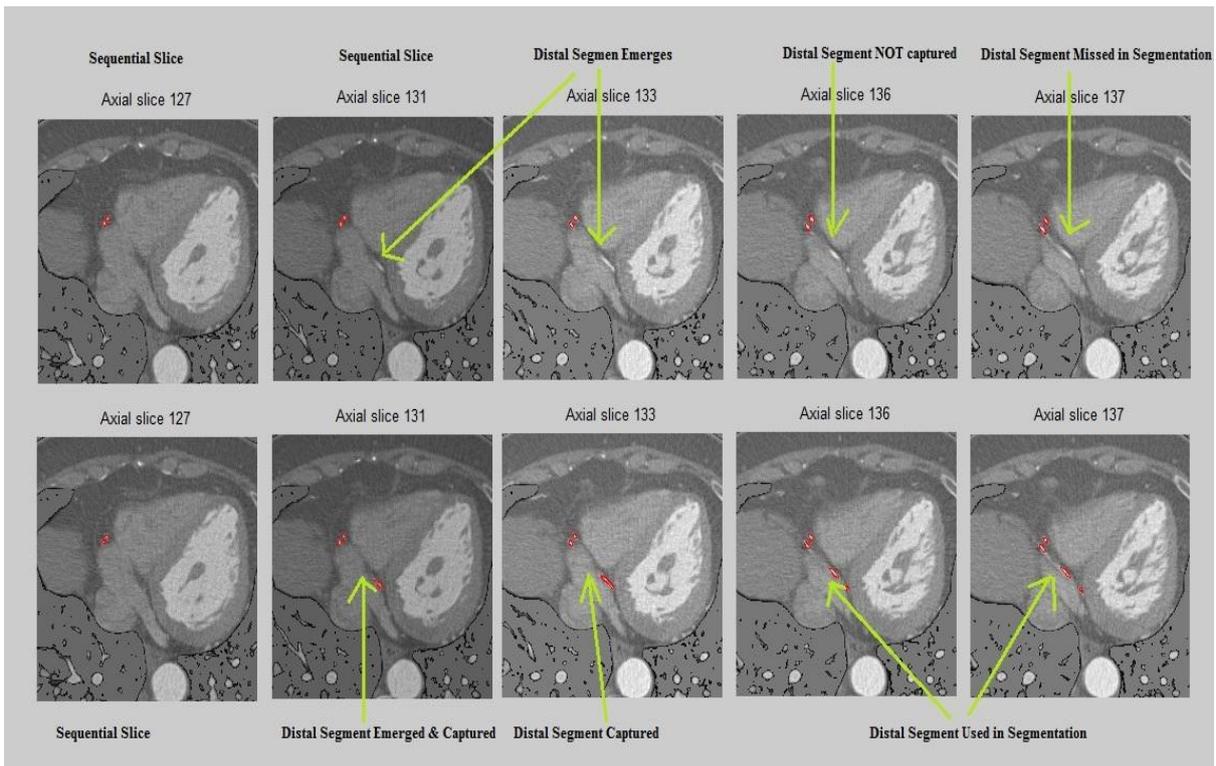

Fig.5.36 Auto capturing of Emerging side branches for CTA Volume 10

Loss of arterial information occurs depending upon the location of reference slice. If the reference slice is exactly axial cross section where respective coronary artery comes out of aorta, complete tree can be segmented efficiently. But due to the condition that same reference slice to be used for both seed points(Left & Right coronary artery) , usually middle of CTA volume is chosen as reference slice by putting value of Cr in range of [0.4, 0.6]. The resultant reference slice ensures to produce seed for both arteries, but arterial information exists in both directions (towards initial slices that contain aorta as well as towards distal end points of artery). This middle slice based seed point detection is apportioned with provision of backward segmentation mechanism that scans all axial slices in reverse direction till the artery joins aorta. The final segmented arterial tree is the logical combination of two directional segmentations, hence contains the complete information. Figure 5.37 represents segmentation in forward direction (red) & backward direction (green) after detecting seed point at axial slice number 66.

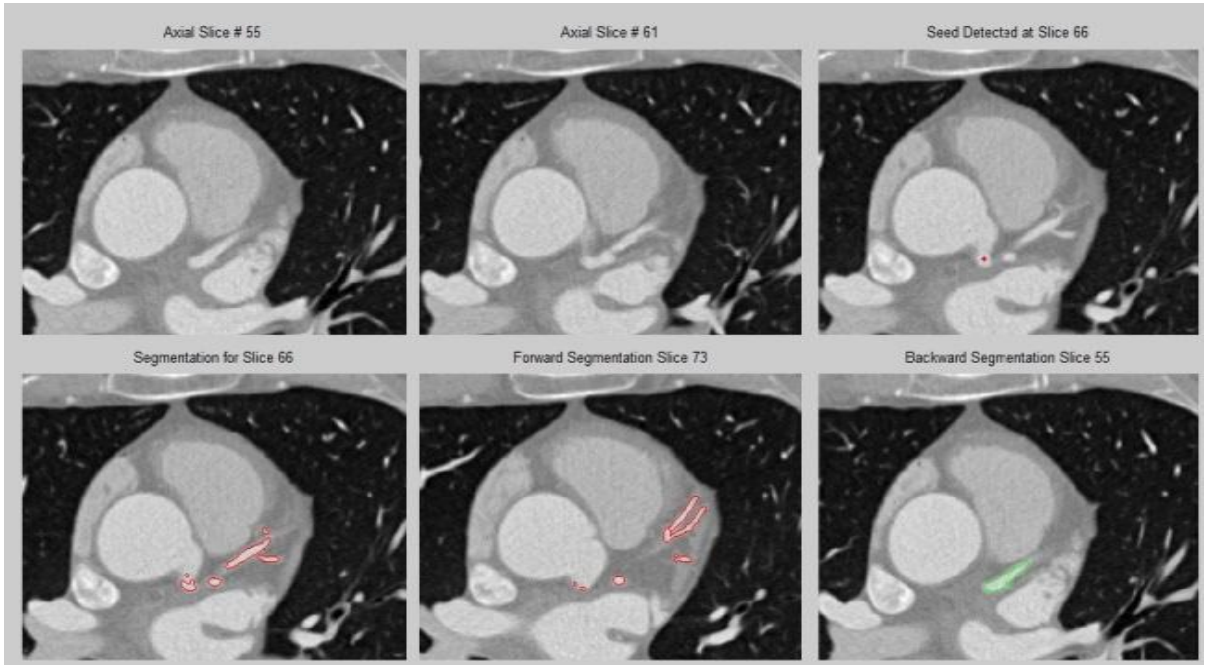

Figure 5.37 Illustration of two way segmentation  (A) Axial slice 55. (B) Axial slice 61 (C) Axial slice 66 containing seed. (D) Segmentation for axial slice 66 (E) Red Contour - Segmentation in Forward direction shown axial slice 73. (F) Green Contour - Segmentation in Backward Direction, shown axial slice 55

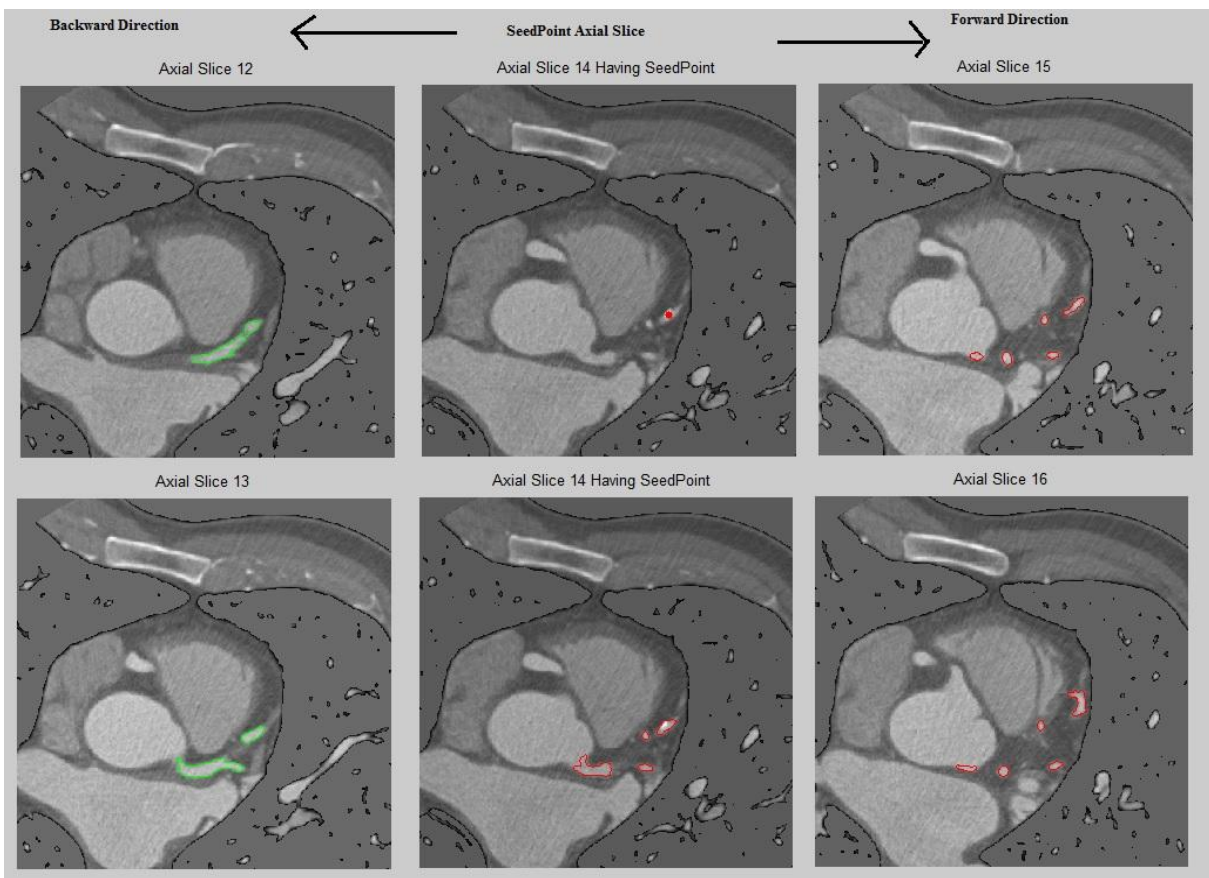

Figure 5.38 Illustration of two way segmentation (A) Axial slice 12 segmented. (B) Axial slice 13 segmented (C) Axial slice 14 containing seed. (D) Segmentation for axial slice 14 (E) Axial slice 15 segmented (F) Axial slice 16 segmented. Green contour represents backward segmentation, Red contour forward segmentation

## 5.5 Centreline extraction based on sub-voxel skeleton

For quantitative analysis of coronary arteries, medial axis generation is very important. After obtaining 3D coronary artery tree for different CTA volumes, centreline is established by using fast marching implementation of sub-voxel skeleton method as proposed in [85]. On the basis of centreline data, orthogonal cross sections will be obtained which represents the luminal behavior at a certain point. Figure 5.39 -5.40 shows centreline for CTA volume 1 & 2 respectively. Centreline points will help not only in orthogonal slice extraction but ray casting will also be based on skeleton points for evaluating local intensity & geometric properties in plaque detection method.

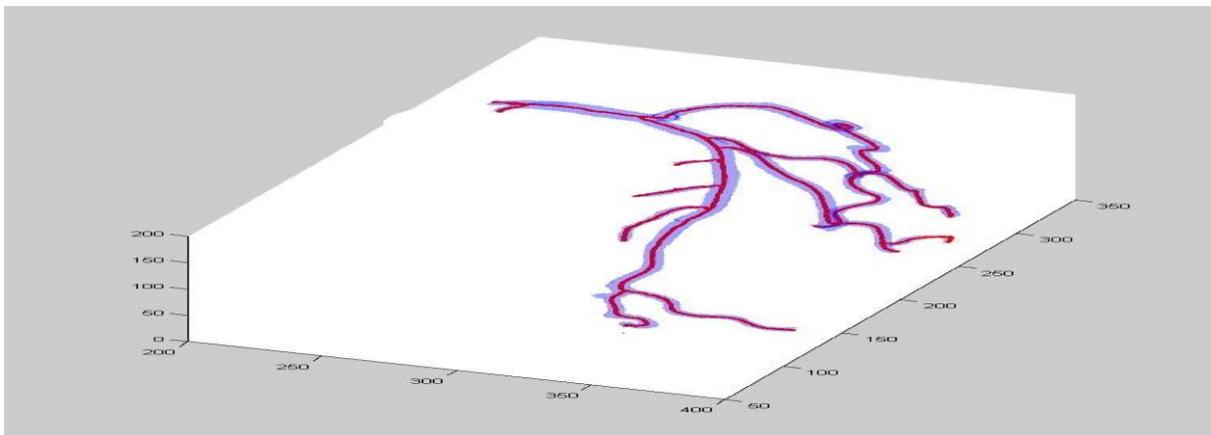

Fig.5.39 Centreline generated for left coronary artery of CTA volume1

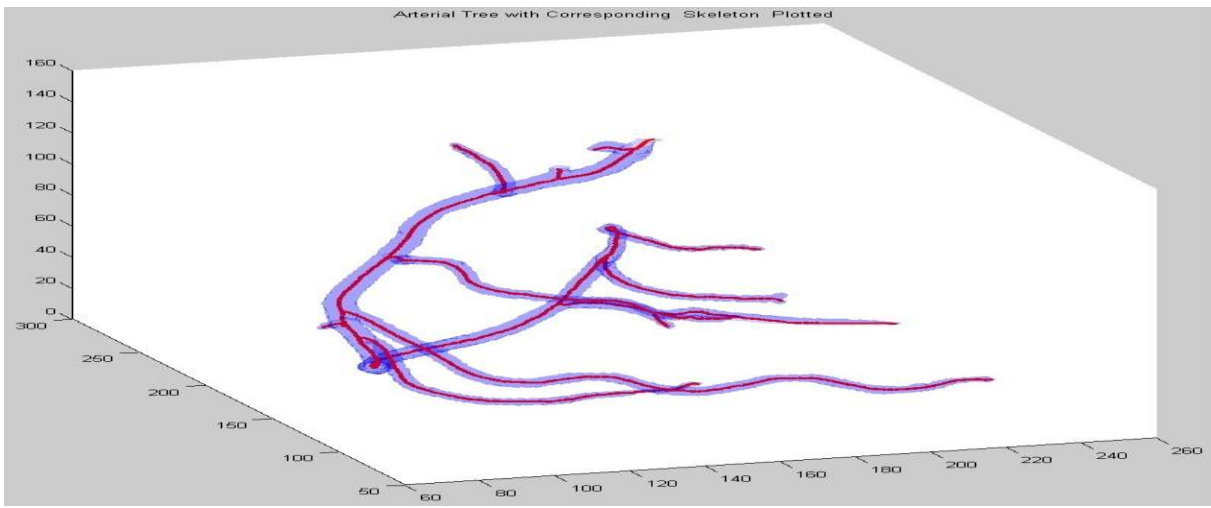

Fig.5.40 Centreline generated for right coronary artery of CTA volume2

**5.6  Conclusion & Future Work**

In this work we proposed an automated framework for extraction of coronary tree from 3-Dimensional CTA data cloud. Automatic seed detection process made this framework robust and less dependent on user however prior anatomical knowledge has been used that coronaries behaves as tubular structures in the mid axial slices of CTA volume. Contrast agent is usually injected in patient via intravenous passage before CTA examination as it ensures brighter appearance of blood filled vessels in image. Depending upon several factors including amount & type of medium as well as heart beat of patient, the diffusion rate of contrast agent is not homogenous for all cases. This non homogeneity is reflected in terms of voxel intensity in 3D volume. Active contour based segmentation is dependent upon initial seed point and the voxel intensity threshold that the impact of contrast agent for particular volume. Setting a hard threshold for all CTA images for differentiating coronaries from fatty muscles could be misleading as it lacks behavioural information of contrast agent in respective CTA volume. Usually under or over segmentation occurs due to the improper diffusion of contrast agent in different branches.

We proposed the integration of contrast medium information by approximating intensity histogram of segmented aorta. Based on the fact that coronary arteries originate from aorta it is expected that blood intensity for coronary can be derived by modelling intensity distribution of aorta. Aorta is isolated by performing geometrical shape based segmentation through initial axial slices and intensity histogram is obtained from segmented aorta. In the following stage intensity histogram is approximated by Gaussian fitting to obtain distribution parameters including mean & deviation values for contrast agent impact in CTA volume. A notable difference among mean values for different volumes emphasizes the need of establishing separate threshold value for respective volume. Obtained range is validated by equating intensity value of manually sampled coronary segments for each volume at 50 random points.

The proposed method has been tested on 12 CTA volumes and preliminary results are promising. Extracted tree validates the standard coronary model visually whereas statistical quantification of segmentation quality is in process. Automatic adjustment feature of mask ensure that all the emerging side branches are captured whereas bidirectional segmentation is used for extracting coronary information in both directions with respect to reference slice (seed point). This method alleviates the need of user interaction & prior knowledge in terms of defining initial mask or input seed points. In comparison with existing approaches the proposed methods requires least pre or post processing that make it feasible for physicians.

In second phase of this work extracted coronary tree will be evaluated for non-calcified plaques. Soft plaques generally lie inside coronary walls so focus of the second stage of work is vascular wall analysis in terms of detailed geometrical & intensity based metrics. Vessel remodelling analysis in context of non-calcified plaques is eventually topic of interest in future.

# References

[1] American Heart Association, "A.H.A. Heart Disease and Stroke Statistics - 2006 Update", (2006).

[2] W. G., Roger, V. L., Go, A. S., Lloyd-Jones et al, Heart disease and stroke statistics-2012 update. Circulation, 125(1):e2–e220.

[3] McMinn, R. M. H., Hutchings, R. T., Color Atlats of Human Anatomy, Year Book Medical Publishers, Inc. 35E. Wacker Drive Chicago, (1977)

[4] Netter, F. H., Atlas of Human Anatomy, Second Edition, Novartis, East Hanover, New Jersey (1999).

 [5] Jang, I-K, Bouma, B. E., Kang, D-H., Park, S-J., Park, S-W., Seung, K-B., Choi, K-B., Shishkov, M., Schlendorf, K., Pomerantsev, E., Houser, S. L., Aretz, H. T., Tearney, G. J., "Visualization of Coronary Atherosclerotic Plaques in Patients Using Optical Coherence Tomography: Comparison With Intravascular Ultrasound", Journal of American College of Cardiology, Vol. 39, No. 4, (2002) pp. 604-609

[6] Kass, M., Witkin, A., and Terzopoulos, D., "Snakes: Active Contour Models", International Journal of Computer Vision, (1988), pp. 321-331.

[7] Terzopoulos, D., Witkin, A., and Kass, M. 1988. Constraints on deformable models: Recovering 3D shape and nonrigid motions.Artificial Intelligence, 36:91–123

[8] Caselles, V., Catte, F., Coll, T., and Dibos, F. 1993. A geometric model for active contours.Numerische Mathematik , 66:1–31

[9] Malladi, R., Sethian, J. A., Vemuri, B. C., "Shape Modeling with Front Propagation: A Level Set Approach", IEEE Trans. Pattern Analysis and Machine Intelligence 17 (1995) pp. 158-174.

[10] Caselles, V., Kimmel, R., and Sapiro, G., "Geodesic Active Contours," Int. Journal Computer Vision, 22(1) (1997) pp. 61-79.

 [11] Chan, T.F. and Vese, L.A., "Active Contours Without Edges", IEEE Trans. Imag. Proc., Vol. 10, No. 2, (2001) pp.266-277.

[12] Kichenassamy, S., Kumar, A., Olver, P., Tannenbaum, A., and Yezzi, A., "Conformal Curvature Flows: From Phase Transitions to Active Vision," Archive of Rational Mechanics and Analysis, 134 (1996) pp. 275-301.

[13] Osher, S. and Fedkiw, R., Level Set Methods and Dynamic Implicit Surfaces, Springer-Verlag, 2003.

[14] Sethian, J., "Level Set Methods and Fast Marching Methods", Cambridge University Press (1999).

[15] Zeng, X., Staib, L. H., Schultz, R. T., and Duncan, J. S., "Segmentation and Measurement of the Cortex from 3-D MR Images Using Coupled-Surfaces Propagation", IEEE Trans. Med. Imag. 18 (1999) pp. 927-937.

[16] Yezzi, A., Kichenassamy, S., Kumar, A., Olver, P., and Tannenbaum, A., "A geometric Snake Model For Segmentation of Medical Imagery," IEEE Trans. on Medical Imaging, 16 (1997) pp. 199-209.

[17] Leventon, M. E., Grimson, W. E. L. and Faugeras, O., "Statistical Shape Influence in Geodesic Active Contours" Proc. Conf. Computer Vis. and Pattern Recog, (2000).

[18] McInerney, T. and Terzopoulos, D., "Deformable Models in Medical Image Analysis: A Survey" Medical Image Analysis 1 pp. 91-108.

[19] Baillard, C., Barillot, C., and Boutherny, P., "Robust Adaptive Segmentation of 3D Medical Images with Level Sets" INRIA, Rennes Cedex, France, Res. Rep. 1369, Nov. 2000.

[20] Pichon, E., Tannenbaum, A. and Kikinis, R., "A Statistically Based Flow for Image Segmentation", Medical Image Analysis, 8 (2004) pp. 267-274.

[21] Frangi, A. F., Niessen, W. J., Vincken, K. L. and Viergever, M. A., "Multi-scale Vessel Enhancement Filtering", MICCAI'98, LNCS 1496, (1998) pp. 130-137

[22] Krissian, K., "Flux-Based Anisotropic Diffusion Applied to Enhancement of 3-D Angiogram", IEEE Trans. Med. Imag., 21, (2002) pp.1440-1442

[23] Krissian, K., Malandain, G., Ayache, N., "Directional Anisotropic Diffusion Applied to Segmentation of Vessels in 3D Images", Scale-Space Theory in Computer Vision, LNCS 1252, (1997) pp. 345-348

[24] Gonzalez, R. C., Woods, R. E., "Digital Image Processing", Second Edition, Prentice Hall.

[25] Pal, N. R., Pal, S. K., "A Review on Image Segmentation Techniques", Pattern Recognition 26 (1993) pp. 716.

[26] Yang, Y., Tannenbaum, A., and Giddens, D., "Knowledge-Based 3D Segmentation and Reconstruction of Coronary Arteries Using CT Images", In Proceedings of the 26th Annual International Conference of the IEEE EMBS, (2004) pp. 1664-1666.

[27] Bezdek, J.C., Hall, L.O., and Clarke, L.P., "Review of MR Image Segmentation Techniques Using Pattern Recognition" Med. Phys. 20 (1993) pp.1033-1048.

[28] Clarke, L. P., Velthuizen, R. P., Camacho, M. A., Heine, J. J., Vaidyanathan, M., et al. "MRI Segmentation: Methods and Applications" Magn. Reson. Imaging 13 (1995) pp. 343-368.

[29] Pham, D. L., Xu, C., and Prince, J. L., "Current Methods in Medical Image Segmentation", Annu. Rev. Biomed. Eng. 2 (2000) pp. 315-337.

[30] Antiga, L., Ene-Iordache, B., and Remuzzi, A., "Computational Geometry for Patient-Specific Reconstruction and Meshing of Blood Vessels From MR and CT Angiography", IEEE Trans. Med. Imag., 22 (2003) pp. 674-684

[31] Andel, H.A.F.G., Meijering, E., Lugt, A., Vrooman, H.A., Stokking, R., "VAMPIRE: Improved Method for Automated Center Lumen Line Definition in Atherosclerotic Carotid Arteries in CTA Data", MICCAI'04, LNCS 3216, (2004) pp. 525-532.

[32] Scherl, H., Hornegger, J., Prummer, M., Lell, M., "Semi-Automatic Level-set Based Segmentation and Stenosis Quantification of the Internal Carotid Artery in 3D CTA Data Sets", Medical Image Analysis, In Press, (2006)

[33] Hernandez, M., Frangi, A. F., Sapiro, G., "Three-Dimensional Segmentation of Brain Aneurysms in CTA Using Non-parametric Region-Based Information and Implicit Deformable Models: Method and Evaluation", MICCAI 2003, LNCS 2879 pp. 594-602.

[34] Holtzman-Gazit, M., and Kimmel, R., "Segmentation of Thin Structures in Volumetric Medical Images", IEEE Trans. Imag. Proc., Vol. 15 (2006), pp. 354-363.

[35] Olabarriaga, S. D., Rouet, J-M., Fradkin, M., Breeuwer, M., Niessen, W. J., "Segmentation of Thrombus in Abdominal Aortic Aneurysms From CTA With Nonparametric Statistical Grey Level Appearance Modeling", IEEE Trans. Med. Imag., Vol. 24, No. 4, (2005), pp. 477-485

[36] Blondel, C., Malandain, G., Vaillant, R., and Anyche, N., "Reconstruction of Coronary Arteries From a Single Rotational X-Ray Projection Sequence", IEEE Trans. Med. Imag., Vol. 25, No. 5 (2006), pp. 653-663

[37] Chen, Z., Molloi, S., "Automatic 3D Vascular Tree Construction in CT Angiography", Computerized Medical Imaging and Graphics, 27, (2003) pp. 469-479.

[38] Szymczak, A., Stillman, A., Tannenbaum, A., Mischaikow, K., "Coronary Vessel Trees From 3D Imagery: A Topological Approach", Medical Image Analysis, 10 (2006) pp. 548-559.

[39] C. Lorenz, I.C. Garlsen, T.M. Buzug, C. Fassnacht and J. Weese, "Multi-scale line segmentation with automatic estimation of width, contrast and tangential direction in 2D and 3D medical images", *Lecture Notes in Computer Science*, Springer, Vol.1205, pp.233-242, 1997.

[40] Yoshinobu Sato, Shin Nakajima, Nobuyuki, Shiraga, et al, "3D Multi-scale Line Filter For Segmentation and Visualization of Curvilinear Structures in Medical Images", In *Medical Image Analysis*, Vol.2, No.2, pp.143-168, 1998.

[41] Bennink, H.E and Van-Assen.,H.C, "A novel 3D multi-scale lineness filter for vessel detection", In *Medical Image Computing and Computer-Assisted Intervention,* Vol.4792, Springer Berlin, 2007.

[42] Changhua Wu, Gady Agam and Peter Stanchev, "A hybrid filtering approach to retinal vessel segmentation", In *Biomedical Imaging: 4th IEEE International Symposium,* pp.604-607, 2007.

[43] Chuan Zhou, Heang-Ping Chan, Berkman Sahiner, "Automatic multiscale enhancement and segmentation of pulmonary vessels in CT pulmonary angiography images for CAD applications", *Med. Phys.* Vol. 34(12), 2007.

[44] Qiang Li, Shusuke Sone and Kunio Doi, "Selective enhancement filters for nodules, vessels and airway walls in two and three dimensional CT scans", In *Proceedings of the 18th International Congress and Exhibition, Computer Assisted Radiology and Surgery*, pp.929-934, 2004.

[45] Boskamp, T., Rinck, Link, F., Kmmerlen, B., Stamm, G., Mildenberger, P., "New vessel analysis tool for morphometric quantification and visualization of vessels in CT and MR imaging datasets", *Radiographics,* Vol.24, pp.287-297, 2004.

[46] Yi, J. 2003, "A locally adaptive region growing algorithm for vascular segmentation", *International Journal of Imaging Systems and Technology,* Vol. 13 (4), pp. 208-214.

[47] Tschirren, J. Hoffman, E.A., Mclennan, G., Sonka, M., "Intrathoracic airway trees: segmentation and airway morphology analysis from low dose CT scans", *IEEE Transactions on Medical Imaging*, Vol.24, pp.1529-1539, 2005.

[48] Metz, C., Schaap, M., Van Der Giessen, A., Van Walsum, T., Niessen, W., "Semi-automatic coronary artery centerline extraction in computed tomography angiography data", *4th IEEE International Symposium on In Biomedical Imaging,* pp. 856-859.

[49] M. de Bruijine, B. Van Ginneken, "Adapting active shape models for 3D segmentation of tubular structures in medical images", *Information Processing in Medical Imaging,* pp.136-147, 2003.

[50] Jun Feng and Horace H.S.Ip, "A statistical assembled model for segmentation of entire 3D vasculature", *18th International Conference on Pattern Recognition*, pp.95-98, 2006.

[51] Florez Valencia, L , Montagnat,J. and Orkisz,M "3D models for vascular lumen segmentation in MRA images and for artery-stenting simulation", *IRBM Elsevier,* Vol.28(2), pp. 65-71, 2007.

[52] Yim, P. J., Cebral, J. J., Mullick, R. M., Marcos, H. B., and Choyke, P. L., "Vessel Surface Reconstruction With a Tubular Deformable Model", IEEE Trans. Med. Imag. 20, (2001) pp. 1411-1421

[53] Worz, S., and Rohr, K., "A New 3D Parametric Intensity Model for Accurate Segmentation and Quantification of Human Vessels", MICCAI 2004, LNCS 3216, (2004) pp. 491-499.

[54] Olabarriaga, S. D., Breeuwer, M., Niessen, W. J., "Minimum Cost Path Algorithm for Coronary Artery Central Axis Tracking in CT Images," MICCAI 2003, LNCS 2879 pp. 687-694.

[55] Avants, B.B, Williams, J.P., "An Adaptive Minimal Path Generation Techniques for Vessel Tracking in CTA/CE-MRA Volume Images", MICCAI'00 LNCS 1935, (2000), pp. 707-716.

[56] Li, H., Yezzi, A., "Vessels as 4D Curves: Global Minimal 4D Paths to 3D Tubular Structure Extraction", IEEE Computer Society Workshop on Mathematical Methods in Biomedical Image Analysis (MMBIA'06), (2006)

[57] Felkel, P., Wegenkittl, R., Kanitsar, A., "Vessel Tracking in Peripheral CTA Datasets - An Overview", Spring Conference on Computer Graphics - SCCCG'01, (2001) pp. 232-239.

[58] Wink, O., Niessen, W. J., and Viergever, M. A., "Fast Delineation and Visualization of Vessels in 3-D Angiographic Image", IEEE Trans. Med. Imag, 19, (2000) pp. 337-346

[59] Whitaker, R. T. and Breen, D. E., "Level-set Models For the Deformation of Solid Objects", Proc. 3rd Intl. workshop on Implicit Surfaces, Eurographics Association, (1998), pp. 19-35.

[60] Chen, J. and Amini, A. A., "Quantifying 3-D Vascular Structures in MRA Images Using Hybrid PDE and Geometric Deformable Models", IEEE Trans. Med. Imag., Vol. 23, No. 10, (2004) pp. 1251-1262

[61] Nain, D., Yezzi, A., and Turk, G., "Vessel Segmentation Using a Shape Driven Flow", MICCAI 2004, LNCS 3216, (2004) pp. 51-59.

[62] Zhu, S. C. and Yuille, A., "Region Competition: Unifying Snakes, Region Growing, and Bayes/MDL for Multi-band Image Segmentation", IEEE Trans. Pattern Analysis and Machine Intelligence, Vol. 18, No. 9, (1996) pp. 884-900

[63] Yezzi, A., Tsai, A, and Willsky, A, "A statistical approach to snakes for bimodal and trimodal imagery, in *Proc. Int. Conf. Comput. Vis.,* 1999.

[64] Rousson, M, and Deriche, R, "A variational framework for active and adaptive segmentation of vector valued images", in *Proc. Workshop Motion Vid. Computi,* pp.56, 2002.

[65] Chunming Li  Chiu-Yen Kao  Gore, J.C.  Zhaohua Ding," Implicit Active Contours Driven by Local Binary Fitting Energy, IEEE conference on Computer Vision and Pattern Recognition**,** pp.1-7, 2007.

[66] Lankton S. and Tannenbaum A**.,** "Localizing Region-Based Active Contours", IEEE Transactions on Image Processing, Vol.17, pp. 2029-2039, 2008.

[67] S. Voros, S. Rinehart, Z. Qian et al., "Coronary atherosclerosis imaging by coronary CT angiography: current status, correlation with intravascular interrogation and meta-analysis," *Journal of the American College of Cardiology*, vol. 4, no. 5, pp. 537– 548, 2011.

[68] Saur, S., Alkadhi, H., Desbiolles, L., Szekely, G., Cattin, P.: Automatic detection of calcified coronary plaques in computed tomography data sets. In: Proceedings of Medical Imagingand Computing and Computer Assisted Intervention. (2008) 170–177

[69] Brunner, G., Kurkure, U., Chittajallu, D., Yalamanchili, R., Kakadiaris, I.: Toward unsupervisedclassification of calcified arterial lesions. In: Proceedings of Medical Imaging andComputing and Computer Assisted Intervention. (2008) 144–152

[70] Virmani, R., Burke, A., Farb, A., Kolodgie, F.: Pathology of the vulnerable plaque. Journal of the American College of Cardiology 47(8-C) (2006) 13–18

[71] Yang, G, Bousse, A, Toumoulin, C, Shu, H, "A multiscale tracking algorithm for the coronary extraction in MSCT angiography", *Proc. of 28th IEEE EMBS Annual Inter Conf,*2006.

[72] Lankton,S., Nain, D., Yezzi, A., and Tannenbaum, A., "Hybrid Geodesic Region- Based Curve Evolutions for Image Segmentation". In SPIE Medical Imaging, Vol. 6510, 2007.

[73] Zhou, C., Chan, H-P.,Chughtai, A., Patel, S. ,Kazerooni, A., "Automated coronary artery tree extraction in coronaryCT angiography using a multiscale enhancement and dynamic balloontracking (MSCAR-DBT) method," Comput. Med. Imaging Graphics **36**,1–10 (2012).

[74] Kubisch,C., Glaer, S., Neugebauer, M., and Preim.B., "Vessel visualization with volume rendering" , *Visualization in Medicine and Life Sciences II,* Mathematics and Visualization, pages 109-134. Springer Berlin Heidelberg, 2012.

[75] Hennemuth A., BoskampT., Fritz D.,et al.: One-click coronary tree segmentation in CT angiographic images. CARS'05: Computer Assisted Radiology and Surgery 1281, (2005), 317–321.

[76] Hong C.,Becker C. R.,Schoepf U. J., et al.: Coronary Artery Calcium: Absolute Quantification in Non-enhanced and Contrastenhaced Multi-Detector Row CT Studies. Radiology 223, (2002), 474–480.

[77] Agatston A. S., Janowitz W. R., HIildner F., Zusmer N. R., Viamonte M. J., Detrano R.: "Quantification of coronary artery calcium using ultrafast computed tomography". Journal of American College of Cardiology 15, 4 (1990), 827–832.

[78] Brodoefel,H., Burgstahler, C., Sabir, A., et al., "Coronary plaque quantification by voxel analysis: dual-source MDCT angiogra- phy versus intravascular sonography," The American Journal of Roentgenology,vol.192,no. 3, pp.W84–W89,2009.

[79] Sun,J., Zhang,Z., Lu, B., et al., "Identification and quantifica- tion of coronary atherosclerotic plaques: a comparison of 64- MDCT and intravascular ultrasound," The American Journal of Roentgenology,vol.190,no. 3, pp.748–754,2008.

[80] Lesage, D., Angelini, E.D., Bloch, I., Funka-Lea, G.A review of 3D vessel lumen segmentation techniques: Models, features and extraction schemes (2009) Medical Image Analysis, 13 (6), pp. 819-845.

[81] Renard, F.; Yongyi Yang, "Coronary artery extraction and analysis for detection of soft plaques in MDCT images," *Image Processing, 2008. ICIP 2008. 15th IEEE International Conference on* , vol., no., pp.2248,2251, 12-15 Oct. 2008.

[82] Yezzi, A., "A Fully Global Approach to Image Segmentation via Coupled Curve Evolution Equations", Journal of Visual Communication an Image Representation, 13, (2002) pp. 195-216

[83] Isgum, I., van Ginneken and Olree M., "Automatic detection of calcification in the aorta from CT scans of the abdomen", *Academic Radiology,* Vol.11, pp.247-257, 2004.

[84] Hong C.,Becker C. R., Schoepf U. J., *et al.*: Coronary Artery Calcium: Absolute Quantification in Nonenhanced and Contrastenhanced Multi-Detector Row CT Studies. *Radiology 223*, (2002), 474–480.

[85]. Uitert R,Bitter I, "Subvoxel precise skeletons of volumetric data based on fast marching methods" Med Phys. 2007 Feb;34(2):627-38.

[86] G. Yang, P. Kitslaar, M. Frenay, A. Broersen, M.J. Boogers, J.J.Bax, et al., Automatic centerline extraction of coronaryarteries in coronary computed tomographic angiography,Int. J. Cardiovasc. Imaging 28 (April) (2012) 921–933.

[87] Han D, Doan NT, Shim H, JeonB, Lee H, Hong Y, Chang HJ, "A fast seed detection using local geometric feature for automatic tracking of coronary arteries in CTA", Comput Methods Programs Biomed. 2014 Nov;117(2):179-88. doi: 10.1016/j.cmpb.2014.07.005. Epub 2014 Jul 24.

[88] Wang Y., "Blood Vessel Segmentation and Shape Analysis forQuantification of Coronary Artery Stenosis in CTAngiography" (Ph.D. thesis), School of Engineering &Mathematical Sciences, City University London, 2011.

[89] Tannenbaum, A., "Three Snippets of Curve Evolution Theory in Computer Vision", Mathematical and computer modelling, 24, (1996), pp. 103-119

[90] Lankton, S., Stillman, A, Raggi P, and Tannenbaum A, "Soft plaque detection and automatic vessel segmentation", in Proc. Int. Conf. Med. Image Comput. Comput. Assist. Intervent. (MICCAI) Workshop: Probabilist. Models Med. Image Anal., London, U.K., Sep. 2009, pp. 1–9.

[91] Arumuganainar Ponnapan, "Automatic Soft Plaque Detection From CTA" , MS Thesis, Department of Biomedical Engineering, Georgia Institute of Technology. https://smartech.gatech.edu/bitstream/handle/1853/26690/ arumuganainar_ponnappan_200808_ms.pdf?sequence=1.

[92] Wei, J., Zhou, C., Chan, H.-P., Chughtai, A., Agarwal, P., Kuriakose, J., … Kazerooni, E. (2014), "Computerized detection of noncalcified plaques in coronary CT angiography: Evaluation of topological soft gradient prescreening method and luminal analysis". Medical Physics, 41(8), 081901. http://doi.org/10.1118/1.4885958.

[93] Glaber S., Oeltze S, Kubisch C et al, "Automatic Transfer Function Specifcaton For Visual Emphasis Coronary Artery Plaque", Computer Graphics Forum, 2010; 29(1):191 -201.

[94] Mirunalini, P.; Aravindan, C., "Automatic segmentation of coronary arteries and detection of stenosis," in TENCON 2013 - 2013 IEEE Region 10 Conference (31194) , vol., no., pp.1-4, 22-25 Oct. 2013. doi: 10.1109/TENCON.2013.6719010.

[95] Clouse, M, Sabir,A, Yam,C-S, Yoshimura,N, Lin, S, Welty, F, Martinez-Clark,P,  and Raptopoulos,V, "Measuring Noncalcified Coronary Atherosclerotic Plaque Using Voxel Analysis with MDCT Angiography: A Pilot Clinical Study",  American Journal of Roentgenology 2008, 190:6, 1553-1560.

[96] Ying, Li, Chen,W, Liu,K, et al., "A Voxel-Map Quantitative Analysis Approach for Atherosclerotic Noncalcified Plaques of the Coronary Artery Tree," Computational and Mathematical Methods in Medicine, vol. 2013, Article ID 957195, 9 pages, 2013. doi:10.1155/2013/957195.

[97] Mazinani, M.; Dehmeshki, J.; Hosseini, R.; Ellis, T.; Qanadli, S.D., "Automatic Segmentation of Soft Plaque by Modeling the Partial Volume Problem in the Coronary Artery," in Digital Society, 2010. ICDS '10. Fourth International Conference on , vol., no., pp.274-278, 16-16 Feb. 2010. doi: 10.1109/ICDS.2010.61.

**ANNEX-A (Segmented Coronary Tree for CTA Volumes)**

**Segmented Arteries for CTA Volume1  RCA,  LCA and Combined Arterial Structures**

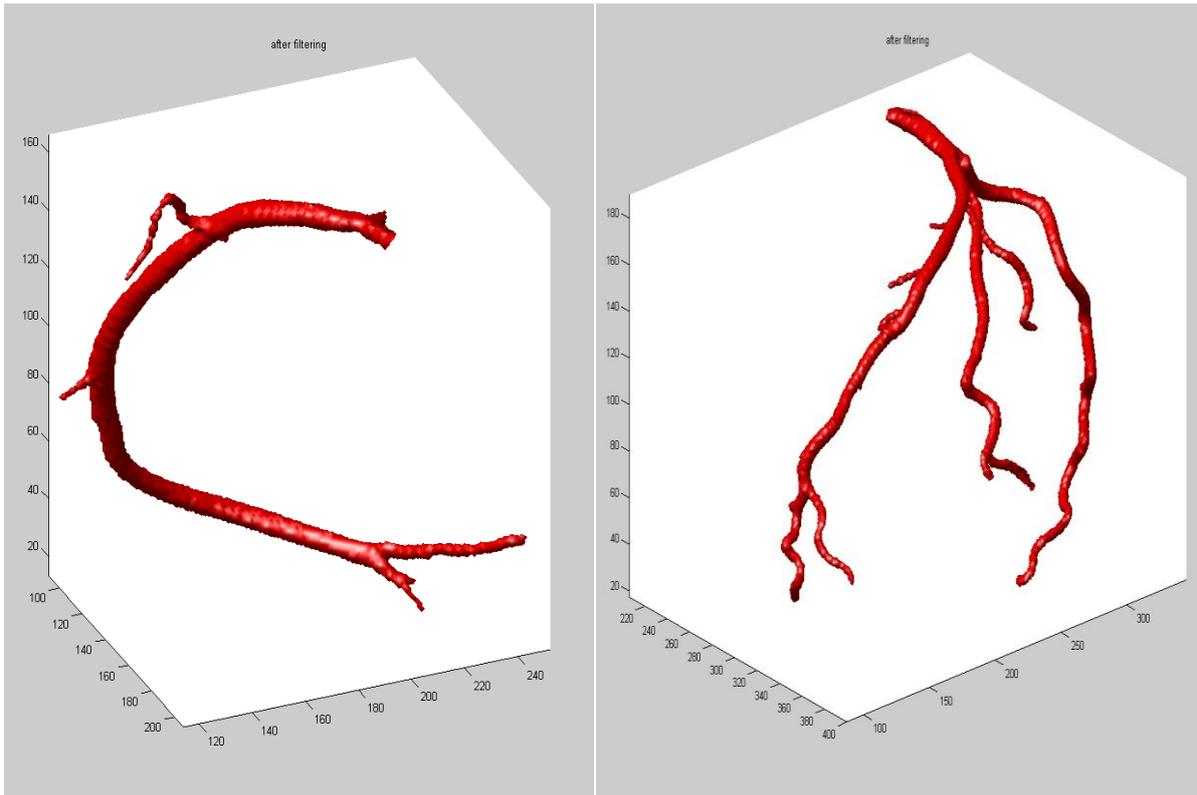

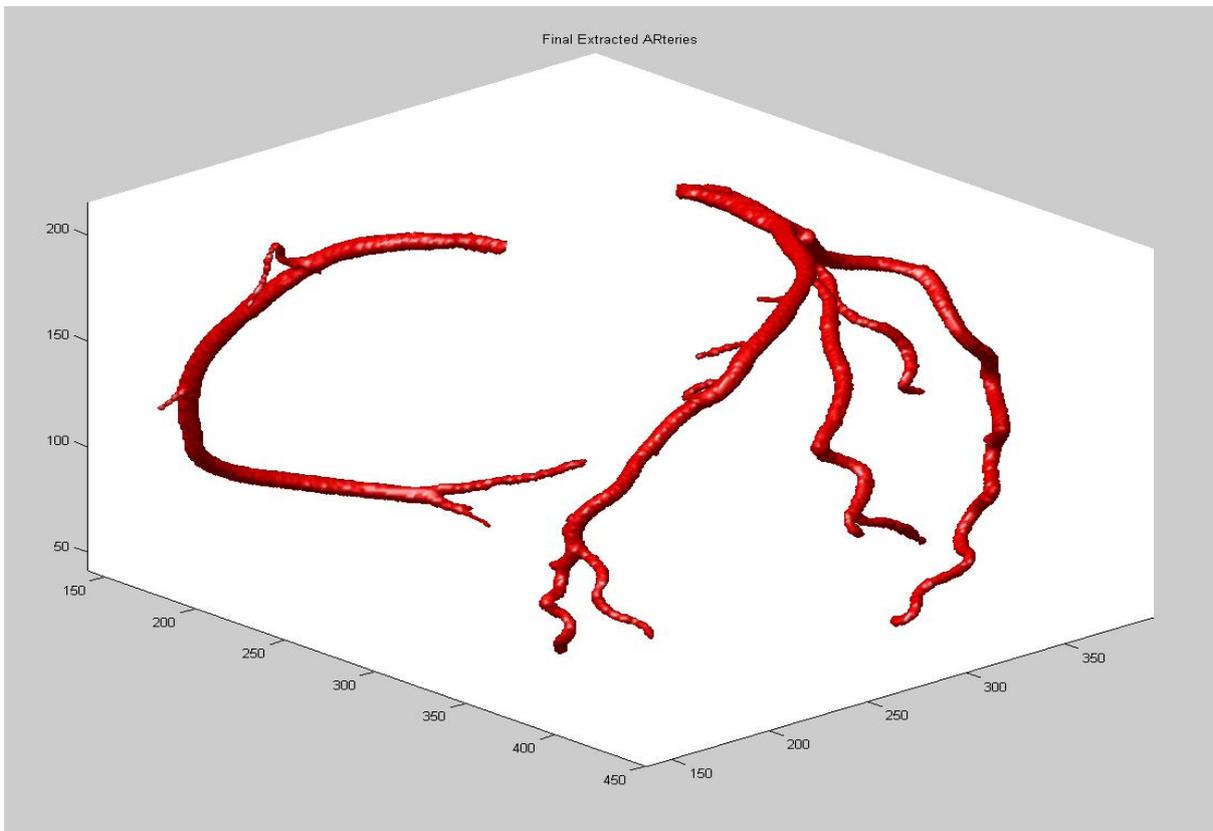

**Segmented Arteries for CTA Volume2   RCA,  LCA and Combined Arterial Structures**

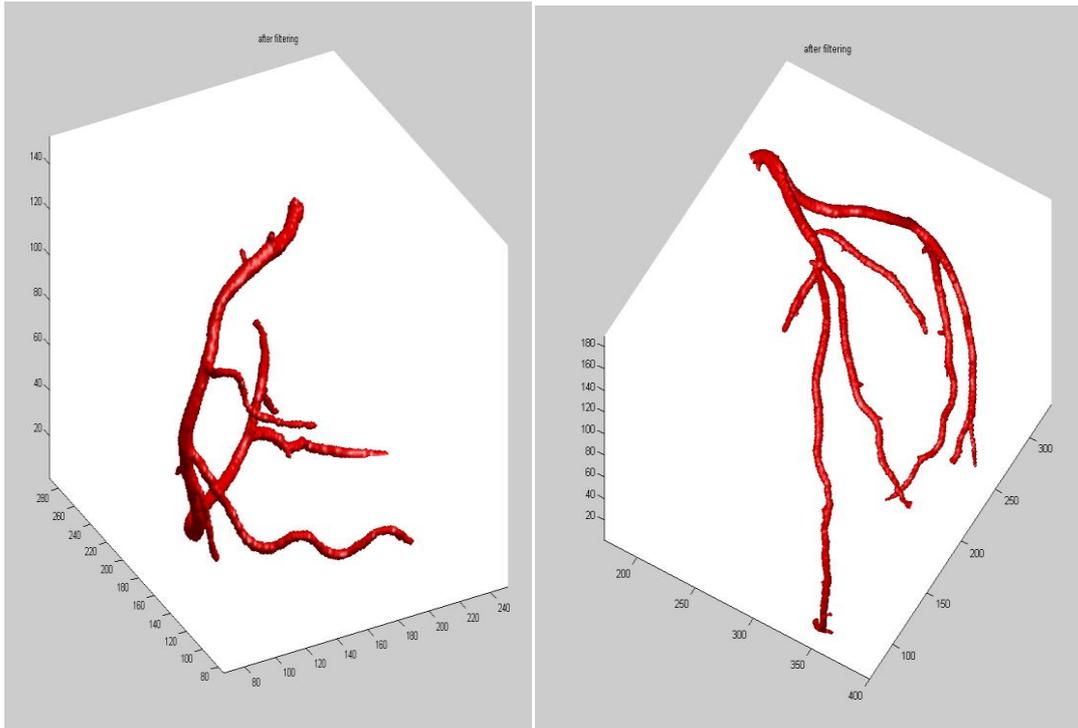

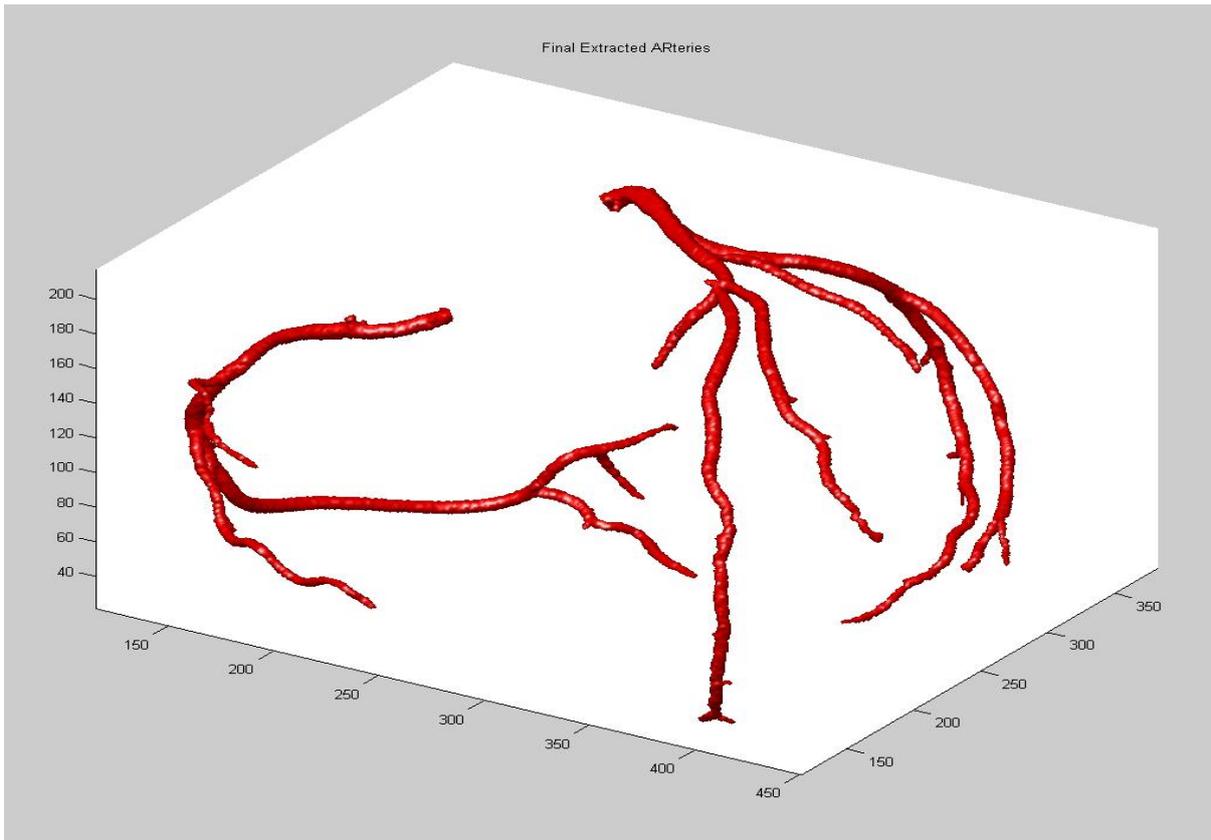

**Segmented Arteries for CTA Volume3   RCA,  LCA and Combined Arterial Structures**

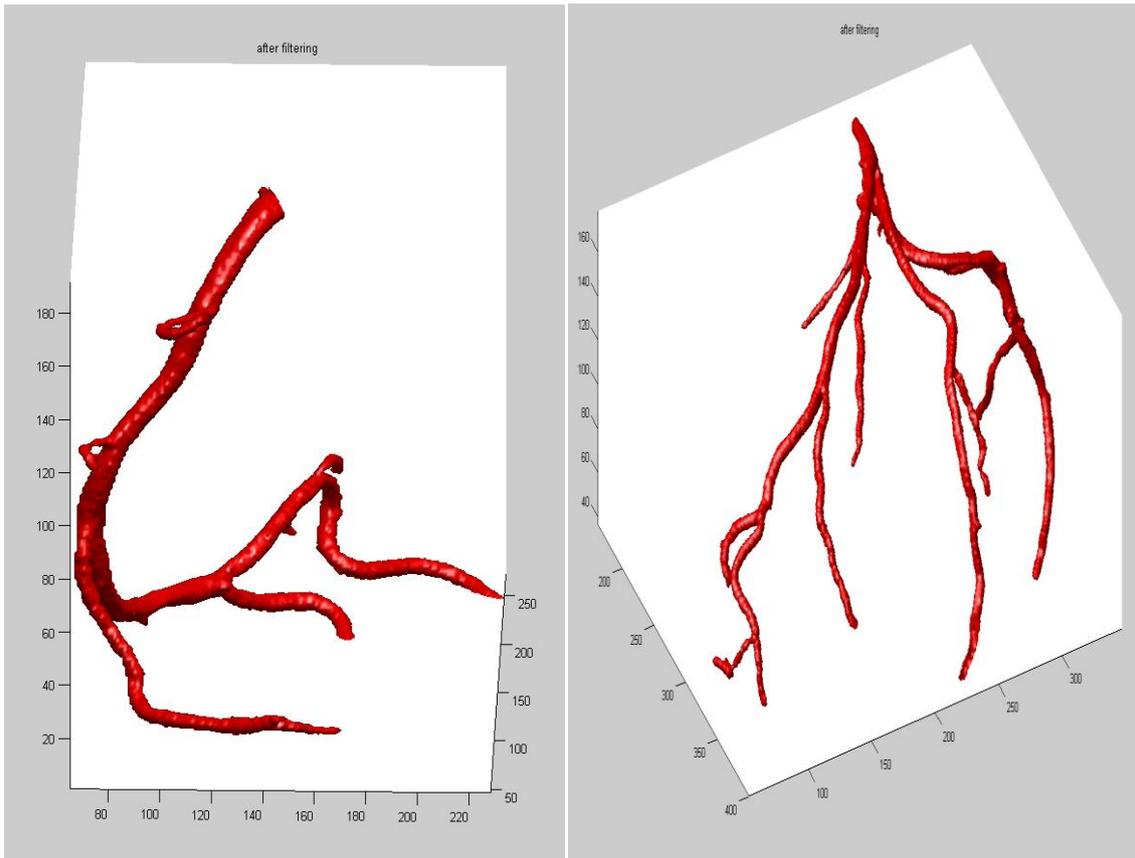

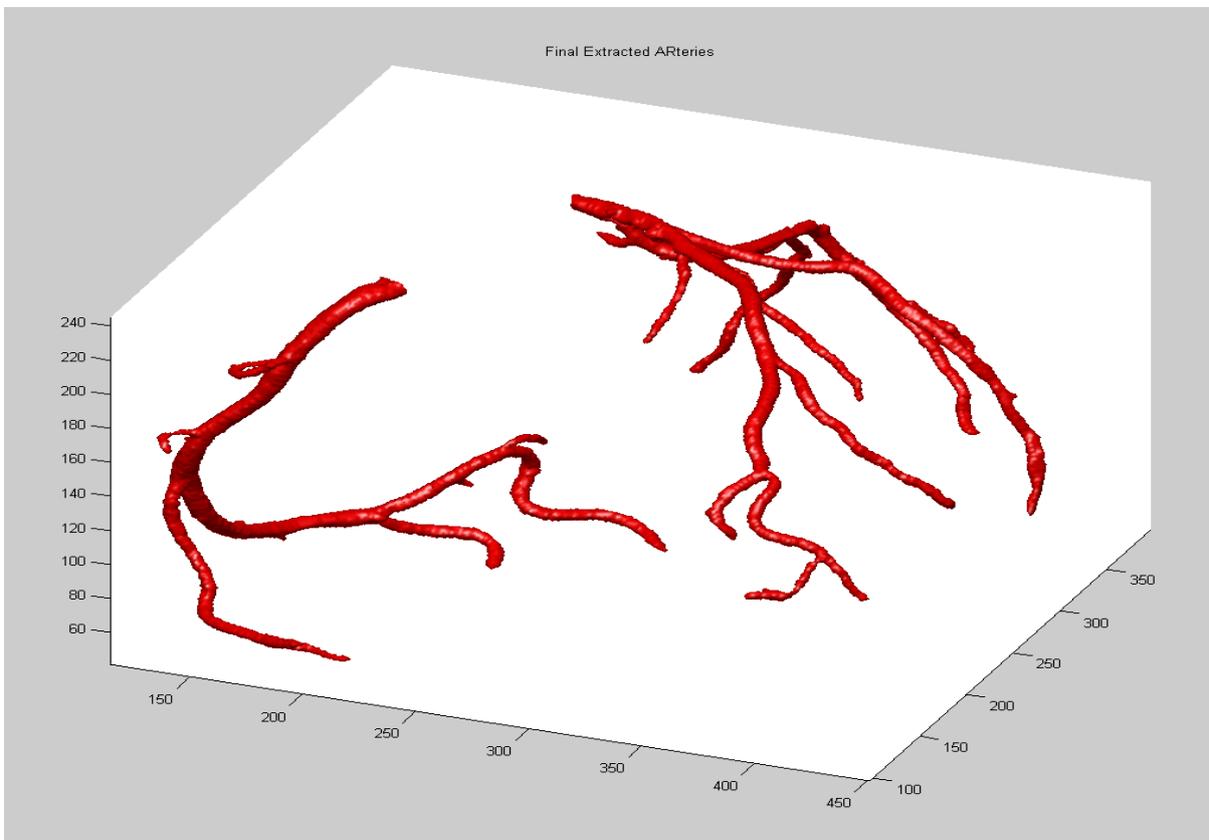

**Segmented Arteries for CTA Volume 4   RCA,  LCA and Combined Arterial Structures**

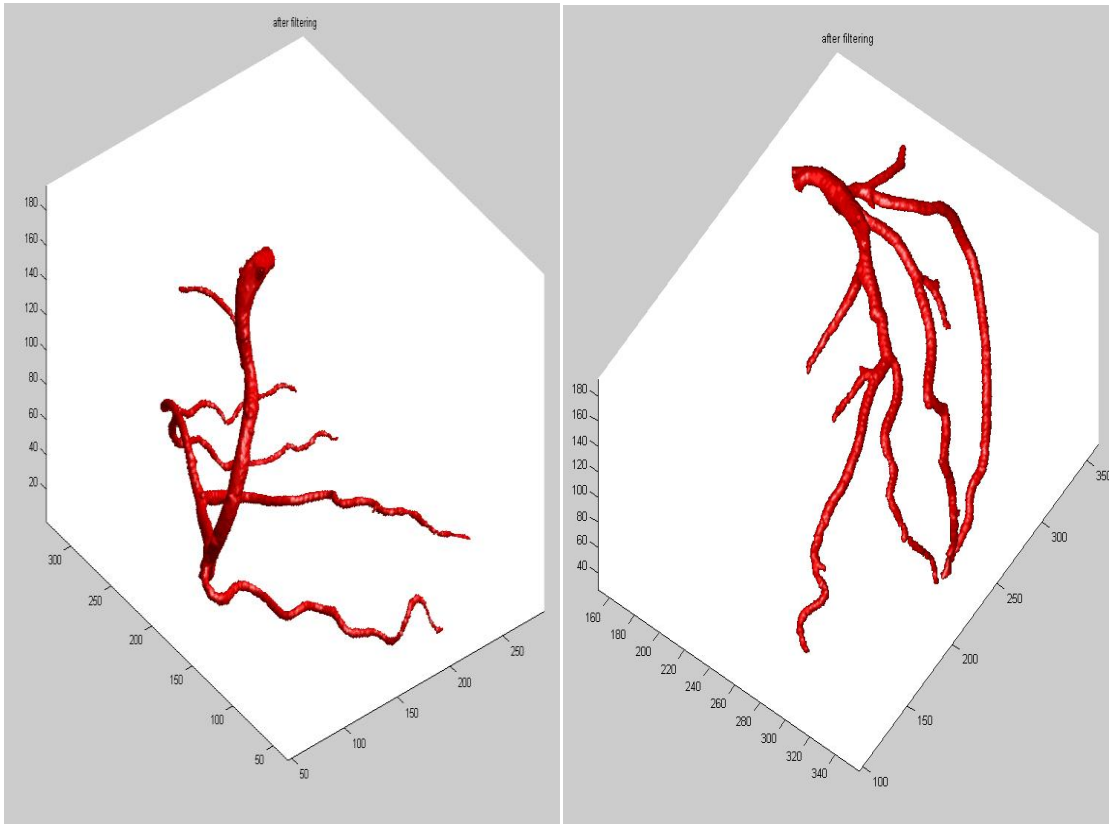

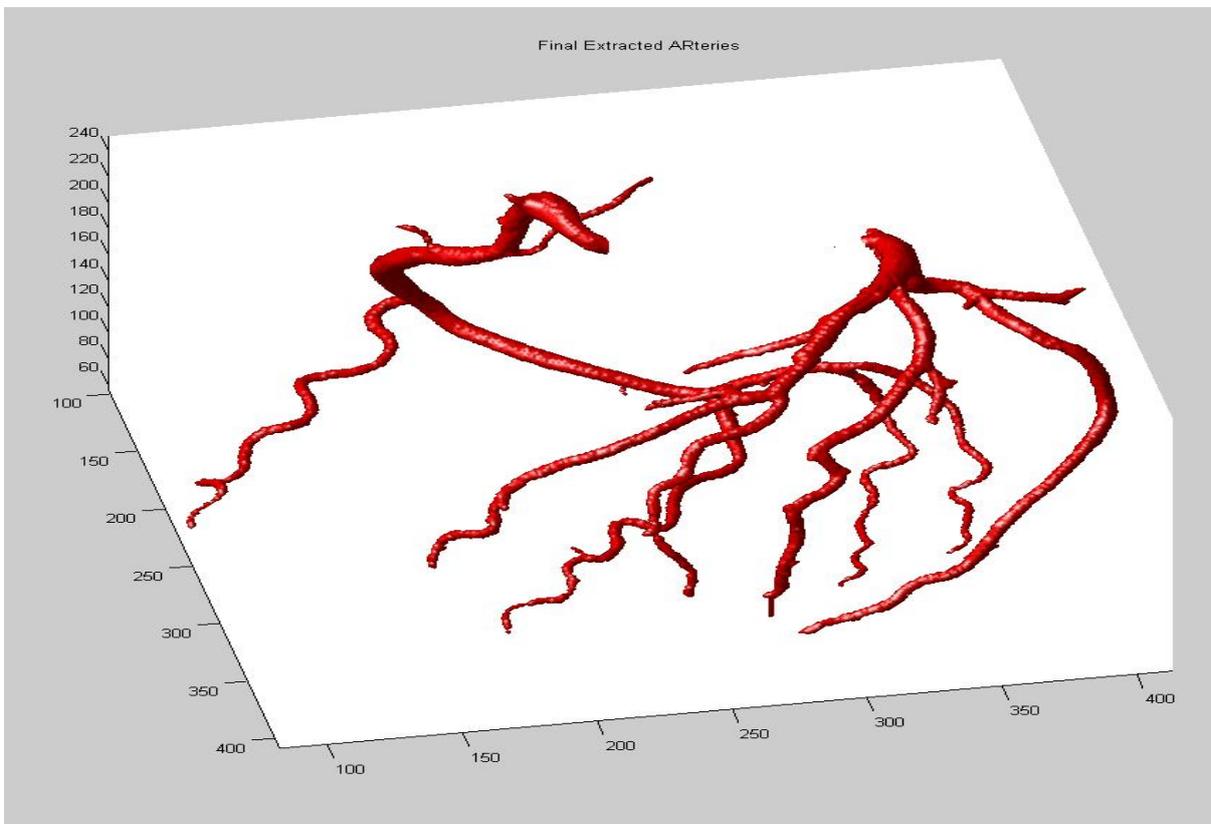

**Segmented Arteries for CTA Volume5   RCA,  LCA and Combined Arterial Structures**

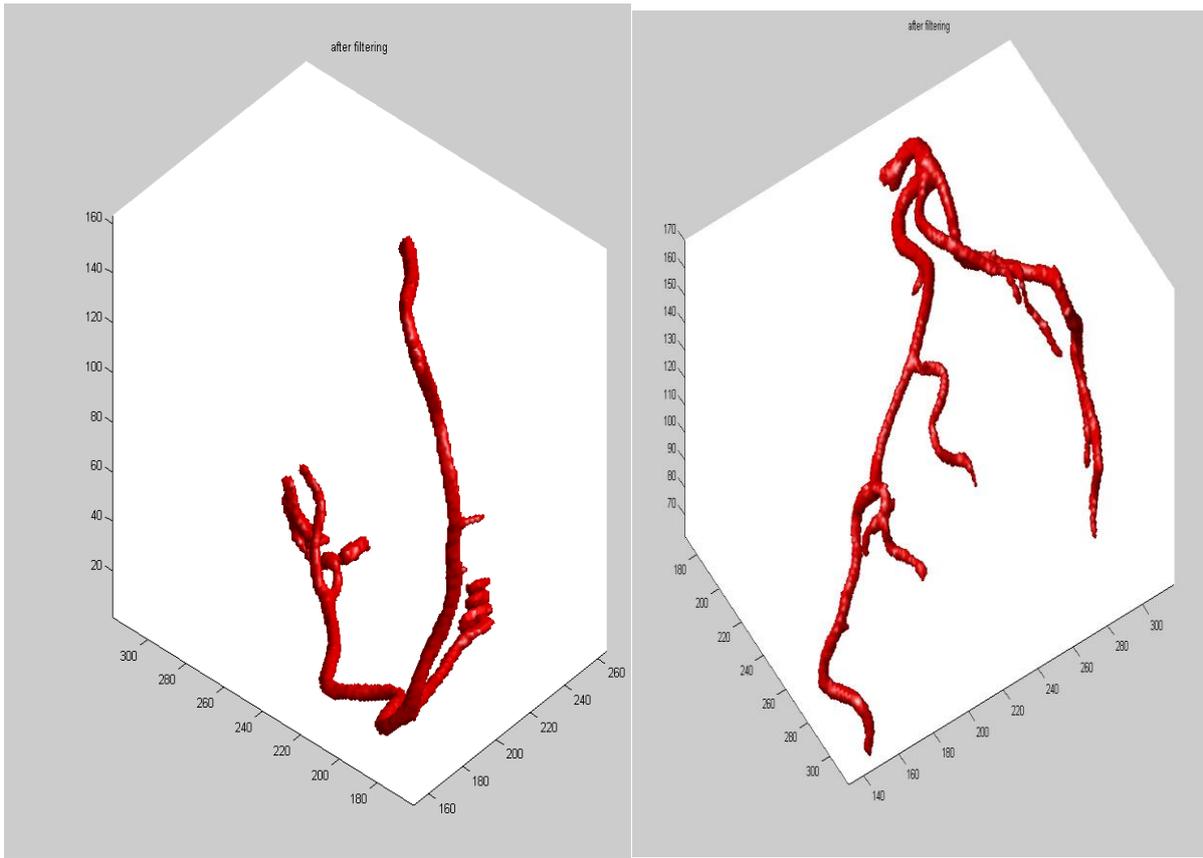

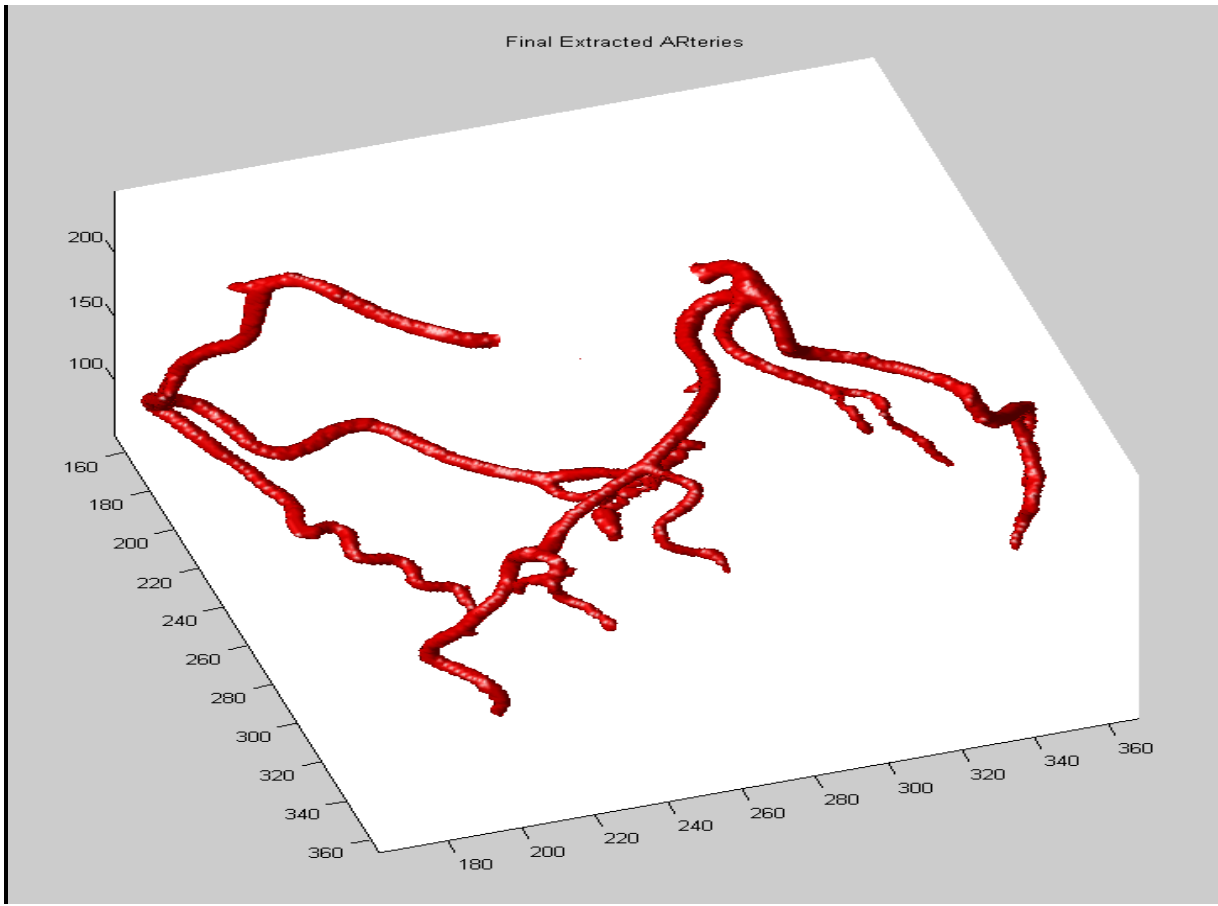

**Segmented Arteries for CTA Volume 6   RCA,  LCA and Combined Arterial Structures**

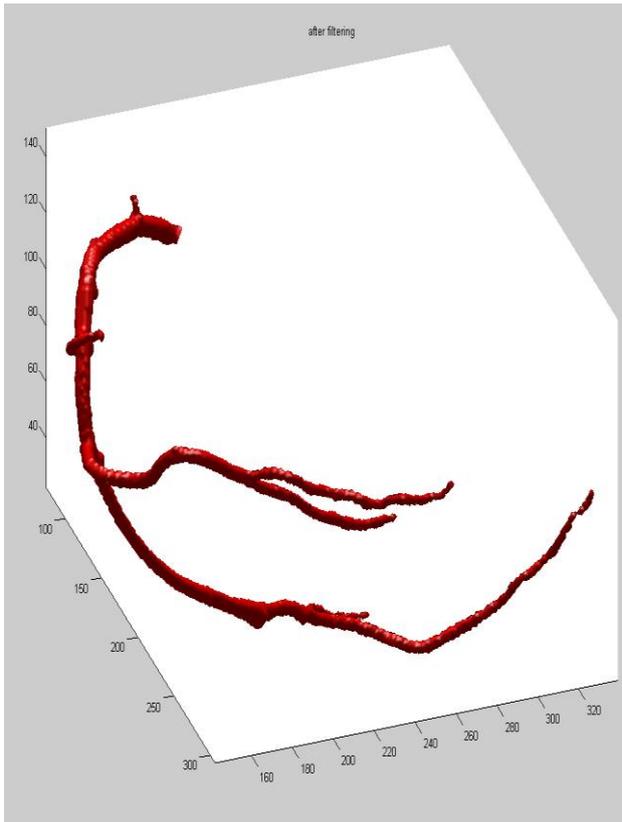

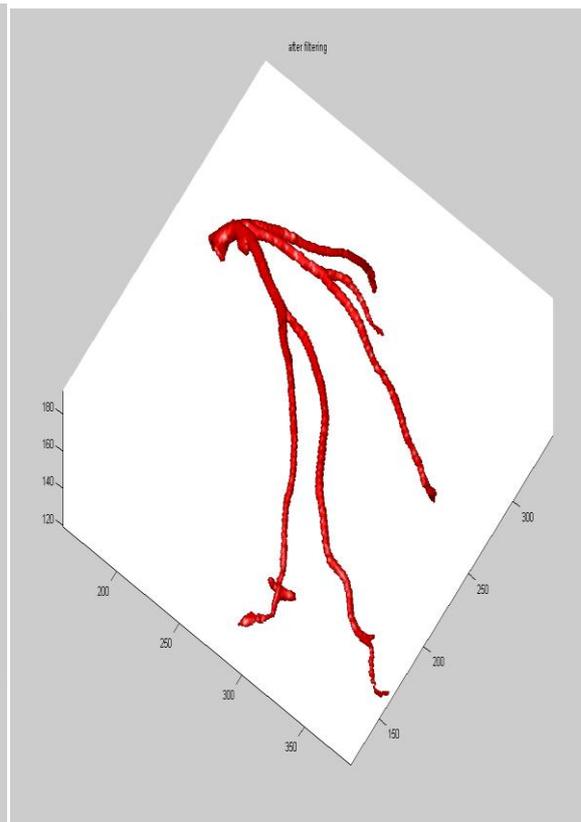

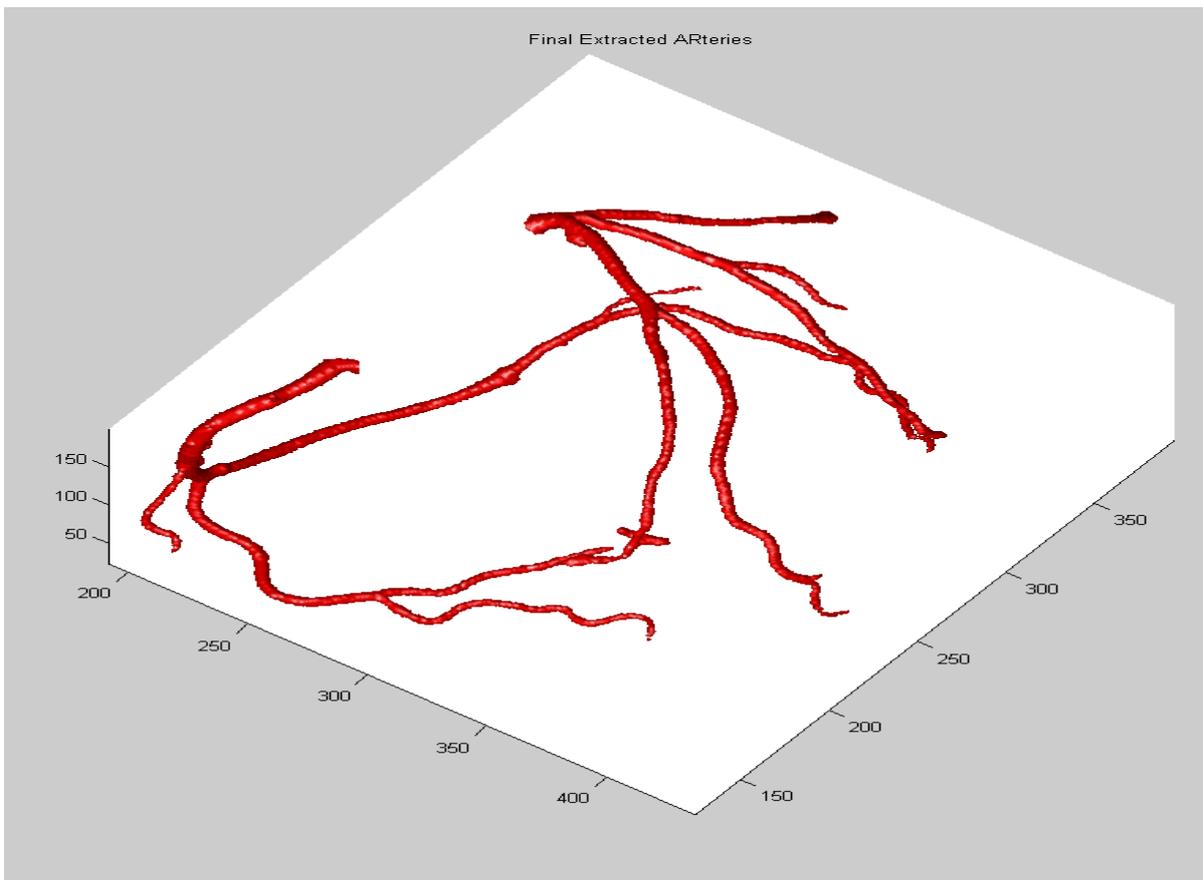

**Segmented Arteries for CTA Volume 7   RCA,  LCA and Combined Arterial Structures**

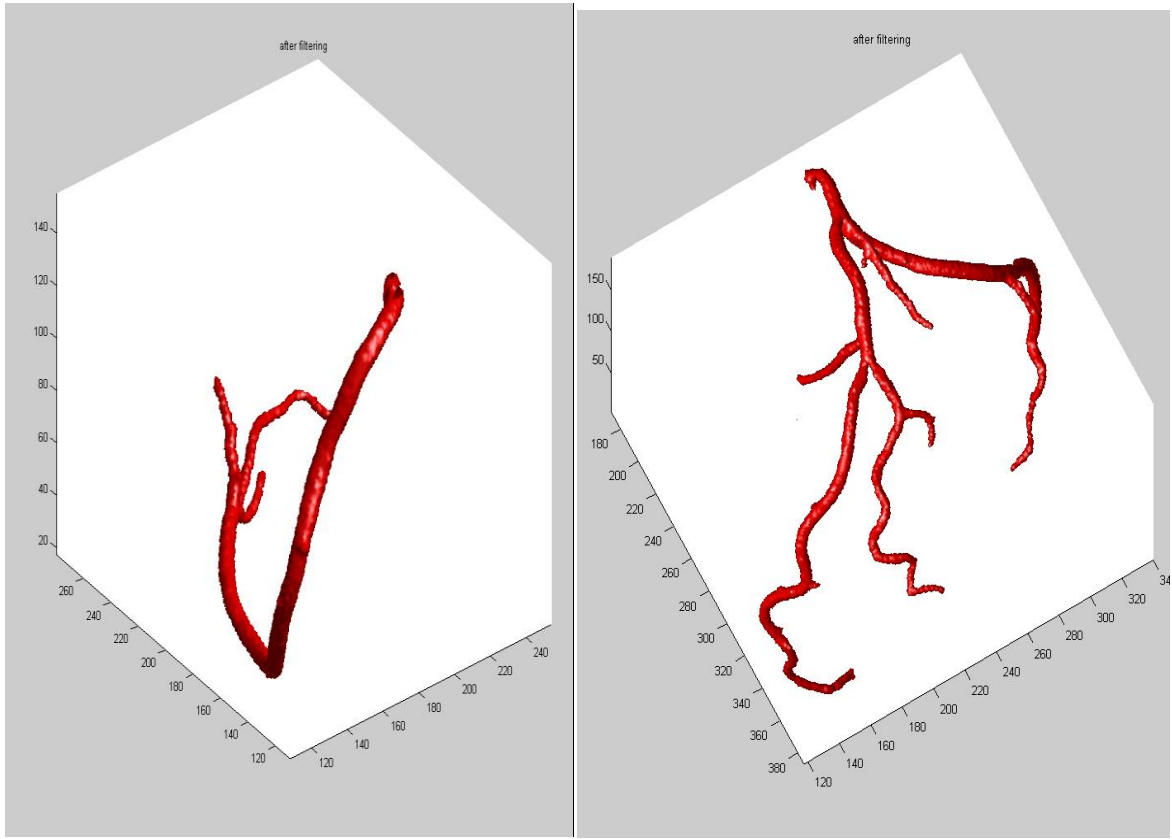

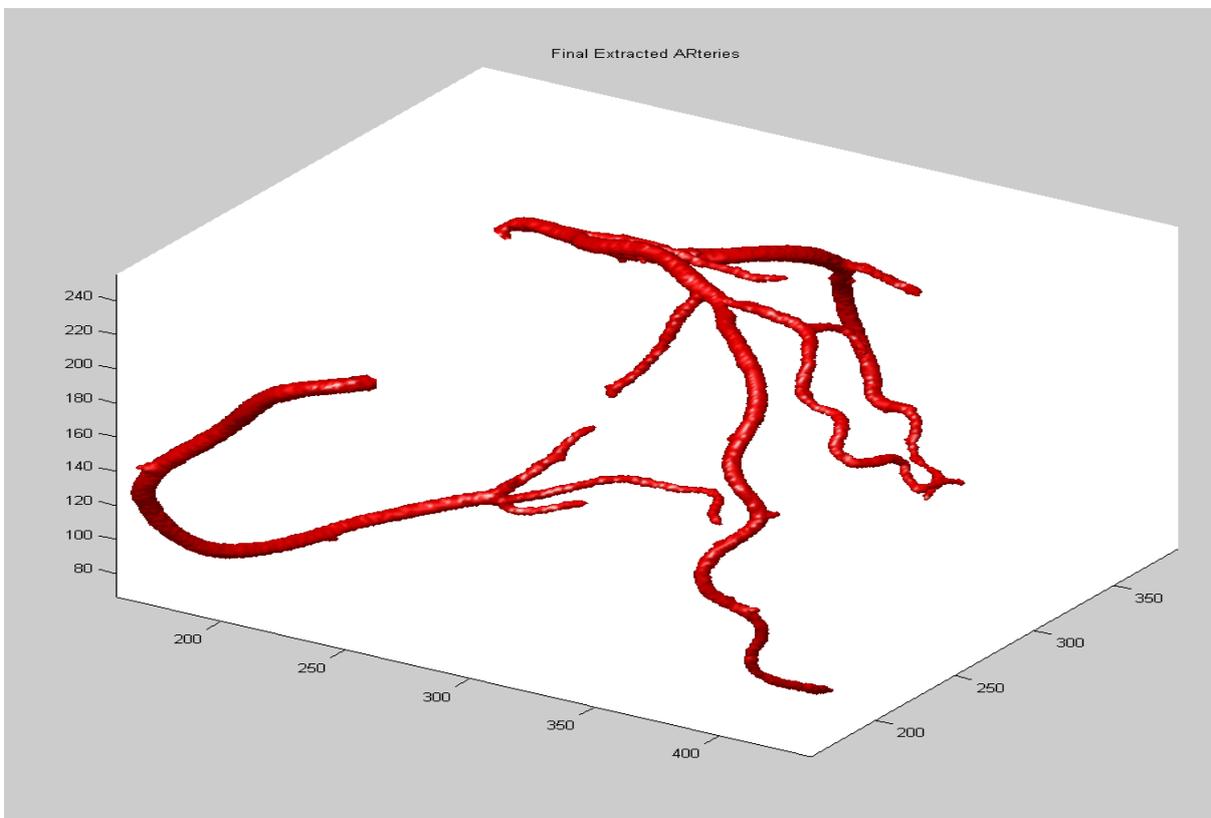

**Segmented Arteries for CTA Volume 8   RCA,  LCA and Combined Arterial Structures**

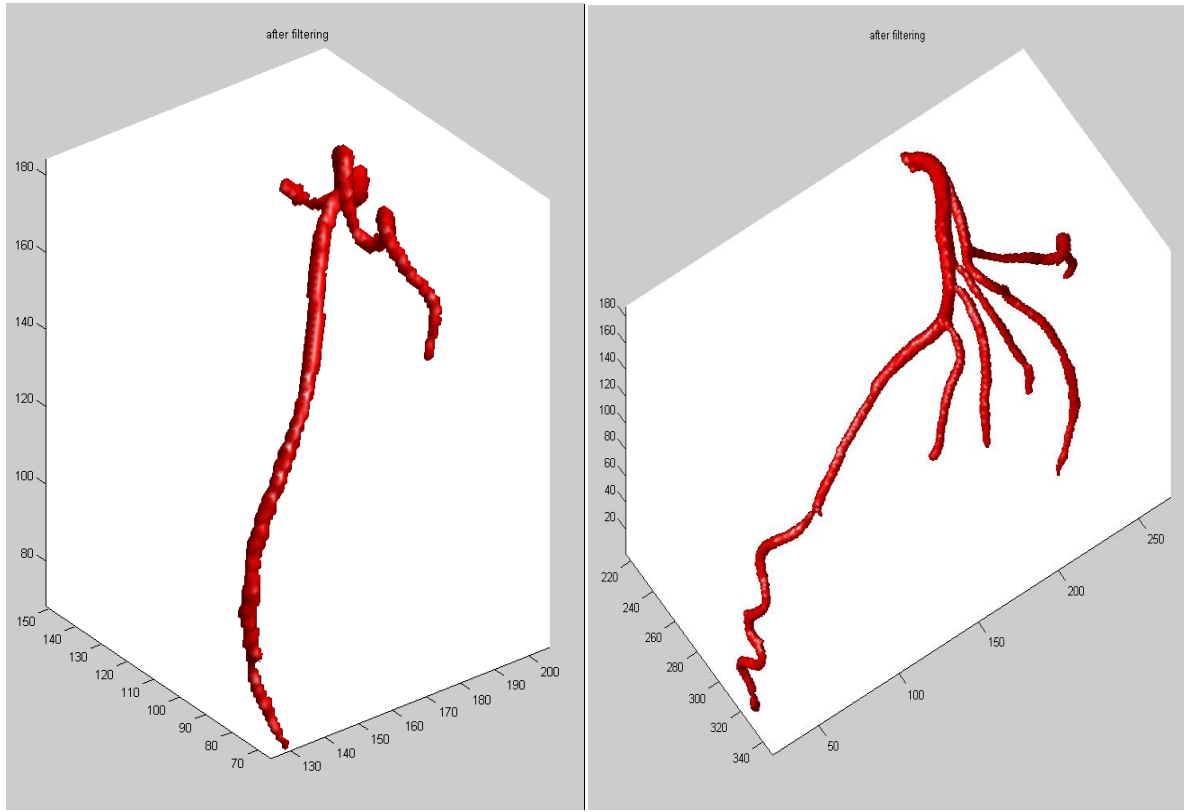

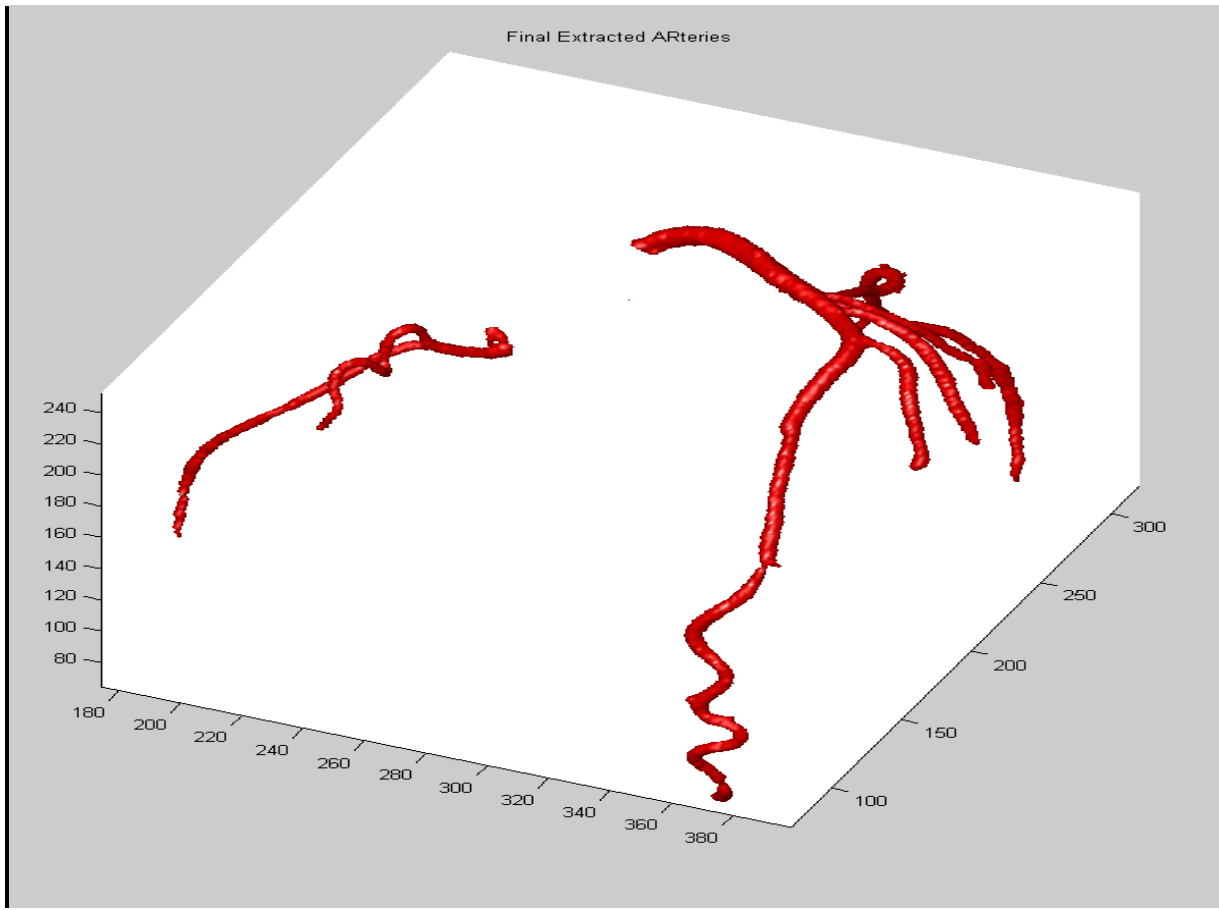

**Segmented Arteries for CTA Volume 9   RCA,  LCA and Combined Arterial Structures**

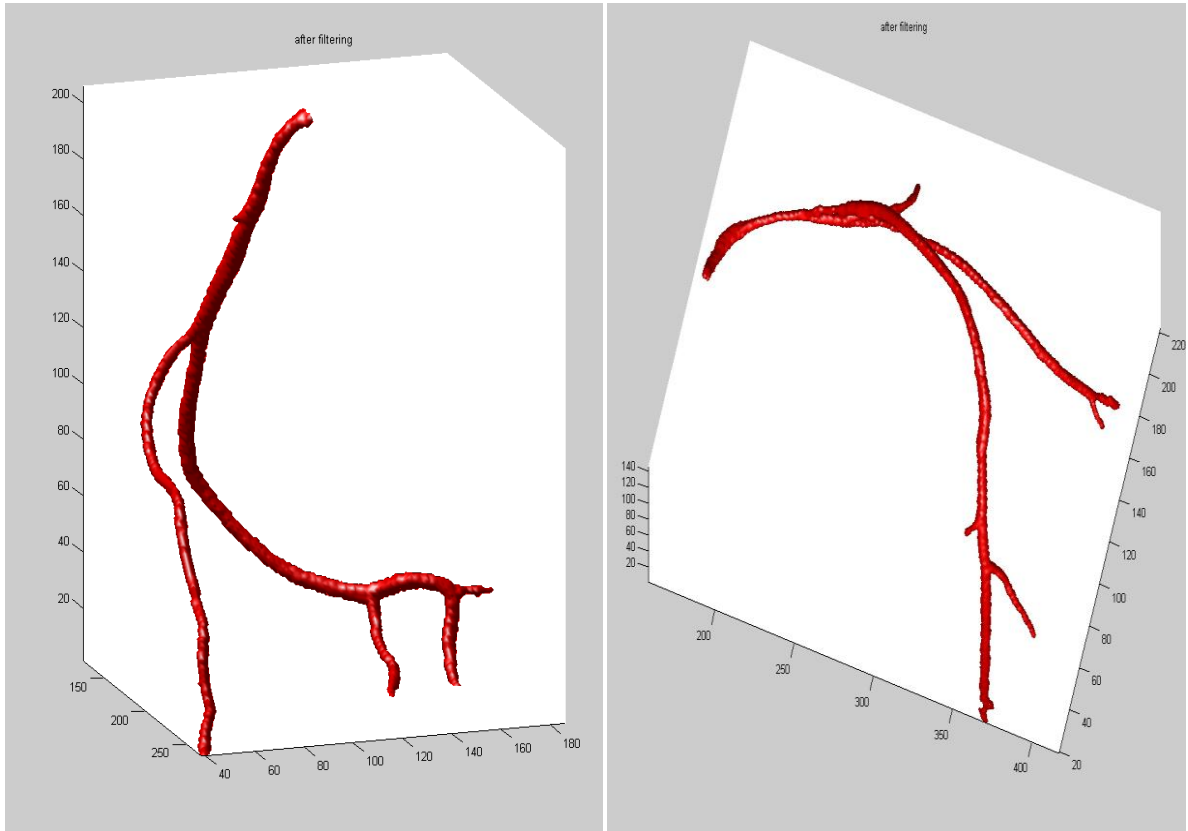

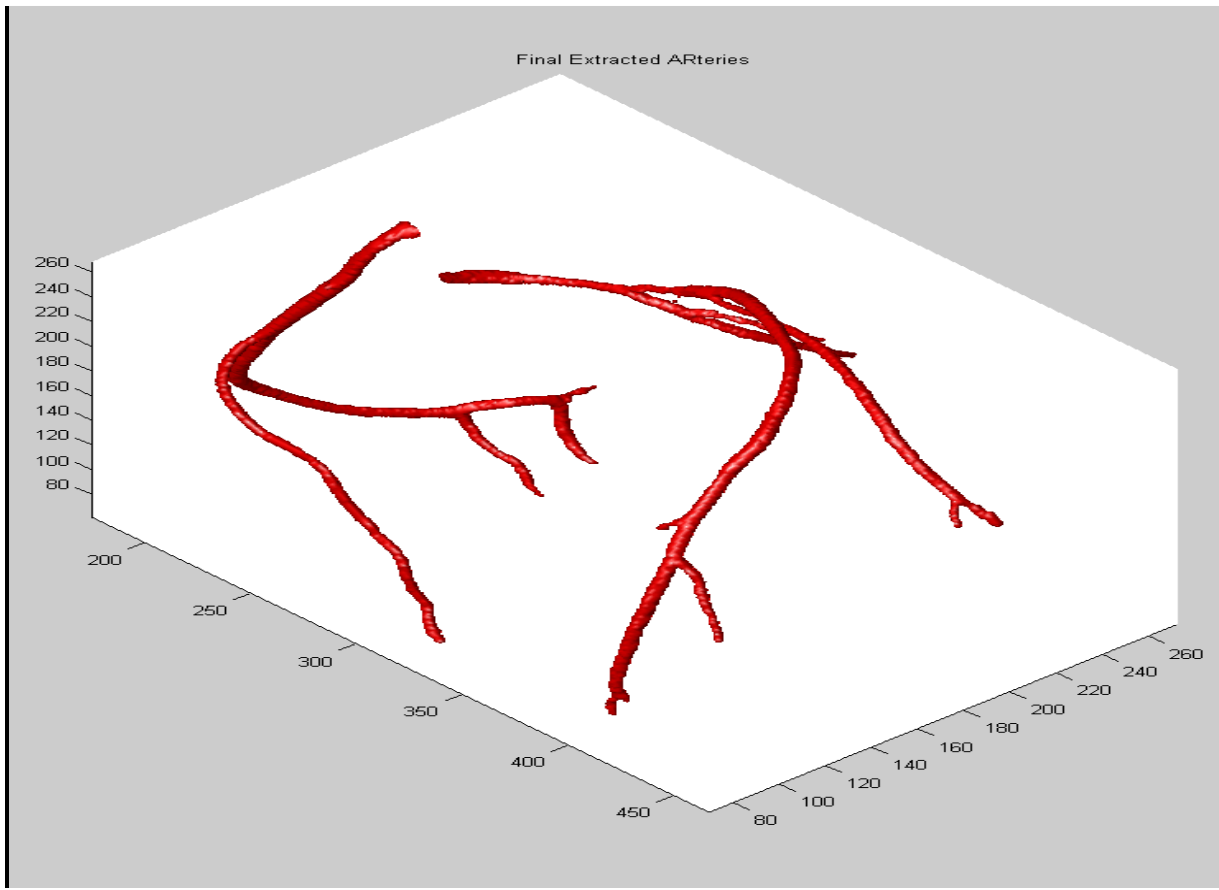

**Segmented Arteries for CTA Volume 10   RCA,  LCA and Combined Arterial Structures**